\documentclass[lettersize,journal]{IEEEtran}
\usepackage{amsmath,amsfonts}
\usepackage{algorithmic}
\usepackage{algorithm}
\usepackage{array}
\usepackage[caption=false,font=normalsize,labelfont=sf,textfont=sf]{subfig}
\usepackage{textcomp}
\usepackage{stfloats}
\usepackage{url}
\usepackage{verbatim}
\usepackage{graphicx}
\usepackage{cite}

\usepackage{cuted}
\usepackage{capt-of} 
\usepackage{float}
\usepackage[T1]{fontenc}
\usepackage{CJKutf8}
\usepackage{makecell}
\usepackage{ragged2e}
\usepackage{tabularx}
\usepackage{booktabs}
\usepackage{multirow}
\usepackage[table]{xcolor}
\usepackage{wrapfig}
\usepackage{arydshln}
\usepackage{threeparttable}
\usepackage{nicefrac}      
\usepackage{microtype}
\usepackage[utf8]{inputenc}
\usepackage{pifont}

\definecolor{mycustompurple}{RGB}{154, 36, 79}
\definecolor{hlcolor}{RGB}{246,254,236}
\definecolor{darkerblue}{rgb}{0,0.08,0.45}
\definecolor{forestgreen}{RGB}{34,139,34}
\definecolor{codemaroon}{RGB}{175, 35, 75}
\usepackage{hyperref}
\hypersetup{
        colorlinks=true,
        linkcolor=darkerblue,
        citecolor=teal,
        filecolor=mycustompurple,
        urlcolor=magenta
    }

\usepackage{tikz}
\usepackage{forest}
\useforestlibrary{edges}
\definecolor{hidden-draw-blue}{RGB}{106,142,189} 
\definecolor{hidden-blue}{RGB}{194,232,247} 

\newcommand{\tg}[1]{\textcolor{gray}{#1}}
\newcommand{\ul}[1]{\underline{#1}}
\newcommand{\tb}[1]{\textbf{#1}}
\newcommand{\cb}[1]{{\setlength{\fboxsep}{0pt}\colorbox[HTML]{AEE4F2}{#1}}}  % gray!25,green!15,green!20,[HTML]{E2EFDA}

\begin{document}

% \title{\raisebox{-0.35em}{\includegraphics[width=0.05\textwidth]{agvbench_logo.png}} 
% AGVBench: A Reliability-Oriented Benchmark of Data Augmentation for Vein Recognition
% }

\title{AGVBench: A Reliability-Oriented Benchmark of Data Augmentation for Vein Recognition
}

\author{
        Haiyang Li$^\star$,
        Yuming Fu$^\star$,
        Qun Song$^\star$,
        Hongchao Liao,
        Jing Chen, \\
        ~\IEEEmembership{Senior,~IEEE,} Mounim A. El-Yacoubi,
        ~\IEEEmembership{Fellow,~IEEE,} Yang Liu,
        Siyuan Ma$^{\dagger}$,
        Xin Jin$^{\dagger}$
        % <-this % stops a space
\thanks{H. Li, Y. Fu, J. Chen and Q. Song are with Chongqing Technology and Business University, Chongqing 400067, China (E-mail: lihaiyang@ctbu.edu.cn).
}
\thanks{X. Jin is with the School of Engineering, Westlake University, Hangzhou, Zhejiang Province, China (E-mail: jinxin4@ctbu.edu.cn).
}
\thanks{M. A. El-Yacoubi is with SAMOVAR, Telecom SudParis, Institute Polytechnique de Paris, 91120 Palaiseau, France (E-mail: mounim.el\_yacoubi@telecom-sudparis.eu).
}
\thanks{H. Liao is with the Guangzhou College of Applied Science and Technology, Guangzhou, Guangdong Province, China.
}
\thanks{S. Ma, Y. Liu are with the Nanyang Technological University, Singapore.
}
\thanks{
This work was supported by Supported by National Natural Science Foundation of China (Grant No. 62506054) and the Science and Technology Research Program of Chongqing Municipal Education Commission (Grant No.KJQN202400841).
}
\thanks{$\star$: Equal contribution. $\dagger$: Corresponding author.
}
}


\markboth{AGVBench: A Reliability-Oriented Benchmark of Data Augmentation for Vein Recognition}
{Shell \MakeLowercase{\textit{et al.}}: A Sample Article Using IEEEtran.cls for IEEE Journals}

% Remember, if you use this, you must call \IEEEpubidadjcol in the second

\maketitle
\begin{abstract}
Vein recognition is a secure biometric technology often constrained by limited annotated data and imaging variations. While data augmentation mitigates this, strategies designed for natural images may disrupt the fine-grained topology and textures essential for identity discrimination. We present AGVBench, which evaluates 30 representative augmentation strategies on five public palm- and finger-vein datasets with seven backbone architectures, covering classic CNNs, vision transformers, and vein-specific recognition models.
Our results show that multi-image mixing methods (e.g., MixUp, PuzzleMix, StarMixup) generally provide the strongest recognition performance. However, they are often poorly calibrated and vulnerable to adversarial perturbations, revealing a clear inconsistency between clean accuracy and adversarial security. We also find that severe geometric transformations frequently degrade recognition, which is potentially due to feature misalignment or spatial cropping, and that augmentation effectiveness varies across palm and finger vein datasets. These findings prove that accuracy-centric evaluation is insufficient for biometric augmentation. AGVBench provides standardized protocols to support reproducible research and guide the design of reliable, secure, and robust vein recognition systems. Our codebase is available at 
\href{https://github.com/Advance-VeinTech-Innovators/AGVBench}{https://github.com/Advance-VeinTech-Innovators/AGVBench}.
\end{abstract}

\begin{IEEEkeywords}
Data Augmentation, Vein Identification, Biometrics, Computer Vision.
\end{IEEEkeywords}

\section{Introduction}
\IEEEPARstart{I}{n} the digital era, biometric authentication, particularly vein recognition~\cite{qin2025wtxgrn, luo2024scutpv, Qin2021pvcnn, kamaruddin2019new}, has become a cornerstone of security due to its unique internal, spoof-resistant, and privacy-preserving vascular patterns. While deep learning architectures have significantly enhanced discriminative performance, their efficacy remains constrained by a persistent ``small-sample'' dilemma inherent in biometric acquisition. To mitigate this data scarcity, researchers extensively employ data augmentation~\cite{cubuk2019autoaugment, cubuk2020randaugment, devries2017cutout, Zhang2018mixup, jin2025starmixup} (DA). However, most studies inherit strategies optimized for natural images, often overlooking the fundamental morphological differences between semantic objects and vascular structures. This empirical transfer can be counterproductive, as aggressive transformations risk obliterating the delicate topological and high-frequency details essential for identity discrimination. These issues underscore a critical need for a specialized evaluation framework.

\textbf{\textit{Why do we call for a vein-specific augmentation benchmark?}} While advanced augmentation strategies~\cite{Liu2022automix, qin2024adautomix, kang2023guidedmixup} have shown immense potential in natural image tasks, their direct application to vascular biometrics remains poorly understood due to the delicate and high-frequency nature of vein patterns. Without a systematic evaluation, it is uncertain whether complex augmentations, such as policy-based~\cite{cubuk2019autoaugment, cubuk2020randaugment, muller2021trivialaugment, suzuki2022teachaugment} or mixup-based~\cite{Zhang2018mixup, Yun2019cutmix} methods, consistently serve as superior alternatives to traditional ones across diverse neural architectures like Convolutional Neural Networks~\cite{He2016deep} (CNNs) and Vision Transformers~\cite{Dosovitskiy2021vit} (ViTs). Moreover, a thorough and standardized assessment of how these techniques affect biometric fidelity is conspicuously missing in the community. A benchmark is precisely the mechanism to establish such an understanding, playing a pivotal role in driving research progress by integrating an agreed-upon set of tasks, impartial comparisons, and rigorous assessment criteria. However, there has been a lack of a comprehensive benchmark for vein recognition to facilitate unbiased comparisons and practical deployment.

\textbf{\textit{Why do we need an open-source vein augmentation codebase?}} Notably, most existing data augmentation techniques used in vein recognition are implemented with varied settings, hyperparameters, and coding styles across different research papers. This lack of standardization not only hinders user-friendly reproduction and deployment but also imposes costly trial-and-error on practitioners to determine the most effective augmentation strategy for their specific biometric needs in real-world applications. Hence, it is essential to develop a unified vein representation learning codebase for standardized data pre-processing, diverse augmentation module selection, network architecture integration, model training, and empirical analysis. Such a platform would bridge the gap between theoretical research and practical implementation, fostering further innovation in robust vein recognition systems.

In this paper, we introduce AGVBench, the first comprehensive benchmark specifically curated to rethink and evaluate data augmentation strategies for vein recognition. Our work addresses the existing knowledge gap by providing a rigorous experimental framework and actionable insights. The primary contributions of this research can be summarized as:
\begin{itemize}
    \item We construct AGVBench, the first large-scale benchmark for data augmentation in vein recognition, systematically evaluating 30 methods on five datasets and seven architectures across six dimensions: recognition and verification performance, calibration, corruption robustness, adversarial robustness, occlusion robustness, and computational efficiency.
    \item We reveal a critical decoupling among reliability metrics: top-performing mixup methods in clean accuracy often exhibit severe calibration errors and adversarial vulnerability, demonstrating the necessity of multidimensional evaluation.
    \item We establish a standardized experimental protocol and release our complete library to the community. This serves as a foundational resource for facilitating reproducible research and guiding the design of more robust, vein-specific augmentation pipelines.
\end{itemize}

\begin{figure*}[!ht]
    \centering
    \includegraphics[width=1.0\textwidth]{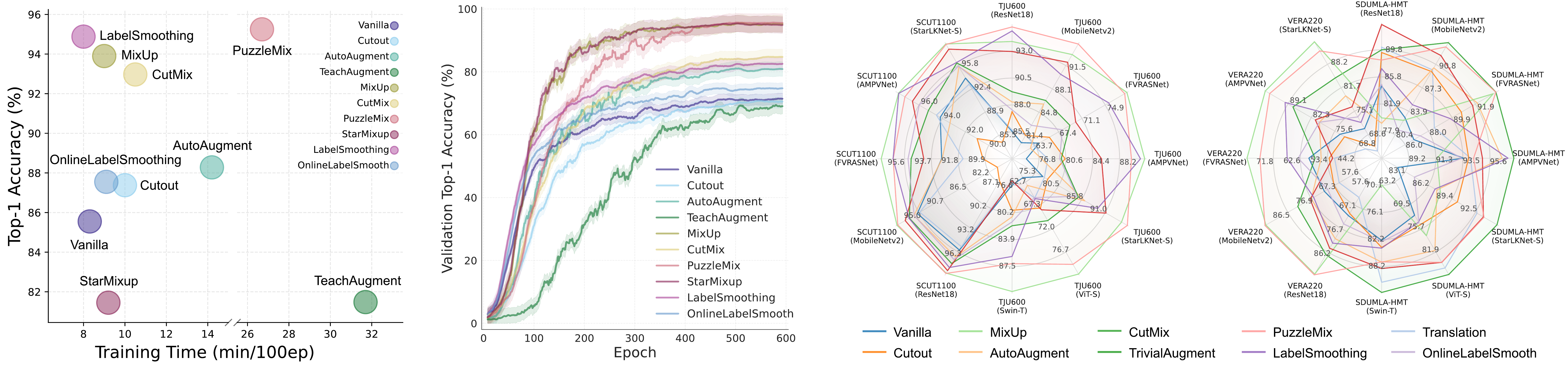}
    \vspace{-2.em}
    \caption{Comprehensive performance evaluation of various data augmentation and regularization methods. (a) Trade-off of Accuracy \textit{vs.} Training Time. (b) The fitting curve of accuracy and training epochs. (c) Evaluation across various vein datasets and network architectures via radar charts.}
    \vspace{-1.em}
    \label{fig:teaser}
\end{figure*}
\section{Background and Related Works}
\label{sec:2}

\subsection{Problem Formulation}
\label{sec:2.1}
The core objective of vein recognition is to learn a discriminative mapping function that projects raw vascular patterns into a high-dimensional feature space for identity verification. Let $\mathcal{X} = \{x_i \mid x_i \in \mathbb{R}^{H \times W \times C}\}^N_{i=1}$ denote the input space of vein images, where $H$, $W$, and $C$ represent the height, width, and the number of channels, respectively. Although raw near-infrared vein images are typically grayscale ($C=1$), $C$ is retained to accommodate standard backbone architectures that may duplicate channels for compatibility. Let $\mathcal{Y} = \{1, 2, \dots, K\}$ be the discrete label space, where $K$ is the total number of identity classes. A typical vein dataset containing $N$ samples is defined as $\mathcal{D} = \{(x_i, y_i)\}_{i=1}^N$, where $x_i \in \mathcal{X}$ and its corresponding label $y_i \in \mathcal{Y}$.

The goal is to optimize a deep neural network $f_\theta(\cdot)$, where $\theta$ denotes the learnable parameters of the model. We aim to minimize the empirical risk:
\begin{align}
    \mathcal{R}(\theta) = \frac{1}{N} \sum_{i=1}^{N} \mathcal{L}_\text{CE}\big(f_\theta(x_i), y_i \big),
\end{align}
where $\mathcal{L}_\text{CE}(\cdot)$ represents the standard Cross-Entropy loss function. For a sample $(x_i, y_i)$, it is formulated as:
\begin{align}
    \mathcal{L}_\text{CE}\big(f_\theta(x_i), y_i \big) = -\log \big( f_\theta(x_i)_{y_i} \big),
\end{align}

However, in the ``small-sample'' context of AGVBench, the empirical distribution of $\mathcal{D}$ is a sparse approximation of the true distribution $P(\mathcal{X}, \mathcal{Y})$. To bridge this gap, here we introduce a generalized augmentation policy $\mathcal{A}$ that generates virtual samples $(\tilde{x}, \tilde{y})$ from the vicinity of the original training data $(x, y)$. Since a practical training pipeline implicitly incorporates both original and synthesized data (e.g., by treating the identity mapping as a subset of $\mathcal{A}$), the augmented training objective can be rigorously formulated by conditioning the augmentation distribution on the original empirical distribution:
\begin{align}
    \min_{\theta} \mathbb{E}_{(x, y) \sim \mathcal{D}} \left[ \mathbb{E}_{(\tilde{x}, \tilde{y}) \sim \mathcal{A}(x, y)} [\mathcal{L}_\text{CE}\big(f_\theta(\tilde{x}), \tilde{y}\big)] \right].
\end{align}
Here, policy $\mathcal{A}$ acts as a specific conditional data augmentation mapping, and different designs of this mapping can significantly alter the distribution of vein features in the latent space.

\begin{figure*}[ht]
    \centering
    \includegraphics[width=0.98\linewidth]{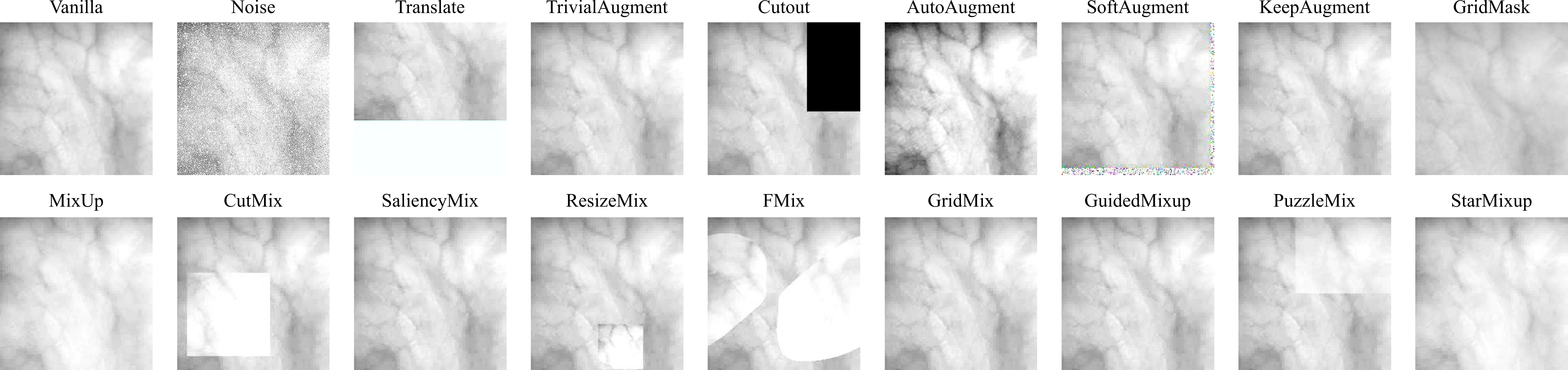}
    \caption{Visualization of various data augmentation techniques applied to a sample vein image. The original (Vanilla) image is shown alongside standard single-image transformations (top row) and advanced image mixing strategies (bottom row).}
    \label{fig:visual}
    \vspace{-1.em}
\end{figure*}

\subsection{Data Augmentation}
\label{sec:2.2}
Data augmentation techniques artificially expand dataset diversity to enhance model generalization and robustness, broadly encompassing Single Image Augmentation, Multi-Image Augmentation, and Label Enhancement.
Single-image augmentation methods apply various transformations to individual training samples. These approaches have evolved from foundational geometric and pixel-level perturbations to advanced strategies that force networks to learn globally distributed features via spatial occlusion~\cite{devries2017cutout, zhong2020randome}. To address the suboptimal nature of manual designs, researchers leveraged search algorithms to dynamically identify optimal transformation sequences~\cite{cubuk2019autoaugment, cubuk2020randaugment, muller2021trivialaugment}, while also developing sophisticated operations such as saliency-guided~\cite{gong2021keepaugment} and learnable transformations~\cite{suzuki2022teachaugment}.
To expand the virtual sample space, multi-image augmentation methods construct smoother decision boundaries by blending multiple images and interpolating their corresponding labels. This paradigm originated from global linear interpolation across sample pixels~\cite{Zhang2018mixup}, and was refined by spatial mask-based cutting and pasting to preserve local structural integrity while blending~\cite{Yun2019cutmix, Baek2021gridmix, takahashi2019ricap, jin2025starmixup}. To prevent the loss of critical foreground information, subsequent works incorporated saliency-guided priors~\cite{Uddin2020saliencymix, Kim2020puzzle} and automated end-to-end adaptive mixing frameworks~\cite{Liu2022automix, qin2024adautomix}. Furthermore, recent advancements have expanded these mixing paradigms into the generative domain, leveraging diffusion models and self-saliency mechanisms to synthesize context-aware and label-preserving mixed samples~\cite{islam2026s, islam2025context, islam2026genmix, islam2024diffusemix}.
Beyond manipulating visual inputs, label enhancement approaches alleviate model overconfidence by directly regularizing the training targets. The methodology has evolved from applying static uniform softening to one-hot vectors~\cite{szegedy2016labelsmooth, pereyra2017confidencepenalty}, to generating adaptive dynamic soft labels using historical predictions~\cite{zhang2021onlinelabelsmooth} or parameterized distributions~\cite{cheng2021dirichletlabelsmoothLoss}.

\subsection{Advanced Vein Recognition}
\label{sec:2.3}
Due to the strong absorption of near-infrared light by hemoglobin, infrared imaging captures highly discriminative vein patterns for biometrics~\cite{Qin2021pvcnn}. However, the fine-grained nature and topological sensitivity of vein structures make robust feature extraction particularly challenging.

Early vein recognition primarily relied on traditional methods, based on hand-crafted feature extractors~\cite{kang2014contactless,syarif2017enhanced} or traditional machine learning algorithms~\cite{ponnusamy2019palm,kamaruddin2019new,shazeeda2019finger}. Despite their computational efficiency, these heuristic-driven approaches heavily depend on prior knowledge. Consequently, they exhibit limited representational capacity and robustness against image noise, spatial misalignment, and complex backgrounds, leading to their gradual obsolescence in tackling real-world modality degradations.
To overcome the bottlenecks of manual feature engineering, deep learning has been widely adopted due to its powerful end-to-end automated representation learning capabilities. Deep learning-based methods can automatically extract profound, highly discriminative features directly from raw pixels. Besides optimizing system security~\cite{Yang2020fvrasnet} and lightweight architectures~\cite{shen2021finger}, deep learning-based vein recognition methods primarily focus on improving feature representation. To address the complex topological branches of blood vessels, recent network architectures emphasize extracting structure-aware information. These methods typically incorporate attention mechanisms \cite{qin2025wtxgrn} or adaptive spatial modeling components \cite{Qin2021pvcnn,luo2024scutpv,luo2025rsnet} to capture fine-grained geometric features, thereby improving the generalization performance of the model.

Nevertheless, the superior performance of deep neural networks is inherently contingent upon massive amounts of training data. Constrained by the requirement for specialized infrared capture devices and privacy concerns, existing publicly available vein datasets are generally small-scale and lack sample diversity. This severe data scarcity easily causes heavily parameterized deep models to overfit. Consequently, investigating and applying effective data augmentation techniques to artificially expand dataset scale and feature boundaries has become crucial for unlocking the full potential of deep learning and building highly robust vein recognition systems.
\section{AGVBench: Benchmark Design}
\label{sec:3}
\begin{figure*}[t]
    \centering
    \includegraphics[width=0.9\linewidth]{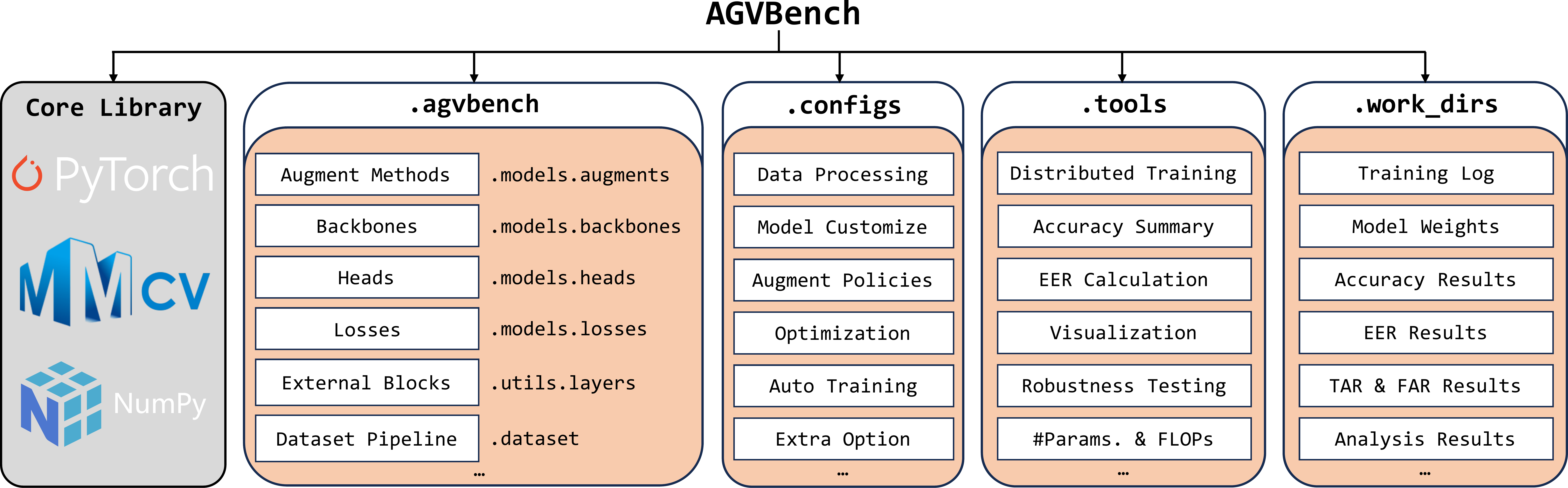}
    \caption{Overview of the AGVBench codebase framework. Built upon PyTorch and MMCV, the framework is structured into four functional modules: (1) \textcolor{codemaroon}{\texttt{.agvbench}} for data pipeline and model component registration; (2) \textcolor{codemaroon}{\texttt{.configs}} for centralized hyperparameter and experimental setup; (3) \textcolor{codemaroon}{\texttt{.tools}} for distributed training execution and comprehensive metric evaluation; and (4) \textcolor{codemaroon}{\texttt{.work\_dirs}} for storing training logs, model weights, and analysis results.}
    \vspace{-1.em}
    \label{fig:codebase}
\end{figure*}

This section introduces the AGVBench framework design from five aspects: supported augmentation methods, backbone models, datasets, evaluation protocol, and the experimental pipeline of the codebase. AGVBench provides a unified framework implemented in PyTorch and built upon the OpenMMLab Computer Vision Foundation (MMCV) ecosystem for model design, training, and evaluation.
We start with the overview of the composition of AGVBench. As illustrated in Fig. \ref{fig:codebase}, AGVBench is decoupled into multiple modular components, including model architectures (\textcolor{codemaroon}{\texttt{.agvbench.models.backbones}}), data preprocessing pipelines (\textcolor{codemaroon}{\texttt{.agvbench.datasets}}), augmentation methods (\textcolor{codemaroon}{\texttt{.agvbench.models.augments}}), and execution scripts (\textcolor{codemaroon}{\texttt{.tools}}). 
Vision models are divided into standard building blocks (e.g., necks and heads) in \textcolor{codemaroon}{\texttt{.agvbench.models}}. This enables researchers to easily construct and modify models by combining different components through centralized configuration files in \textcolor{codemaroon}{\texttt{.configs}}. Consequently, users can readily customize specific vein recognition pipelines and training strategies. The detailed benchmarking configurations and theoretical designs are discussed in the subsequent subsections.

\subsection{Augmentation Schemes}
\label{sec:3.1}

\textbf{Single Image Augmentation Methods:}
Single-image augmentation schemes incorporate methods ranging from basic operations to policy-driven methods. Basic operations perform foundational geometric and photometric perturbations, including Flip, Rotate, Affine Transformation (Trans.), Blur, and Noise. Additionally, to evaluate regional spatial robustness, Cutout~\cite{devries2017cutout}, RandomErasing~\cite{zhong2020randome}, and GridMask~\cite{chen2020gridmask} implement hand-crafted spatial occlusion and dropping policies. 
Policy-driven methods, such as AutoAugment~\cite{cubuk2019autoaugment}, RandAugment~\cite{cubuk2020randaugment}, and TrivialAugment~\cite{muller2021trivialaugment}, use predefined search spaces to discover optimal augmentation policies. Furthermore, advanced dynamic and domain-adaptive strategies are integrated, where TeachAugment~\cite{suzuki2022teachaugment} and KeepAugment~\cite{gong2021keepaugment} apply importance-guided or teacher-guided transformations, while YOCO~\cite{han2022yoco}, SoftAugment~\cite{liu2023softaug}, and RQ~\cite{wu2023rq} perform structural and feature-level augmentations specific to complex visual patterns.

\textbf{Multi Image Augmentation Methods:}
Multi-image augmentation methods involve cross-sample feature interaction and label interpolation. Foundational methods like MixUp~\cite{Zhang2018mixup} perform global pixel-level or feature-level convex interpolation. To preserve local structural integrity, CutMix~\cite{Yun2019cutmix}, ResizeMix~\cite{Qin2020resizemix}, GridMix~\cite{Baek2021gridmix}, and RICAP~\cite{takahashi2019ricap} implement hand-crafted spatial cutting and pasting policies. FMix~\cite{Harris2020fmix} utilizes Fourier-guided cutting masks to blend frequency domain representations. To further optimize the mixing regions, SaliencyMix~\cite{Uddin2020saliencymix} and GuidedMixup~\cite{kang2023guidedmixup} apply saliency-guided and attention-guided cutting, respectively, ensuring the preservation of critical foreground textures. Some dynamic approaches like PuzzleMix~\cite{Kim2020puzzle} utilize optimal transport-based alignment before interpolation. Moreover, the latest automated blending mechanism, like StarMixup~\cite{jin2025starmixup}, performs end-to-end online-optimizable mixing, dynamically adapting the blending policy to the current training state of the model.

\textbf{Label Enhancement Methods:}
Regarding label enhancement, AGVBench implements methods that directly manipulate the target probability distributions to mitigate model overconfidence. LabelSmooth~\cite{szegedy2016labelsmooth} and ConfidencePenalty~\cite{pereyra2017confidencepenalty} apply static, hand-crafted uniform distribution softening to the ground-truth vectors. To mitigate the impact of potentially noisy annotations and leverage self-guided learning, Bootstrapping~\cite{reed2014bootstrapp} dynamically updates the training targets by blending the original hard labels with current predictive distributions of the model. OnlineLabelSmooth~\cite{zhang2021onlinelabelsmooth} performs dynamic label updating based on the statistical distribution of historical predictions. Furthermore, DirichletLabelSmooth~\cite{cheng2021dirichletlabelsmoothLoss} exploits parameterized Dirichlet distributions to generate dynamic soft labels, adapting the regularization intensity through training optimization. These label-level strategies provide critical calibration when deep architectures fit highly complex biometric features.

\subsection{Backbones}
\label{sec:3.2}
To evaluate how augmentations influence backbone architectures, AGVBench incorporates two representative models: classic visual backbones and domain-specific vein recognition models. 
For the first category, we select models spanning both CNNs and ViTs. Specifically, we employ MobileNetv2 (Mobv2)~\cite{Mark2018mobilenetv2} as a representative of lightweight CNNs, ResNet18 (R18)~\cite{He2016deep} for standard residual learning, and ViT-small (ViT-S)~\cite{Dosovitskiy2021vit} along with Swin Transformer-Tiny (Swin-T)~\cite{iccv2021swin} to evaluate global and hierarchical attention mechanisms, respectively. 
In parallel, we evaluate domain-specific architectures tailored for vein pattern extraction, namely FVRASNet (FVN)~\cite{Yang2020fvrasnet}, AMPVNet (APN)~\cite{luo2024scutpv}, and StarLKNet-S (SLN-S)~\cite{jin2025starmixup}, which possess specialized priors for biometric recognition. 

\begin{figure}
    \centering
    \includegraphics[width=0.9\linewidth]{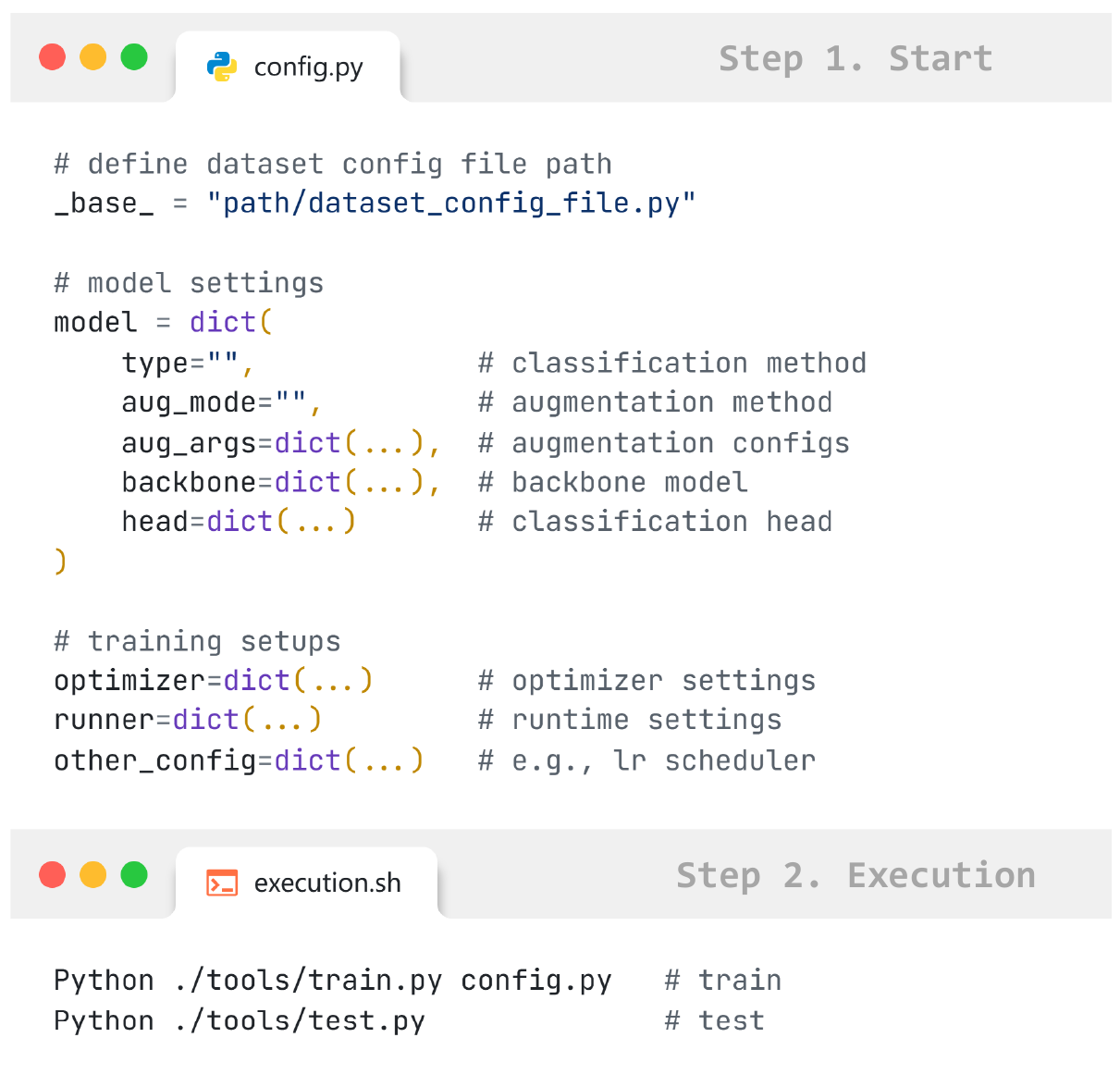}
    \vspace{-1.5em}
    \caption{Experimental pipeline in AGVBench codebase. The workflow is configuration-driven, where users define all experimental settings in a \textcolor{codemaroon}{\texttt{config.py}} file before running the standardized execution scripts.}
    \vspace{-1.em}
    \label{fig:exp_pipeline}
\end{figure}

\begin{table*}[t]
\caption{Top-1 Accuracy (\%)$\uparrow$, EER (\%)$\downarrow$, and TAR@FAR (T@R)=0.0001 (\%)$\uparrow$ of various augmentations across different models on VERA220.}
\centering
\setlength{\tabcolsep}{2.mm}
\resizebox{1.\linewidth}{!}{
    \begin{tabular}{lccccccccccccccc}
    \toprule
    \multirow{2}{*}{\textbf{VERA220}} 
        & \multicolumn{3}{c}{\textbf{R18}} 
        & \multicolumn{3}{c}{\textbf{Mobv2}} 
        & \multicolumn{3}{c}{\textbf{FVN}} 
        & \multicolumn{3}{c}{\textbf{APN}} 
        & \multicolumn{3}{c}{\textbf{SLK-S}} \\
    \cmidrule(lr){2-4} 
    \cmidrule(lr){5-7} 
    \cmidrule(lr){8-10} 
    \cmidrule(lr){11-13} 
    \cmidrule(lr){14-16}
        & Acc & EER & T@F 
        & Acc & EER & T@F 
        & Acc & EER & T@F 
        & Acc & EER & T@F 
        & Acc & EER & T@F \\
    \midrule
    Vanilla              & 71.45 & 5.20 & 51.00 & 71.55 & 5.94 & 47.18 & 59.73 & 6.38 & 35.91 & 74.00 & 3.95 & 57.18 & 70.27 & 5.46 & 44.55  \\
    \midrule
    Flip                 & \tg{67.55} & \tg{6.27} & \tg{42.55} & \tg{67.27} & \tg{6.40} & \tg{43.82} & \tg{51.09} & \tg{10.98} & \tg{19.36} & \tg{66.73} & \tg{4.71} & \tg{46.82} & \tg{63.45} & \tg{7.31} & \tg{35.27}  \\
    Rotate               & \tg{63.27} & \tg{7.54} & \tg{37.73} & \tg{63.00} & \tg{8.77} & \tg{43.36} & \tg{50.55} & \tg{10.90} & \tg{19.45} & \tg{66.55} & \tg{7.49} & \tg{37.27} & \tg{60.36} & \tg{9.34} & \tg{37.45}  \\
    Translation          & \tg{64.00} & \tg{7.33} & \tg{42.82} & \tg{65.45} & \tg{8.21} & \tg{41.45} & \tg{51.00} & \tg{11.07} & \tg{22.27} & \tg{66.64} & \tg{5.93} & \tg{44.27} & \tg{64.00} & \tg{8.28} & \tg{38.82}  \\
    Noise                & \tg{68.82} & \tg{5.39} & \tg{44.64} & \tg{69.73} & 5.82 & 49.64 & \tg{39.64} & \tg{14.27} & \tg{7.00} & \tg{70.00} & \tg{4.44} & \tg{50.45} & \tg{68.45} & \tg{6.46} & \tg{39.45}  \\
    Cutout~\cite{devries2017cutout}               & \tg{70.55} & \tg{5.33} & \tg{50.45} & \tg{67.55} & \tg{6.35} & 48.91 & \tg{52.82} & \tg{10.06} & \tg{25.73} & \tg{72.73} & \tg{5.45} & \tg{54.27} & \tg{65.36} & \tg{6.66} & 47.18  \\
    GridMask~\cite{chen2020gridmask}             & \tg{66.00} & \tg{8.15} & \tg{40.55} & \tg{68.45} & \tg{6.45} & \tg{44.82} & \tg{37.09} & \tg{14.55} & \tg{10.00} & \tg{60.36} & \tg{8.26} & \tg{41.91} & \tg{57.55} & \tg{10.56} & \tg{29.73}  \\
    RandomErasing~\cite{zhong2020randome}        & \tg{62.73} & \tg{7.20} & \tg{44.00} & \tg{69.82} & \tg{6.17} & 47.73 & \tg{48.73} & \tg{12.07} & \tg{18.09} & 74.45 & 3.66 & \tg{54.45} & \tg{68.18} & \tg{6.25} & \tg{42.91}  \\
    RandomQuant~\cite{wu2023rq}          & \ul{81.91} & \ul{1.91} & 65.00 & \ul{82.09} & 2.98 & 62.55 & \tg{40.36} & \tg{9.63} & \tg{8.18} & \ul{81.27} & 3.39 & \ul{65.55} & 78.73 & 2.72 & 61.45  \\
    AutoAugment~\cite{cubuk2019autoaugment}          & 80.82 & 2.55 & \ul{65.09} & 76.91 & 3.33 & 59.73 & \tg{55.55} & \tg{7.27} & \tg{24.45} & 80.00 & \ul{2.87} & 64.45 & 79.45 & 2.83 & 62.91  \\
    RandAugment~\cite{cubuk2020randaugment}          & 74.91 & 3.47 & 56.73 & \ul{82.09} & \ul{2.66} & \ul{68.91} & \tg{36.18} & \tg{11.69} & \tg{8.91} & 76.64 & 3.08 & 60.91 & \ul{84.18} & \tb{1.94} & \ul{65.64}  \\
    KeepAugment~\cite{gong2021keepaugment}          & 74.36 & 5.10 & 52.73 & \tg{70.82} & \tg{5.96} & \tg{47.00} & \tb{60.45} & \tg{\tb{6.45}} & \tg{\tb{32.36}} & 75.82 & \tg{4.18} & \tg{51.09} & \tg{68.27} & 4.93 & \tg{42.55}  \\
    TrivialAugment~\cite{muller2021trivialaugment}       & \tb{87.36} & \tb{1.52} & \tb{74.82} & \tb{82.45} & \tb{1.99} & \tb{70.45} & \tg{\ul{57.64}} & \tg{\ul{7.26}} & \tg{20.45} & \tb{87.64} & \tb{1.81} & \tb{80.64} & \tb{84.45} & \ul{2.45} & \tb{70.36}  \\
    TeachAugment~\cite{suzuki2022teachaugment}         & \tg{69.27} & 4.84 & \tg{45.82} & \tg{59.82} & \tg{9.11} & \tg{36.64} & \tg{51.27} & \tg{9.71} & \tg{22.64} & \tg{58.55} & \tg{11.46} & \tg{53.00} & \tg{65.18} & \tg{7.78} & \tg{29.64}  \\
    SoftAugment~\cite{liu2023softaug}          & \tg{67.18} & \tg{6.58} & \tg{45.64} & \tg{69.82} & 5.77 & \tg{47.00} & \tg{56.45} & \tg{8.48} & \tg{\ul{27.27}} & 75.55 & 3.91 & \tg{56.36} & \tg{69.27} & \tg{5.49} & 45.00  \\
    YOCO~\cite{han2022yoco}                 & \tg{64.73} & \tg{8.52} & \tg{27.73} & \tg{68.00} & \tg{10.07} & \tg{33.82} & \tg{43.73} & \tg{14.72} & \tg{11.91} & 74.55 & \tg{7.13} & \tg{34.18} & \tg{63.55} & \tg{8.46} & \tg{34.45}  \\
    \midrule
    RICAP~\cite{takahashi2019ricap}                & 78.64 & 4.73 & 59.91 & \tg{70.73} & \tg{6.55} & \tg{46.45} & \tg{40.82} & \tg{11.15} & \tg{11.82} & \tg{71.82} & \tg{5.30} & \tg{51.27} & \tg{65.36} & \tg{8.54} & \tg{41.09}  \\
    MixUp~\cite{Zhang2018mixup}                & \ul{95.27} & \ul{0.91} & \ul{92.27} & \ul{95.55} & \ul{0.87} & \cb{\tb{93.73}} & \ul{80.73} & \ul{3.14} & \ul{59.64} & \ul{95.64} & \ul{0.74} & \cb{\tb{93.45}} & \cb{\tb{94.55}} & \cb{\tb{0.87}} & \cb{\tb{91.18}}  \\
    CutMix~\cite{Yun2019cutmix}               & 84.73 & 2.92 & 74.00 & 77.36 & 4.49 & 60.00 & \tg{51.64} & \tg{8.38} & \tg{20.64} & 81.18 & 2.74 & 69.82 & 76.36 & \tg{6.24} & 58.09  \\
    FMix~\cite{Harris2020fmix}                 & 81.64 & 3.36 & 64.36 & 75.55 & 4.65 & 58.45 & \tg{56.18} & \tg{7.89} & \tg{27.27} & 81.64 & 3.17 & 71.09 & 70.64 & \tg{7.10} & 51.73  \\
    GridMix~\cite{Baek2021gridmix}             & 78.09 & 3.87 & 61.82 & 75.09 & 4.70 & 57.55 & \tg{42.45} & \tg{11.37} & \tg{12.27} & \tg{73.36} & \tg{4.82} & \tg{52.27} & \tg{68.45} & \tg{8.83} & 48.64  \\
    ResizeMix~\cite{Qin2020resizemix}            & 83.00 & 3.56 & 70.27 & 75.82 & 4.89 & 63.09 & \tg{51.18} & \tg{9.62} & \tg{13.18} & 81.18 & 3.37 & 68.00 & 70.36 & \tg{7.19} & 48.27  \\
    SaliencyMix~\cite{Uddin2020saliencymix}          & 83.36 & 2.80 & 72.45 & 79.09 & 3.64 & 61.55 & \tg{55.09} & \tg{7.64} & \tg{25.91} & 82.18 & 3.12 & 72.00 & 73.45 & \tg{6.73} & 58.91  \\
    PuzzleMix~\cite{Kim2020puzzle}            & \cb{\tb{95.55}} & \cb{\tb{0.83}} & \cb{\tb{93.36}} & \cb{\tb{95.91}} & \cb{\tb{0.83}} & \ul{91.45} & 76.36 & 3.55 & 51.18 & 94.36 & 1.08 & 89.73 & 92.00 & 1.26 & 86.27  \\
    GuidedMixup~\cite{kang2023guidedmixup}          & \tg{66.09} & \tg{7.19} & \tg{44.45} & \tg{69.73} & \tg{6.54} & \tg{44.27} & $-$ & $-$ & $-$  & \tg{65.73} & \tg{29.64} & \tg{0.64} & \tg{69.27} & \tg{6.29} & \tg{44.36}  \\
    StarMixup~\cite{jin2025starmixup}            & 94.91 & 0.96 & \ul{92.27} & 92.64 & 1.11 & 87.55 & \cb{\tb{83.45}} & \cb{\tb{2.62}} & \cb{\tb{67.82}} & \cb{\tb{96.27}} & \cb{\tb{0.71}} & \ul{92.91} & \ul{93.09} & \ul{1.16} & \ul{89.73}  \\
    \midrule
    LabelSmoothing~\cite{szegedy2016labelsmooth}       & \ul{82.64} & \tb{2.54} & \tb{66.09} & \ul{75.91} & \ul{4.88} & \ul{55.27} & \tb{67.09} & \ul{5.35} & \tb{47.91} & \ul{89.91} & \ul{2.19} & \ul{76.00} & \ul{72.00} & \ul{5.28} & \ul{46.09}  \\
    OnlineLabelSmooth~\cite{zhang2021onlinelabelsmooth}    & 74.82 & 4.24 & 54.91 & \tg{70.18} & 5.08 & 50.00 & 63.36 & 5.91 & \tg{33.73} & 80.00 & 3.50 & 63.82 & \tb{73.55} & \tb{5.15} & \tb{48.91}  \\
    ConfidencePenalty~\cite{pereyra2017confidencepenalty}    & \tg{70.73} & 4.82 & 52.27 & \tg{70.82} & 5.72 & 49.91 & 61.18 & \tg{6.80} & \tg{34.09} & 74.64 & 3.36 & \tg{53.27} & \tg{66.45} & \tg{7.81} & \tg{39.00}  \\
    DirichletLabelSmooth~\cite{cheng2021dirichletlabelsmoothLoss} & \tb{83.82} & \ul{2.76} & \ul{64.82} & \tb{77.45} & \tb{4.26} & \tb{55.82} & \ul{66.64} & \tb{4.82} & \ul{44.82} & \tb{91.45} & \tb{2.11} & \tb{81.18} & 71.09 & \tg{5.57} & \tg{39.64}  \\
    Bootstrapping~\cite{reed2014bootstrapp}        & \tg{65.64} & \tg{6.63} & \tg{42.73} & \tg{52.55} & \tg{10.99} & \tg{24.55} & 61.55 & \tg{6.76} & \tg{33.36} & 76.45 & 3.59 & \tg{55.82} & \tg{43.00} & \tg{13.55} & \tg{12.36}  \\
    \bottomrule
    \end{tabular}
}
\vspace{-1.em}
\label{tab:vera220}
\end{table*}

\subsection{Datasets}
\label{sec:3.3}
AGVBench benchmarks 5 public mainstream datasets, including two vascular modalities. Specifically, three palm vein datasets, namely SCUT1100~\cite{luo2024scutpv}, TJU600~\cite{Zhang2018palmprint}, and VERA220~\cite{Tome2015vera}, and two finger vein datasets, namely FV-USM~\cite{asaari2014fvusm} and SDUMLA-HMT~\cite{yin2011sdumla}, are selected. These datasets exhibit significant changes in sensor hardware, environmental lighting, and image quality, providing a challenging scenario for testing the adaptive capabilities of different augmentation operators. 
The details of these vein datasets are given in Section~\ref{sec:4.1}.

\subsection{Evaluation Protocol}
\label{sec:3.4}
To comprehensively assess the evaluated augmentation strategies, we design a multi-dimensional protocol encompassing performance, robustness, and efficiency. For brevity, comprehensive mathematical formulations (e.g., ECE, $P_\text{aug}$, APEX paradigm), matching threshold mechanisms, and exact hyperparameter configurations are uniformly detailed in Appendix~\ref{sec:appendix_protocol}.

\textbf{Performance Metrics:}
To quantitatively evaluate recognition capabilities and feature robustness, we employ Top-1 Accuracy (Acc.), Equal Error Rate (EER), and True Acceptance Rate at FAR=$10^{-4}$ (TAR@FAR=0.0001). These metrics comprehensively reflect both basic classification power and threshold-dependent verification reliability under highly stringent security scenarios. 

\textbf{Robustness Evaluation:}
To assess reliability in real-world complex scenarios, AGVBench integrates a multi-dimensional robustness suite:
\begin{itemize}
    \item \textit{Calibration:} We utilize the Expected Calibration Error (ECE) to quantify model overconfidence, a prevalent issue in modern DNNs trained with intense augmentations. 
    \item \textit{Corruption:} Following ImageNet-C~\cite{hendrycks2019robustness}, we evaluate robustness against 19 out-of-distribution degradations. Given the extreme fragility of fine-grained vein topologies, we tailor the evaluation to lower severity levels (C1--C3) to prevent catastrophic baseline collapse.
    \item \textit{Adversarial Attacks:} We implement white-box attacks, including FGSM~\cite{goodfellow2014explainingfgsm} and PGD~\cite{madry2018towardspgd}, to assess vulnerability against malicious imperceptible perturbations.
    \item \textit{Occlusion:} We simulate sensor smudges by masking continuous regions (occlusion ratios from $0\%$ to $50\%$) to evaluate global structural fault tolerance.
\end{itemize}

\begin{table*}[h!]
\caption{Top-1 Accuracy (\%)$\uparrow$, EER (\%)$\downarrow$, and TAR@FAR (T@R)=0.0001 (\%)$\uparrow$ of various augmentations across different models on TJU600.}
\centering
\setlength{\tabcolsep}{1.mm}
\resizebox{1.\linewidth}{!}{
    \begin{tabular}{lccccccccccccccccccccccc}
    \toprule
    \multirow{2}{*}{\textbf{TJU600}}
        & \multicolumn{3}{c}{\textbf{R18}}
        & \multicolumn{3}{c}{\textbf{Mobv2}}
        & \multicolumn{3}{c}{\textbf{FVN}}
        & \multicolumn{3}{c}{\textbf{APN}}
        & \multicolumn{3}{c}{\textbf{SLK-S}}
        & \multicolumn{3}{c}{\textbf{ViT-S}}
        & \multicolumn{3}{c}{\textbf{Swin-T}} \\
    \cmidrule(lr){2-4}
    \cmidrule(lr){5-7}
    \cmidrule(lr){8-10}
    \cmidrule(lr){11-13}
    \cmidrule(lr){14-16}
    \cmidrule(lr){17-19}
    \cmidrule(lr){20-22}
        & Acc & EER & T@F
        & Acc & EER & T@F
        & Acc & EER & T@F
        & Acc & EER & T@F
        & Acc & EER & T@F
        & Acc & EER & T@F
        & Acc & EER & T@F \\
    \midrule
    Vanilla              & 85.55 & 1.72 & 81.23 & 80.33 & 2.68 & 74.80 & 64.33 & 5.31 & 48.83 & 75.02 & 3.76 & 66.91 & 76.95 & 4.78 & 57.05 & 61.77 & 5.73 & 46.76 & 76.55 & 2.83 & 66.06 \\
    \midrule
    Flip                 & \tg{79.50} & \tg{2.75} & \tg{73.55} & \tg{76.15} & \tg{3.65} & \tg{69.18} & \ul{66.83} & \tg{7.16} & \tg{45.85} & \tg{69.68} & \tg{4.91} & \tg{59.46} & \tg{73.80} & \tg{5.79} & \tg{49.81} & \tg{58.58} & \tg{6.48} & \tg{42.81} & \tg{71.90} & \tg{3.71} & \tg{61.90} \\
    Rotate               & \tg{82.75} & \tg{1.86} & \tg{77.83} & \tg{79.75} & \tg{3.06} & \tg{73.75} & \tg{59.58} & \tg{5.83} & \tg{46.26} & \tg{73.97} & \tg{3.98} & \tg{65.56} & \tg{75.30} & 4.09 & 58.61 & 64.85 & 4.98 & 51.78 & \tg{75.55} & \tg{2.91} & \tg{65.75} \\
    Translation          & 85.67 & \tg{1.76} & 81.73 & 81.52 & \tg{2.76} & 75.85 & \tg{63.57} & 4.95 & 50.58 & 78.07 & 3.58 & 71.78 & 77.30 & 3.76 & 58.11 & 65.87 & 4.85 & 52.66 & \tg{76.13} & 2.60 & 67.36 \\
    Noise                & \tg{85.17} & \tg{2.38} & \tg{77.61} & 81.28 & \tg{2.86} & 75.88 & \tg{59.90} & \tg{6.08} & \tg{44.13} & \tg{73.87} & \tg{3.91} & \tg{64.11} & 80.03 & 2.63 & 73.61 & \tg{59.70} & \tg{6.01} & \tg{44.91} & \tg{67.30} & \tg{4.61} & \tg{53.83} \\
    Cutout~\cite{devries2017cutout}               & 87.38 & \ul{1.35} & 84.35 & 81.38 & 2.46 & \tg{74.53} & 65.95 & \tb{4.38} & 53.03 & \tb{80.53} & \tb{2.53} & 74.45 & 78.00 & 2.98 & 69.75 & 67.88 & 3.85 & 56.13 & 79.93 & \ul{2.08} & 72.00 \\
    GridMask~\cite{chen2020gridmask}             & \tg{81.82} & \tg{2.22} & \tg{73.63} & \tg{77.73} & \tg{3.18} & \tg{69.58} & \tg{56.52} & \tg{6.28} & \tg{36.56} & \tg{69.45} & \tg{4.39} & \tg{57.73} & \tg{75.20} & 3.51 & 65.26 & \tg{61.68} & 4.64 & 47.75 & \tg{73.75} & \tg{3.00} & \tg{63.45} \\
    RandomErasing~\cite{zhong2020randome}        & 87.37 & 1.56 & 83.68 & 82.92 & \tg{2.69} & 77.53 & \tg{63.43} & 4.94 & 49.70 & 78.77 & \ul{2.91} & 72.11 & 78.23 & 4.21 & 58.70 & 64.53 & 4.61 & 52.05 & 77.53 & 2.58 & 68.55 \\
    RandomQuant~\cite{wu2023rq}          & \tg{82.82} & \tg{2.56} & \tg{78.16} & 80.33 & \tg{3.03} & \tg{74.68} & \tg{62.83} & \tg{5.96} & \tg{47.80} & \tg{70.58} & \tg{5.20} & \tg{61.35} & 83.82 & \ul{2.01} & \ul{78.85} & 68.70 & \ul{3.66} & 57.55 & \tg{56.32} & \tg{6.50} & \tg{40.85} \\
    AutoAugment~\cite{cubuk2019autoaugment}          & \ul{88.28} & 1.59 & \ul{85.23} & \ul{85.93} & \ul{2.00} & \ul{82.03} & \tg{62.93} & \tg{5.78} & \tg{48.06} & 77.17 & \tg{3.83} & 70.08 & \tb{86.43} & \tb{1.88} & \tb{83.55} & 63.33 & 5.25 & 48.05 & 80.82 & 2.12 & 74.05 \\
    RandAugment~\cite{cubuk2020randaugment}          & 87.78 & 1.50 & 85.00 & 82.85 & \tg{2.76} & 77.65 & \ul{66.83} & 4.77 & \ul{53.33} & \ul{80.12} & 3.12 & \ul{74.86} & \ul{86.27} & 2.03 & 76.71 & \ul{71.25} & \tb{3.58} & \ul{61.05} & 81.12 & 2.19 & 74.46 \\
    KeepAugment~\cite{gong2021keepaugment}          & \tg{85.17} & \tg{1.86} & \tg{80.73} & 81.88 & \tg{2.89} & 76.91 & \tg{63.92} & 5.16 & 50.35 & 75.80 & \tg{4.22} & 68.38 & 79.70 & 2.63 & 73.36 & 62.95 & 5.70 & 48.71 & 76.67 & 2.71 & 67.83 \\
    TrivialAugment~\cite{muller2021trivialaugment}       & \tb{89.23} & \tb{1.18} & \tb{87.78} & \tb{86.45} & \tb{1.93} & \tb{83.40} & \tb{69.23} & \ul{4.63} & \tb{56.65} & 79.97 & 3.00 & 74.55 & 84.97 & 2.70 & 73.73 & 70.40 & 4.01 & 59.98 & \tb{82.02} & 2.09 & \tb{76.21} \\
    TeachAugment~\cite{suzuki2022teachaugment}         & \tg{81.48} & \tg{2.84} & \tg{77.28} & \tg{75.20} & \tg{4.94} & \tg{69.75} & \tg{42.68} & \tg{9.56} & \tg{23.13} & \tg{60.28} & \tg{10.63} & \tg{51.95} & \tg{70.47} & \tg{6.28} & 63.25 & \tb{72.50} & 3.68 & \tb{62.43} & \ul{81.42} & \tb{1.85} & \ul{74.88} \\
    SoftAugment~\cite{liu2023softaug}          & \tg{84.83} & 1.52 & \tg{80.96} & 81.78 & \tg{2.70} & 75.98 & \tg{64.12} & 5.03 & 50.61 & 76.13 & 3.51 & 69.50 & 77.52 & 2.94 & 70.11 & 63.42 & 5.48 & 49.21 & \tg{76.53} & 2.60 & 66.70 \\
    YOCO~\cite{han2022yoco}                 & \tg{81.45} & \tg{2.70} & \tg{73.28} & 82.52 & 2.03 & 77.56 & \tg{62.48} & \tg{5.53} & \tg{48.26} & 80.10 & 3.09 & \tb{75.03} & 80.00 & 2.46 & 73.86 & \tg{61.50} & 5.48 & 49.90 & \tg{68.83} & \tg{3.68} & \tg{57.60} \\
    \midrule
    RICAP~\cite{takahashi2019ricap}                & 87.60 & 1.47 & 82.97 & 84.17 & 2.03 & 77.17 & \tg{63.85} & \tg{5.29} & \tg{48.40} & \tg{69.77} & \tg{5.01} & \tg{57.73} & 87.18 & 1.77 & 82.88 & \tg{53.13} & \tg{13.47} & \tg{41.80} & \tg{61.73} & \tg{6.87} & \tg{48.65} \\
    MixUp~\cite{Zhang2018mixup}                & 93.90 & 0.84 & \ul{92.51} & 93.10 & 1.04 & 90.98 & 78.32 & 3.36 & 67.70 & 91.80 & 1.43 & \ul{89.85} & 94.70 & 0.68 & 92.73 & \cb{\tb{81.12}} & \cb{\tb{2.68}} & \cb{\tb{74.40}} & \ul{90.87} & \ul{1.18} & \ul{88.21} \\
    CutMix~\cite{Yun2019cutmix}               & 92.97 & 0.79 & 91.16 & 91.98 & 1.00 & 89.43 & 70.15 & 4.42 & 58.41 & 85.68 & 2.24 & 80.75 & 91.20 & 0.99 & 88.55 & 68.23 & \tg{6.68} & 58.68 & \tg{75.95} & \tg{3.56} & 67.81 \\
    FMix~\cite{Harris2020fmix}                 & 93.13 & 0.86 & 90.01 & 89.30 & 1.33 & 85.25 & 70.48 & 4.44 & 54.95 & 82.42 & 2.66 & 76.10 & 87.82 & 1.46 & 83.45 & 65.40 & \tg{6.91} & 54.86 & 81.08 & 2.31 & 73.73 \\
    GridMix~\cite{Baek2021gridmix}              & 86.88 & 1.58 & 83.16 & 86.37 & 1.88 & 81.70 & \tg{44.47} & \tg{9.11} & \tg{24.03} & 80.33 & 2.71 & 75.75 & 84.58 & 2.13 & 80.45 & 66.93 & \tg{6.51} & 56.76 & 77.03 & \tg{3.50} & 67.51 \\
    ResizeMix~\cite{Qin2020resizemix}            & 92.85 & 0.79 & 90.96 & 91.38 & 0.95 & 88.76 & 70.60 & 3.88 & 58.58 & 84.67 & 2.38 & 78.53 & 90.57 & 0.96 & 88.48 & 64.28 & \tg{8.49} & 53.56 & \tg{75.07} & \tg{3.90} & \tg{65.03} \\
    SaliencyMix~\cite{Uddin2020saliencymix}          & 92.80 & 0.85 & 90.73 & 91.27 & 1.02 & 88.13 & 70.40 & 4.50 & 58.26 & 86.05 & 2.21 & 81.40 & 90.53 & 1.18 & 88.76 & 69.40 & \tg{6.01} & 61.21 & 78.77 & \tg{2.90} & 71.26 \\
    PuzzleMix~\cite{Kim2020puzzle}            & \cb{\tb{95.25}} & \cb{\tb{0.46}} & \cb{\tb{94.45}} & \cb{\tb{94.63}} & \cb{\tb{0.66}} & \cb{\tb{93.80}} & 77.97 & \cb{\tb{2.96}} & \ul{69.13} & 89.92 & 1.50 & 86.86 & \cb{\tb{96.02}} & \cb{\tb{0.40}} & \cb{\tb{95.45}} & \ul{79.17} & \ul{3.08} & \ul{72.26} & 87.12 & 1.46 & 83.21 \\
    GuidedMixup~\cite{kang2023guidedmixup}          & \tg{77.45} & \tg{4.58} & \tg{69.81} & \tg{74.98} & \tg{4.50} & \tg{67.90} & \tg{36.03} & \tg{14.55} & \tg{18.70} & \tg{70.78} & \tg{6.52} & \tg{61.87} & 80.65 & 2.82 & 74.62 & 65.97 & \tg{7.53} & 51.23 & \tg{71.43} & \tg{5.66} & \tg{57.71} \\
    StarMixup~\cite{jin2025starmixup}            & 93.68 & 0.74 & 91.66 & 92.90 & 0.98 & 91.01 & \cb{\tb{79.65}} & 3.18 & \cb{\tb{70.03}} & \cb{\tb{92.55}} & 1.19 & \cb{\tb{89.98}} & \ul{94.97} & 0.71 & 92.98 & 78.77 & 3.23 & 70.20 & \cb{\tb{90.98}} & \ul{1.18} & \cb{\tb{88.31}} \\
    \midrule
    LabelSmoothing~\cite{szegedy2016labelsmooth}       & \tb{94.88} & \ul{0.84} & \ul{93.71} & \ul{90.23} & \tb{1.58} & \ul{87.11} & \ul{75.43} & \ul{3.73} & \ul{63.51} & \ul{91.20} & \tb{1.63} & \ul{87.80} & \ul{89.33} & \ul{1.38} & \ul{84.75} & \ul{65.98} & \tb{4.64} & 51.56 & \ul{86.15} & \tb{1.51} & \ul{79.68} \\
    OnlineLabelSmooth~\cite{zhang2021onlinelabelsmooth}    & 87.55 & 1.38 & 84.63 & 82.85 & 2.25 & 78.56 & 68.02 & 4.48 & 55.66 & 79.75 & 2.96 & 73.80 & 84.67 & 2.09 & 77.50 & 65.78 & \ul{4.68} & \ul{51.61} & 81.50 & 2.33 & 73.10 \\
    ConfidencePenalty~\cite{pereyra2017confidencepenalty}    & 86.17 & \tg{1.90} & 81.48 & 80.45 & \tg{2.99} & \tg{73.85} & 65.95 & 4.83 & 52.21 & 76.23 & 3.73 & 67.41 & 79.65 & 2.76 & 71.83 & 62.78 & 5.48 & 48.20 & 77.20 & 2.81 & 67.81 \\
    DirichletLabelSmooth~\cite{cheng2021dirichletlabelsmoothLoss} & \ul{94.83} & \tb{0.72} & \tb{93.73} & \tb{91.08} & \ul{1.78} & \tb{87.68} & \tb{76.63} & \tb{3.66} & \tb{64.21} & \tb{91.42} & \ul{1.93} & \tb{87.95} & \tb{90.20} & \tb{1.28} & \tb{85.48} & \tb{66.18} & 4.71 & \tb{51.80} & \tb{86.50} & \ul{1.64} & \tb{79.70} \\
    Bootstrapping~\cite{reed2014bootstrapp}        & \tg{76.13} & \tg{3.11} & \tg{67.43} & \tg{64.23} & \tg{5.68} & \tg{53.20} & \tg{32.97} & \tg{11.34} & \tg{15.56} & 76.50 & \tg{3.94} & 68.76 & \tg{66.97} & \tg{4.83} & \tg{55.53} & 62.92 & 5.31 & 48.61 & 76.93 & 2.71 & 67.68 \\
    \bottomrule
    \end{tabular}
}
\vspace{-1.em}
\label{tab:tju600}
\end{table*}

\textbf{Efficiency Evaluation:}
An ideal augmentation should achieve a superior trade-off between performance gain and resource consumption. We quantify computational overhead using training time per epoch ($T_\text{train}$), peak memory footprint ($M_\text{peak}$), GFLOPs, and extra learnable parameter ratio ($P_\text{aug}$). 

To provide a holistic assessment, we introduce the \underline{A}ugmentation \underline{P}erformance and \underline{E}fficiency E\underline{x}cellence (\textbf{APEX}) rank. Based on the principle of Pareto Efficiency, APEX models evaluation as a multi-objective optimization problem. It identifies non-dominated solutions across the multi-dimensional objective space, thereby classifying methods based on their strategic location on the efficiency-performance frontier rather than an arbitrary weighted sum.

\subsection{Experimental Pipeline of AGVBench Codebase}
\label{sec:3.5}
As illustrated in Fig. \ref{fig:exp_pipeline}, the experimental workflow of AGVBench is structured to support the complete process of vein recognition evaluations. 
The pipeline is primarily controlled through configuration files located in \textcolor{codemaroon}{\texttt{.configs}} directory. 
In these files, users specify the target dataset, data augmentation methods, backbone architecture, and training hyperparameters. During initialization, the framework reads these configurations and instantiates the required components from \textcolor{codemaroon}{\texttt{.agvbench.models}}, which maintains a registry of backbones, classification heads, and loss functions. 
The data pipeline is also built automatically according to the specified dataset format. 
For the execution phase, \textcolor{codemaroon}{\texttt{.tools}} directory provides scripts for both single-node and distributed training. It also includes specific evaluation utilities to extract feature embeddings and calculate verification metrics such as the Equal Error Rate (EER). 
Finally, the repository is accompanied by basic documentation and baseline results to assist researchers in reproducing the benchmarks or integrating new algorithms.
\begin{table*}[t]
\caption{Top-1 Accuracy (\%)$\uparrow$, EER (\%)$\downarrow$, and TAR@FAR (T@F)=0.0001 (\%)$\uparrow$ of various augmentations across different models on SCUT1100.}
\centering
\setlength{\tabcolsep}{1.mm}
\resizebox{1.\linewidth}{!}{
    \begin{tabular}{lcccccccccccccccccccccc}
    \toprule
    \multirow{2}{*}{\textbf{SCUT1100}}
    & \multicolumn{3}{c}{\textbf{R18}}
    & \multicolumn{3}{c}{\textbf{Mobv2}}
    & \multicolumn{3}{c}{\textbf{FVN}}
    & \multicolumn{3}{c}{\textbf{APN}}
    & \multicolumn{3}{c}{\textbf{SLK-S}}
    & \multicolumn{3}{c}{\textbf{ViT-S}}
    & \multicolumn{3}{c}{\textbf{Swin-T}} \\
    \cmidrule(lr){2-4}
    \cmidrule(lr){5-7}
    \cmidrule(lr){8-10}
    \cmidrule(lr){11-13}
    \cmidrule(lr){14-16}
    \cmidrule(lr){17-19}
    \cmidrule(lr){20-22}
    & Acc & EER & T@F
    & Acc & EER & T@F
    & Acc & EER & T@F
    & Acc & EER & T@F
    & Acc & EER & T@F
    & Acc & EER & T@F
    & Acc & EER & T@F \\
    \midrule
    Vanilla              & 96.05 & 0.30 & 97.30 & 95.27 & 0.32 & 96.80 & 92.96 & 0.59 & 94.12 & 94.20 & 0.55 & 95.43 & 93.96 & 0.49 & 95.30 & 76.42 & 2.56 & 72.60 & 94.07 & 0.36 & 95.80 \\
    \midrule
    % \multicolumn{8}{c}{\textit{\textbf{Single Image Augmentation}}} \\
    % \midrule
    Flip                 & \tg{94.89} & \tg{0.41} & \tg{96.05} & \tg{93.44} & \tg{0.56} & \tg{94.85} & \tg{92.60} & \tg{0.66} & \tg{93.80} & \tg{92.07} & \tg{0.70} & \tg{92.85} & \tg{91.65} & \tg{0.78} & \tg{92.94} & \tg{67.55} & \tg{3.90} & \tg{60.90} & \tg{93.67} & \tg{0.43} & \tg{95.03} \\
    Rotate               & \tg{95.05} & \tg{0.42} & \tg{96.34} & \tg{94.53} & \tg{0.51} & \tg{95.89} & \tg{89.91} & \tg{0.99} & \tg{90.63} & \tg{93.07} & \tg{0.59} & \tg{94.16} & \tg{93.38} & \tg{0.56} & \tg{94.63} & \tg{74.29} & \tg{2.73} & \tg{69.87} & \tg{93.87} & \tg{0.45} & \tg{95.58} \\
    Translation          & \tg{95.60} & \tg{0.32} & \tg{97.05} & \tg{94.84} & \tg{0.46} & \tg{96.21} & \tg{91.45} & \tg{0.72} & \tg{92.40} & 94.53 & 0.54 & 95.85 & \tg{93.05} & \tg{0.63} & \tg{94.69} & \tg{75.96} & \tg{2.87} & \tg{71.85} & 95.02 & 0.34 & 96.16 \\
    Noise                & \tg{88.65} & \tg{0.96} & \tg{89.00} & \tg{94.62} & \tg{0.40} & \tg{96.07} & \tg{85.87} & \tg{1.40} & \tg{85.49} & \tg{92.49} & \tg{0.65} & \tg{93.61} & \tg{93.73} & 0.43 & \tg{95.21} & \tg{72.40} & \tg{3.54} & \tg{66.52} & \tg{90.56} & \tg{0.74} & \tg{91.41} \\
    Cutout~\cite{devries2017cutout}               & \tg{86.89} & \tg{1.05} & \tg{85.83} & \tg{80.67} & \tg{1.67} & \tg{77.41} & \tg{90.05} & \tg{0.85} & \tg{89.78} & \tg{91.02} & \tg{0.69} & \tg{91.70} & \tg{84.45} & \tg{1.27} & \tg{83.50} & 82.71 & 1.72 & 80.70 & 94.22 & \tg{0.41} & \tg{95.69} \\
    GridMask~\cite{chen2020gridmask}             & \tg{85.80} & \tg{1.42} & \tg{81.61} & \tg{84.13} & \tg{2.11} & \tg{76.40} & \tg{79.80} & \tg{2.16} & \tg{72.21} & \tg{81.35} & \tg{2.36} & \tg{68.14} & \tg{80.65} & \tg{2.09} & \tg{73.89} & \tg{71.07} & \tg{2.87} & \tg{64.47} & \tg{88.49} & \tg{0.92} & \tg{88.69} \\
    RandomErasing~\cite{zhong2020randome}        & \tg{96.00} & 0.30 & 97.34 & \tg{94.71} & \tg{0.39} & \tg{96.16} & \tg{91.35} & \tg{0.76} & \tg{92.10} & \ul{95.00} & \tb{0.40} & \ul{96.12} & \tg{93.87} & 0.48 & \tg{95.25} & \tg{75.64} & 2.54 & \tg{71.40} & 94.76 & 0.36 & 96.36 \\
    RandomQuant~\cite{wu2023rq}          & \tg{95.64} & \tg{0.40} & \tg{97.01} & 95.27 & \tg{0.41} & \tg{96.45} & \tg{84.49} & \tg{1.69} & \tg{82.94} & \tg{92.58} & \tg{0.71} & \tg{93.72} & 95.24 & 0.43 & 96.67 & 84.82 & 1.34 & 83.78 & \tg{92.25} & \tg{0.52} & \tg{93.29} \\
    AutoAugment~\cite{cubuk2019autoaugment}          & 96.95 & 0.25 & 98.25 & 95.89 & \tg{0.34} & 97.29 & \tg{92.93} & \tg{0.67} & 94.29 & \tg{93.35} & 0.54 & \tg{94.94} & 95.65 & 0.32 & 97.03 & 85.58 & 1.25 & 85.09 & 96.38 & 0.20 & 97.92 \\
    RandAugment~\cite{cubuk2020randaugment}          & \ul{97.36} & \ul{0.24} & \ul{98.47} & \tb{96.82} & \tb{0.23} & \ul{97.96} & \ul{94.78} & \ul{0.44} & \ul{96.41} & 94.78 & 0.47 & 96.09 & \tb{97.09} & \tb{0.23} & \tb{98.29} & \ul{87.71} & \tb{1.00} & \ul{87.92} & 96.29 & 0.20 & \ul{98.03} \\
    KeepAugment~\cite{gong2021keepaugment}          & \tg{95.89} & \tg{0.32} & \tg{97.12} & \tg{94.91} & \tg{0.44} & \tg{96.10} & \tg{92.89} & \tg{0.66} & \tg{93.92} & \tg{93.93} & \tg{0.67} & \tg{95.07} & 94.00 & 0.43 & 95.60 & 76.76 & 2.54 & 72.96 & 94.67 & 0.36 & 96.18 \\
    TrivialAugment~\cite{muller2021trivialaugment}       & \tb{97.82} & \tb{0.18} & \tb{98.83} & \ul{96.73} & \ul{0.27} & \tb{98.20} & \tb{95.11} & \tb{0.39} & \tb{96.56} & \tb{95.22} & \ul{0.43} & \tb{96.63} & \ul{96.20} & \ul{0.31} & \ul{97.60} & 86.89 & 1.10 & 86.72 & \tb{96.58} & \tb{0.17} & \tb{98.09} \\
    TeachAugment~\cite{suzuki2022teachaugment}         & $-$ & $-$ & $-$ & \tg{79.44} & \tg{2.01} & \tg{74.65} & $-$ & $-$ & $-$ & \tg{49.09} & \tg{6.98} & \tg{32.94} & $-$ & $-$ & $-$ & \tb{88.42} & \ul{1.05} & \tb{88.43} & \ul{96.45} & \ul{0.18} & 97.94 \\
    SoftAugment~\cite{liu2023softaug}          & \tg{95.42} & \tg{0.40} & \tg{96.98} & \tg{94.56} & \tg{0.47} & \tg{95.94} & \tg{92.31} & \tg{0.70} & \tg{93.49} & \tg{93.98} & 0.45 & 95.43 & 94.02 & \tg{0.50} & \tg{94.96} & \tg{75.36} & \tg{2.81} & \tg{70.50} & \tg{93.87} & \tg{0.45} & \tg{95.25} \\
    YOCO~\cite{han2022yoco}                 & \tg{92.15} & \tg{3.34} & \tg{50.03} & \tg{78.02} & \tg{2.63} & \tg{69.85} & \tg{82.02} & \tg{7.61} & \tg{42.63} & \tg{84.96} & \tg{8.14} & \tg{19.90} & \tg{74.60} & \tg{46.47} & \tg{0.00} & \tg{75.31} & 2.47 & \tg{71.45} & \tg{92.04} & \tg{0.56} & \tg{93.54} \\
    \midrule
    RICAP~\cite{takahashi2019ricap}                & 98.05 & 0.20 & 98.84 & 95.73 & \tg{0.45} & \tg{95.98} & \tg{85.62} & \tg{1.97} & \tg{81.20} & \tg{93.07} & \tg{0.85} & \tg{92.51} & 97.56 & 0.20 & 98.40 & 80.93 & 2.35 & 77.25 & \tg{93.13} & \tg{0.77} & \tg{93.73} \\
    MixUp~\cite{Zhang2018mixup}                & 99.09 & \cb{\tb{0.07}} & 99.63 & 98.98 & \ul{0.08} & 99.56 & \cb{\tb{97.22}} & \cb{\tb{0.25}} & \cb{\tb{97.98}} & 97.80 & \cb{\tb{0.23}} & 98.47 & \ul{99.02} & \cb{\tb{0.05}} & \cb{\tb{99.60}} & \ul{92.42} & \cb{\tb{0.74}} & 92.67 & \cb{\tb{98.87}} & \cb{\tb{0.07}} & \cb{\tb{99.63}} \\
    CutMix~\cite{Yun2019cutmix}               & 98.76 & 0.10 & 99.41 & 97.49 & 0.25 & 98.23 & 94.22 & 0.48 & 94.90 & 96.60 & 0.38 & 97.78 & 98.25 & 0.11 & 99.27 & 89.11 & 1.09 & 88.78 & 97.60 & 0.14 & 98.72 \\
    FMix~\cite{Harris2020fmix}                 & 98.49 & 0.10 & 99.38 & 96.65 & 0.29 & 97.76 & 95.20 & 0.48 & 96.20 & 95.49 & 0.45 & 96.43 & 97.16 & 0.18 & 98.29 & 82.78 & 1.78 & 80.98 & 97.47 & \ul{0.10} & 98.85 \\
    GridMix~\cite{Baek2021gridmix}              & 97.73 & 0.22 & 98.34 & 96.18 & \tg{0.34} & 97.23 & \tg{92.40} & \tg{0.77} & \tg{92.61} & \tg{93.51} & \tg{0.67} & \tg{94.58} & 97.24 & 0.21 & 98.09 & 85.47 & 1.52 & 84.07 & 95.91 & 0.27 & 97.43 \\
    ResizeMix~\cite{Qin2020resizemix}            & 98.45 & \ul{0.08} & 99.21 & 97.51 & 0.16 & 98.27 & 94.71 & 0.58 & 95.34 & 96.20 & 0.41 & 97.32 & 98.07 & 0.11 & 98.90 & 83.62 & 1.92 & 81.69 & 96.87 & 0.20 & 98.25 \\
    SaliencyMix~\cite{Uddin2020saliencymix}          & 98.60 & 0.09 & 99.41 & 97.82 & 0.18 & 98.69 & 94.75 & 0.56 & 95.54 & 96.58 & 0.31 & 97.50 & 98.33 & 0.09 & 99.21 & 89.78 & 0.96 & 89.98 & 97.49 & 0.12 & 98.69 \\
    PuzzleMix~\cite{Kim2020puzzle}            & 99.13 & 0.10 & 99.43 & 98.84 & 0.14 & 99.29 & 96.38 & 0.41 & 96.94 & 97.25 & 0.38 & 97.74 & 98.96 & 0.11 & 99.50 & 92.00 & 0.81 & 91.96 & 98.07 & 0.16 & 98.87 \\
    GuidedMixup~\cite{kang2023guidedmixup}          & \cb{\tb{99.35}} & \cb{\tb{0.07}} & \cb{\tb{99.75}} & \cb{\tb{99.15}} & \cb{\tb{0.06}} & \cb{\tb{99.73}} & 96.91 & \ul{0.26} & \ul{97.84} & \ul{97.91} & 0.29 & \ul{98.56} & \cb{\tb{99.05}} & \ul{0.09} & \ul{99.58} & \cb{\tb{92.45}} & \ul{0.76} & \cb{\tb{92.85}} & \tg{92.85} & \tg{0.47} & \tg{94.10}\\
    StarMixup~\cite{jin2025starmixup}            & \ul{99.25} & \cb{\tb{0.07}} & \ul{99.70} & \ul{99.11} & \ul{0.08} & \ul{99.61} & \ul{96.95} & 0.33 & 97.70 & \cb{\tb{98.00}} & \ul{0.26} & \cb{\tb{98.65}} & 98.93 & 0.12 & 99.45 & 91.38 & 0.89 & 91.94 & \ul{98.78} & \cb{\tb{0.07}} & \ul{99.49} \\
    \midrule
    LabelSmoothing~\cite{szegedy2016labelsmooth}       & \ul{98.35} & \ul{0.18} & \ul{98.81} & \ul{96.78} & \ul{0.31} & \tb{97.45} & \ul{96.38} & \tb{0.39} & \ul{96.90} & \ul{97.78} & \ul{0.30} & \ul{98.21} & \ul{96.27} & \tb{0.29} & \ul{97.25} & 77.45 & \tg{2.76} & 72.78 & \tb{97.16} & \ul{0.23} & \ul{97.50} \\
    OnlineLabelSmooth~\cite{zhang2021onlinelabelsmooth}    & 97.29 & \ul{0.18} & 98.36 & 95.85 & \tb{0.27} & 97.16 & 95.33 & \ul{0.41} & 96.63 & 96.02 & 0.36 & 97.00 & 95.75 & \tb{0.29} & 96.81 & \tb{78.04} & \tb{2.05} & \tb{74.50} & 95.98 & 0.25 & 97.45 \\
    ConfidencePenalty~\cite{pereyra2017confidencepenalty}    & 96.05 & \tg{0.32} & 97.40 & \tg{94.69} & \tg{0.48} & \tg{96.16} & \tg{92.65} & \tg{0.63} & \tg{93.54} & \tg{94.04} & 0.48 & 95.60 & \tg{93.95} & \tg{0.53} & \tg{95.25} & \tg{76.15} & \tg{2.70} & \tg{72.43} & 94.65 & \tg{0.38} & 96.16 \\
    DirichletLabelSmooth~\cite{cheng2021dirichletlabelsmoothLoss} & \tb{98.45} & \tb{0.16} & \tb{98.94} & \tb{96.85} & 0.32 & \ul{97.41} & \tb{96.56} & 0.45 & \tb{97.05} & \tb{97.84} & \tb{0.29} & \tb{98.23} & \tb{96.95} & \ul{0.32} & \tb{97.60} & \ul{77.78} & \tg{2.74} & \ul{73.18} & \ul{96.89} & \tb{0.19} & \tb{97.60} \\
    Bootstrapping~\cite{reed2014bootstrapp}        & \tg{94.47} & \tg{0.51} & \tg{95.67} & \tg{85.16} & \tg{1.54} & \tg{84.38} & \tg{92.76} & \tg{0.64} & 94.18 & 94.56 & 0.41 & 95.80 & \tg{88.76} & \tg{1.03} & \tg{88.89} & 76.82 & \tg{\ul{2.67}} & 72.90 & 94.67 & 0.34 & 96.20 \\
    \bottomrule
    \end{tabular}
}
\vspace{-1.em}
\label{tab:scut1100}
\end{table*}

\section{Experiments and Results}
\label{sec:4}
In this section, we present the experimental evaluation conducted using the AGVBench framework. We first detail the experimental setup, including dataset configurations, implementation details, and training hyperparameters. Next, we report the baseline performance of the selected general and specialized backbone architectures on multiple public vein datasets. Subsequently, we systematically analyze the impact of various data augmentation strategies on model generalization and robustness. For all experiments, we reported accuracy using the median of the last 10 epochs and marked the best and second-best results in \tb{bold} and \ul{underlined}, respectively. Results worse than the Vanilla are marked in \tg{gray}. The best result in each column is marked with the \cb{blue background}.

\subsection{Experimental Setup}
\label{sec:4.1}
\textbf{Dataset Information:}
We evaluate the methods on 5 vein datasets: (1) For palm vein datasets, \textbf{VERA220}~\cite{Tome2015vera} comprises 110 subjects with 20 samples each (2,200 images), where 5 images per subject are used for training and 5 for testing. It is acquired in open environments with slight pose variations and ambient light interference. \textbf{TJU600}~\cite{Zhang2018palmprint} contains 300 subjects and 40 samples per subject (12,000 images), from which 10 images per subject are allocated for training and 10 for testing. This dataset is collected across two sessions in a semi-enclosed space with diverse hand postures and illumination conditions. \textbf{SCUT1100}~\cite{luo2024scutpv} includes 550 subjects with 20 samples per subject (11,000 images), partitioned into 5 training and 5 testing images per subject, captured in unconstrained dynamic scenarios with obvious out-of-plane rotation and grayscale variations. (2) For finger vein datasets, \textbf{FV-USM}~\cite{asaari2014fvusm} has 123 subjects and 12 samples per subject (1,476 images) from two collection sessions, evenly split into 6 for training and 6 for testing to evaluate temporal intra-class robustness. \textbf{SDUMLA-HMT}~\cite{yin2011sdumla} involves 106 subjects with 36 samples per subject (3,816 images), providing multi-finger vein data with rich variations in finger placement and orientation, where 4 images per finger class are used for training and 2 for testing. 

\textbf{Implementation Details:}
All experiments are implemented within the proposed AGVBench framework, which is built upon PyTorch and the MMCV ecosystem. During the data preprocessing stage, all input vein images are resized to a unified resolution of $224\times 224$ to ensure compatibility across all selected backbone architectures. The models are trained from scratch without relying on external pre-trained weights. All benchmarking experiments are conducted on a workstation equipped with a single NVIDIA A100 GPU. All the configuration parameters of augmentations are shown in Table~\ref{tab:hyper_config}.

\textbf{Training Settings:}
For ResNet18, MobileNetv2, and StarLKNet-S, we employ the SGD~\cite{Loshchilov2016sgdr} optimizer with an initial learning rate of 0.01. For FVRASNet, AMPVNet, ViT-S, and Swin-T, we utilize the AdamW~\cite{Ilya2019AdamW} optimizer with an initial learning rate of 0.001 and a weight decay rate of 0.01. Across all configurations, the learning rate is dynamically decayed using a cosine annealing schedule, gradually decreasing to a minimum learning rate of 0 by the end of the training process. All models are trained for 600 epochs with a batch size of 32.

\subsection{Results of Vein Recognition}
\label{sec:4.2}

\begin{table*}[t]
\caption{Top-1 Accuracy (\%)$\uparrow$, EER (\%)$\downarrow$, and TAR@FAR (T@F)=0.0001 (\%)$\uparrow$ of various augmentations across different models on FV-USM.}
\centering
\setlength{\tabcolsep}{2.mm}
\resizebox{1.\linewidth}{!}{
    \begin{tabular}{lccccccccccccccc}
    \toprule
    \multirow{2}{*}{\textbf{FV-USM}}
    & \multicolumn{3}{c}{\textbf{R18}}
    & \multicolumn{3}{c}{\textbf{Mobv2}}
    & \multicolumn{3}{c}{\textbf{FVN}}
    & \multicolumn{3}{c}{\textbf{APN}}
    & \multicolumn{3}{c}{\textbf{SLK-S}} \\
    \cmidrule(lr){2-4}
    \cmidrule(lr){5-7}
    \cmidrule(lr){8-10}
    \cmidrule(lr){11-13}
    \cmidrule(lr){14-16}
    & Acc & EER & T@F
    & Acc & EER & T@F
    & Acc & EER & T@F
    & Acc & EER & T@F
    & Acc & EER & T@F \\
    \midrule
    Vanilla              & 38.52 & 16.26 & 19.58 & 43.50 & 13.62 & 24.66 & 29.00 & 21.69 & 14.84 & 69.61 & 4.91 & 52.74 & 36.42 & 17.69 & 18.43 \\
    \midrule
    Flip                 & \tg{36.45} & 16.19 & 21.04 & \tg{41.16} & \tg{15.25} & \tg{22.49} & \tg{23.14} & \tg{24.80} & \tg{7.72} & \tg{61.96} & \tg{7.18} & \tg{41.26} & \tg{30.12} & \tg{20.09} & \tg{12.97} \\
    Rotate               & \tg{34.52} & \tg{16.87} & 20.19 & \tg{37.74} & \tg{16.43} & \tg{21.54} & \tg{25.43} & \tg{25.00} & \tg{10.40} & \tg{46.14} & \tg{13.46} & \tg{27.17} & \tg{30.45} & \tg{19.17} & \tg{17.07} \\
    Translation          & \tg{31.44} & \tg{19.41} & \tg{17.14} & \tg{36.35} & \tg{15.65} & \tg{21.88} & \tg{24.16} & \tg{22.39} & \tg{14.46} & \tg{55.10} & \tg{8.81} & \tg{35.40} & 36.48 & \tg{18.66} & 19.38 \\
    Noise                & 40.07 & 14.83 & \tg{18.12} & 48.75 & 12.57 & 27.41 & 31.03 & 20.18 & \tg{14.74} & \tg{66.26} & \tg{5.86} & \tg{48.81} & 41.87 & 14.66 & 22.36 \\
    Cutout~\cite{devries2017cutout}               & 46.85 & 14.39 & 27.74 & 45.36 & \tg{15.34} & 25.71 & 35.67 & 19.41 & 18.33 & \tg{56.50} & \tg{9.24} & \tg{38.31} & 44.85 & 15.74 & 29.17 \\
    GridMask~\cite{chen2020gridmask}             & \tg{27.45} & \tg{21.13} & \tg{14.33} & \tg{33.54} & \tg{17.94} & \tg{16.50} & \tg{17.50} & \tg{25.43} & \tg{11.79} & \tg{40.89} & \tg{16.32} & \tg{20.05} & \tg{28.32} & \tg{23.13} & \tg{11.79} \\
    RandomErasing~\cite{zhong2020randome}        & 39.74 & 16.22 & 21.82 & 46.38 & \tg{14.15} & 27.07 & 30.25 & 21.32 & 15.18 & \tg{67.11} & \tg{5.76} & \tg{52.71} & 38.92 & \tg{17.71} & 20.93 \\
    RandomQuant~\cite{wu2023rq}          & 80.35 & 3.22 & 69.82 & 82.59 & 2.87 & 73.04 & \ul{60.30} & 6.67 & 41.67 & 83.74 & 2.65 & 74.86 & 73.34 & 3.97 & 58.13 \\
    AutoAugment~\cite{cubuk2019autoaugment}          & 88.45 & 1.84 & 85.30 & 85.74 & 2.04 & 80.93 & 53.79 & 8.86 & 33.91 & 92.14 & 1.39 & 90.89 & 83.67 & 2.48 & 77.64 \\
    RandAugment~\cite{cubuk2020randaugment}          & \cb{\tb{93.90}} & \cb{\tb{1.15}} & \cb{\tb{93.50}} & \cb{\tb{94.44}} & \cb{\tb{1.06}} & \cb{\tb{94.65}} & \cb{\tb{79.44}} & \tb{5.22} & \ul{67.34} & \cb{\tb{95.63}} & \cb{\tb{0.71}} & \cb{\tb{95.80}} & \cb{\tb{92.31}} & \ul{1.38} & \cb{\tb{91.50}} \\
    KeepAugment~\cite{gong2021keepaugment}          & 41.29 & 14.95 & 21.04 & 49.63 & 11.31 & 32.62 & 30.56 & 20.22 & 16.16 & \tg{67.34} & \tg{5.79} & \tg{52.00} & 38.86 & 16.54 & 19.00 \\
    TrivialAugment~\cite{muller2021trivialaugment}       & \ul{92.21} & \ul{1.26} & \ul{91.12} & \ul{92.51} & \ul{1.09} & \ul{91.33} & 78.42 & \ul{5.49} & \tb{65.31} & \ul{94.72} & \ul{0.92} & \ul{94.55} & \ul{89.80} & \cb{\tb{1.35}} & \ul{87.53} \\
    TeachAugment~\cite{suzuki2022teachaugment}         & \tg{25.76} & \tg{19.21} & \tg{12.06} & 58.42 & 9.83 & 40.65 & \tg{26.24} & 21.31 & \tg{9.28} & \tg{47.27} & \tg{15.45} & \tg{42.34} & \tg{28.35} & \tg{48.92} & \tg{0.00} \\
    SoftAugment~\cite{liu2023softaug}          & 39.30 & 14.16 & 23.17 & \tg{42.89} & 12.12 & \tg{23.88} & 29.17 & 19.31 & 15.62 & \tg{64.57} & \tg{6.10} & \tg{46.92} & 42.62 & 14.29 & 23.04 \\
    YOCO~\cite{han2022yoco}                 & 41.80 & 13.58 & 25.00 & \tg{42.62} & 10.60 & 27.51 & \tg{28.31} & \tg{24.83} & \tg{9.45} & \tg{66.06} & \tg{5.29} & \tg{47.80} & 37.13 & 14.53 & 21.04 \\
    \midrule
    RICAP~\cite{takahashi2019ricap}                & 58.81 & 10.03 & 41.87 & 61.75 & 8.67 & 46.99 & 31.06 & \tg{22.15} & \tg{9.38} & \tg{58.84} & \tg{10.33} & \tg{42.21} & 53.73 & 13.42 & 41.43 \\
    MixUp~\cite{Zhang2018mixup}                & \ul{86.75} & 3.58 & \tb{84.25} & \ul{87.30} & \ul{3.45} & \cb{\tb{85.23}} & \tb{74.59} & \cb{\tb{3.87}} & \cb{\tb{67.92}} & \ul{82.86} & 4.74 & \tb{80.18} & \ul{86.14} & 3.83 & \ul{83.03} \\
    CutMix~\cite{Yun2019cutmix}               & 61.42 & 10.57 & 49.93 & 59.76 & 9.72 & 47.43 & 38.21 & \tg{22.26} & 17.85 & \ul{82.86} & \tg{6.44} & 64.60 & 53.22 & 14.57 & 42.82 \\
    FMix~\cite{Harris2020fmix}                 & 51.29 & 14.03 & 38.72 & 53.01 & 10.26 & 38.52 & 36.25 & 21.14 & 18.22 & \tg{68.29} & \tg{7.24} & 53.18 & 46.27 & 16.91 & 31.57 \\
    GridMix~\cite{Baek2021gridmix}              & 42.04 & \tg{17.22} & 30.18 & \tg{42.95} & \tg{15.34} & 30.42 & \tg{28.26} & \tg{24.25} & \tg{14.40} & \tg{55.69} & \tg{9.69} & \tg{46.04} & 41.77 & 17.55 & 26.66 \\
    ResizeMix~\cite{Qin2020resizemix}            & 52.88 & 11.04 & 37.87 & 57.69 & 9.35 & 44.99 & 33.50 & 21.24 & 16.12 & 71.21 & \tg{5.30} & 57.05 & 47.59 & 12.60 & 29.88 \\
    SaliencyMix~\cite{Uddin2020saliencymix}          & 62.36 & 9.38 & 52.24 & 60.67 & 10.47 & 50.27 & 37.26 & \tg{23.21} & 16.60 & 77.20 & \ul{4.21} & 65.41 & 59.21 & 12.19 & 48.07 \\
    PuzzleMix~\cite{Kim2020puzzle}            & \tb{87.20} & \tb{3.36} & 82.38 & \tb{88.35} & \tb{2.99} & 84.15 & \ul{72.19} & \ul{4.13} & \ul{58.71} & \tb{84.42} & \tb{3.83} & \ul{79.23} & \tb{87.36} & \tb{3.25} & \tb{83.81} \\
    GuidedMixup~\cite{kang2023guidedmixup}          & \tg{29.35} & \tg{18.50} & 24.09 & \tg{40.72} & \tg{17.95} & \tg{22.12} & \tg{16.52} & \tg{24.06} & 15.65 & \tg{48.71} & \tg{16.26} & \tg{36.72} & \tg{28.66} & 13.37 & 30.49 \\
    StarMixup~\cite{jin2025starmixup}            & 85.98 & \ul{3.55} & \ul{82.76} & 86.89 & 3.62 & \ul{85.06} & 69.38 & 4.81 & 48.41 & 81.71 & 4.88 & 77.41 & 84.93 & \ul{3.42} & 80.18 \\
    \midrule
    LabelSmooth~\cite{szegedy2016labelsmooth}          & \tb{61.31} & \tb{7.99} & \tb{47.46} & \tb{54.13} & \tb{9.62} & \tb{38.28} & \ul{42.48} & \tb{15.89} & \tb{22.97} & \tb{93.97} & \tb{0.98} & \ul{88.14} & \ul{45.90} & \ul{12.66} & \ul{28.25} \\
    OnlineLabelSmooth~\cite{zhang2021onlinelabelsmooth}    & 46.99 & 12.16 & 26.69 & 45.12 & 13.55 & 28.01 & 31.00 & 20.46 & \tg{13.62} & 81.71 & 4.00 & 56.47 & 42.01 & 16.23 & 23.07 \\
    ConfidencePenalty~\cite{pereyra2017confidencepenalty}    & 44.99 & 13.59 & 25.27 & 45.49 & 12.94 & 30.08 & 30.79 & 20.59 & \tg{14.40} & \tg{68.12} & \tg{5.45} & \tg{51.29} & 42.01 & 14.93 & 21.75 \\
    DirichletLabelSmooth~\cite{cheng2021dirichletlabelsmoothLoss} & \ul{59.38} & \ul{9.31} & \ul{44.58} & \ul{51.22} & \ul{10.61} & \ul{35.06} & \tb{43.60} & \ul{16.90} & \ul{22.49} & \ul{93.60} & \ul{1.06} & \tb{88.45} & \tb{52.64} & \tb{10.94} & \tb{33.71} \\
    Bootstrapping~\cite{reed2014bootstrapp}        & \tg{28.36} & \tg{20.96} & \tg{14.70} & \tg{25.10} & \tg{18.60} & \tg{15.65} & \tg{17.52} & \tg{22.70} & \tg{12.40} & \tg{68.12} & \tg{5.76} & \tg{51.76} & \tg{26.54} & \tg{19.10} & \tg{17.14} \\
    \bottomrule
    \end{tabular}
}
\label{tab:fv_usm}
\end{table*}
\begin{table*}[t]
\caption{Top-1 Accuracy (\%)$\uparrow$, EER (\%)$\downarrow$, and TAR@FAR (T@F)=0.0001 (\%)$\uparrow$ of various augmentations across different models on SDUMLA-HMT.}
\centering
\setlength{\tabcolsep}{1.0mm}
\resizebox{1.\linewidth}{!}{
    \begin{tabular}{lcccccccccccccccccccccc}
    \toprule
    \multirow{2}{*}{\textbf{SDUMLA-HMT}}
    & \multicolumn{3}{c}{\textbf{R18}}
    & \multicolumn{3}{c}{\textbf{Mobv2}}
    & \multicolumn{3}{c}{\textbf{FVN}}
    & \multicolumn{3}{c}{\textbf{APN}}
    & \multicolumn{3}{c}{\textbf{SLK-S}}
    & \multicolumn{3}{c}{\textbf{ViT-S}}
    & \multicolumn{3}{c}{\textbf{Swin-T}} \\
    \cmidrule(lr){2-4}
    \cmidrule(lr){5-7}
    \cmidrule(lr){8-10}
    \cmidrule(lr){11-13}
    \cmidrule(lr){14-16}
    \cmidrule(lr){17-19}
    \cmidrule(lr){20-22}
    & Acc & EER & T@F
    & Acc & EER & T@F
    & Acc & EER & T@F
    & Acc & EER & T@F
    & Acc & EER & T@F
    & Acc & EER & T@F
    & Acc & EER & T@F \\
    \midrule
    Vanilla              & 84.51 & 1.90 & 80.03 & 82.94 & 2.20 & 78.77 & 86.71 & 1.49 & 83.96 & 93.55 & 0.56 & 93.79 & 82.15 & 1.80 & 77.52 & 72.17 & 7.49 & 66.12 & 82.39 & 2.37 & 76.49 \\
    \midrule
    Flip                 & \tg{79.09} & \tg{2.56} & \tg{74.84} & \tg{77.52} & \tg{2.60} & \tg{70.83} & \tg{84.59} & \tg{1.64} & \tg{80.58} & \tg{87.89} & \tg{1.42} & \tg{84.67} & \tg{81.84} & \tg{2.10} & \tg{76.42} & \tg{66.90} & \tg{9.46} & \tg{58.41} & \tg{80.11} & \tg{3.07} & \tg{74.61} \\
    Rotate               & \ul{91.19} & 1.00 & 90.25 & 90.88 & 0.87 & 90.17 & \tg{84.98} & \tg{2.05} & \tg{82.08} & \tg{93.40} & \tg{0.64} & \tg{93.47} & 92.69 & 0.94 & 92.53 & 83.49 & 2.47 & 80.58 & 90.33 & 0.95 & 89.23 \\
    Translation          & 90.17 & 0.85 & 89.23 & 89.94 & 0.96 & 88.68 & 88.44 & 1.25 & 86.32 & 94.10 & 0.55 & 94.18 & 92.69 & 0.60 & 91.90 & 86.16 & 1.98 & 82.47 & 91.75 & 0.85 & 90.64 \\
    Noise                & \tg{80.82} & \tg{2.38} & \tg{77.20} & \tg{82.15} & 2.19 & \tg{77.04} & \tg{85.93} & \tg{1.66} & \tg{82.47} & \tg{91.59} & \tg{0.77} & \tg{91.04} & 91.43 & 0.95 & 91.04 & \tg{69.89} & 6.31 & \tg{63.68} & 82.78 & \tg{2.99} & 77.28 \\
    Cutout~\cite{devries2017cutout}               & 89.39 & 0.87 & 87.97 & 89.70 & \tb{0.60} & 88.44 & 90.72 & 1.16 & 89.78 & 93.95 & \tg{0.58} & 93.79 & 90.09 & 0.70 & 89.31 & 74.69 & 3.84 & 68.08 & 83.96 & 1.63 & 79.48 \\
    GridMask~\cite{chen2020gridmask}             & 86.64 & 1.30 & 83.02 & 89.78 & 0.85 & 88.52 & 87.66 & \tg{1.51} & 84.59 & \tg{92.69} & \tg{0.70} & \tg{92.53} & 90.33 & 0.72 & 88.60 & 79.48 & 2.16 & 72.56 & \tg{81.45} & \tg{2.99} & \tg{76.18} \\
    RandomErasing~\cite{zhong2020randome}        & 85.61 & 1.56 & 82.39 & 87.81 & 1.27 & 85.93 & 87.50 & 1.24 & 85.06 & 94.10 & \tg{0.64} & 94.42 & 92.45 & 0.79 & 92.53 & 72.25 & 4.78 & \tg{64.07} & 84.83 & 1.65 & 81.13 \\
    RandomQuant~\cite{wu2023rq}          & \tg{77.83} & \tg{3.00} & \tg{71.62} & \tg{75.00} & \tg{3.22} & \tg{69.81} & \tg{82.55} & \tg{1.58} & \tg{78.62} & \tg{89.86} & \tg{1.40} & \tg{88.44} & 88.52 & 1.33 & 85.46 & \tg{59.98} & \tg{8.49} & \tg{54.17} & \tg{75.94} & \tg{3.13} & \tg{70.52} \\
    AutoAugment~\cite{cubuk2019autoaugment}          & \tg{79.09} & \tg{2.77} & \tg{74.61} & 90.25 & 0.83 & 89.15 & 92.06 & 0.56 & 91.51 & \ul{96.62} & 0.31 & \ul{97.33} & 87.34 & 0.93 & 83.18 & 81.53 & 1.66 & 75.71 & 87.26 & 0.79 & 84.43 \\
    RandAugment~\cite{cubuk2020randaugment}          & 89.70 & \ul{0.71} & 88.84 & 90.17 & 0.86 & 88.36 & \cb{\tb{94.18}} & \cb{\tb{0.32}} & \cb{\tb{94.97}} & 96.38 & \ul{0.24} & \ul{97.33} & \ul{94.18} & \ul{0.56} & \ul{94.97} & 87.81 & 1.09 & 85.53 & \ul{93.87} & \ul{0.47} & 94.10 \\
    KeepAugment~\cite{gong2021keepaugment}          & \tg{84.28} & \tg{2.05} & \tg{79.32} & \tg{81.53} & 1.90 & \tg{77.75} & \tg{86.64} & 1.41 & \tg{83.02} & \tg{92.37} & \tg{0.93} & \tg{91.59} & 88.44 & 1.06 & 85.06 & 75.24 & 6.28 & 69.81 & \tg{82.08} & 2.18 & 77.52 \\
    TrivialAugment~\cite{muller2021trivialaugment}       & 89.78 & 1.00 & 87.89 & \tb{93.95} & \ul{0.70} & \tb{94.18} & \ul{93.63} & \ul{0.41} & \ul{93.71} & \cb{\tb{97.56}} & \cb{\tb{0.22}} & \cb{\tb{98.58}} & \cb{\tb{95.36}} & \tb{0.39} & \cb{\tb{95.75}} & 87.89 & \ul{0.88} & 85.22 & \tb{94.03} & 0.63 & \tb{94.42} \\
    TeachAugment~\cite{suzuki2022teachaugment}         & \tg{32.00} & \tg{11.71} & \tg{21.46} & \tg{38.29} & \tg{7.06} & \tg{22.64} & $-$ & $-$ & $-$ & \tg{53.30} & \tg{5.98} & \tg{46.30} & $-$ & $-$ & $-$ & \tb{93.55} & \cb{\tb{0.48}} & \tb{93.47} & 93.63 & \cb{\tb{0.32}} & \ul{94.18} \\
    SoftAugment~\cite{liu2023softaug}          & \ul{91.19} & 0.72 & \ul{90.80} & \ul{92.37} & 0.77 & \ul{92.53} & 87.50 & 1.34 & 85.46 & 95.13 & 0.54 & 95.44 & 92.22 & 0.79 & 92.14 & 87.81 & 2.27 & 85.06 & 90.09 & 1.01 & 88.99 \\
    YOCO~\cite{han2022yoco}                 & \tb{92.92} & \tb{0.39} & \tb{92.61} & 91.27 & 1.18 & \tg{76.73} & 91.43 & 1.03 & 87.19 & 95.99 & 0.38 & 95.44 & 90.49 & 1.28 & \tg{75.31} & \ul{89.62} & 1.30 & \ul{87.97} & 92.14 & 0.68 & 92.22 \\
    \midrule
    RICAP~\cite{takahashi2019ricap}                & \cb{\tb{98.66}} & \cb{\tb{0.10}} & \cb{\tb{99.53}} & \cb{\tb{97.56}} & \cb{\tb{0.11}} & \cb{\tb{97.88}} & 92.77 & 0.81 & 92.06 & \tg{91.51} & \tg{1.18} & \tg{90.72} & $-$ & $-$ & $-$ & \cb{\tb{94.26}} & \tb{0.70} & \cb{\tb{93.71}} & \cb{\tb{95.91}} & \tb{0.40} & \cb{\tb{96.70}} \\
    MixUp~\cite{Zhang2018mixup}                & \tg{79.87} & \tg{2.25} & \tg{74.29} & \tg{82.55} & 1.90 & \tg{77.99} & \ul{93.40} & \ul{0.63} & \tb{93.24} & \tg{91.43} & \tg{0.94} & \tg{91.27} & 87.81 & 1.33 & 86.08 & 77.59 & 2.98 & 72.17 & \tg{69.97} & \tg{5.03} & \tg{64.07} \\
    CutMix~\cite{Yun2019cutmix}               & 93.47 & 0.55 & 91.67 & 92.06 & 0.66 & 92.14 & 91.67 & 0.69 & 91.12 & \ul{95.05} & \tg{0.63} & \ul{94.81} & 93.55 & \ul{0.41} & \ul{92.92} & 84.67 & 1.87 & 78.85 & 88.68 & 1.22 & 86.64 \\
    FMix~\cite{Harris2020fmix}                 & 90.72 & 1.10 & 87.74 & 88.29 & 1.48 & 85.93 & 89.39 & 1.18 & 87.03 & 94.50 & \tg{0.72} & \tg{93.00} & 90.49 & 0.94 & 88.29 & 79.32 & 2.44 & 74.06 & 85.77 & 1.97 & 80.74 \\
    GridMix~\cite{Baek2021gridmix}              & 90.02 & 1.12 & 88.68 & 92.61 & 0.86 & 91.59 & 91.90 & 0.88 & 90.72 & 93.95 & \tg{0.85} & \tg{93.71} & 92.85 & \ul{0.41} & 92.77 & 86.87 & 1.72 & 84.04 & 89.15 & 1.41 & 87.74 \\
    ResizeMix~\cite{Qin2020resizemix}            & \ul{95.83} & \ul{0.23} & \ul{96.70} & \ul{93.87} & \ul{0.41} & \ul{93.79} & 90.72 & 0.80 & 89.86 & \tb{95.75} & \ul{0.48} & \tb{96.46} & \tb{94.65} & \cb{\tb{0.31}} & \tb{95.28} & 87.19 & 1.64 & \ul{84.98} & \ul{91.75} & 1.10 & \ul{89.70} \\
    SaliencyMix~\cite{Uddin2020saliencymix}          & 90.09 & 0.77 & 88.92 & 89.07 & 0.83 & 87.11 & 91.59 & \tb{0.57} & 89.94 & \tg{93.00} & \ul{0.48} & \tg{92.85} & 91.12 & 1.00 & 89.94 & 80.35 & 2.23 & 75.47 & 86.56 & 1.41 & 83.49 \\
    PuzzleMix~\cite{Kim2020puzzle}            & 88.29 & 1.22 & 85.93 & 93.32 & 0.68 & 91.67 & \tb{93.63} & 0.64 & \ul{93.00} & 94.34 & \tg{0.71} & \tg{93.32} & \ul{93.63} & 0.55 & 92.37 & 84.67 & 1.80 & 79.01 & 87.19 & 1.72 & 84.28 \\
    GuidedMixup~\cite{kang2023guidedmixup}          & 88.29 & 0.88 & 86.56 & 89.31 & 0.85 & 87.81 & 92.85 & 0.60 & 92.14 & \tg{91.19} & \tg{0.95} & \tg{90.88} & 89.78 & 0.76 & 88.76 & \ul{87.42} & \ul{1.32} & \ul{84.98} & 87.81 & \ul{0.91} & 85.77 \\
    StarMixup~\cite{jin2025starmixup}            & \tg{79.40} & \tg{1.96} & \tg{75.71} & \tg{81.76} & 1.96 & \tg{77.44} & 91.67 & 0.70 & 91.12 & \tg{89.54} & \tg{1.25} & \tg{88.60} & 84.36 & \tg{2.11} & 80.03 & 78.14 & 2.90 & 73.66 & \tg{67.69} & \tg{5.73} & \tg{61.40} \\
    \midrule
    LabelSmoothing~\cite{szegedy2016labelsmooth}       & \tb{87.03} & \ul{1.79} & \tb{84.51} & \ul{84.83} & \tb{1.67} & \ul{80.90} & \ul{89.78} & \ul{1.02} & \ul{87.03} & \ul{97.09} & \ul{0.25} & \tb{97.72} & \ul{87.11} & \tb{1.01} & 84.28 & 73.98 & \ul{5.81} & 66.75 & \tb{84.12} & \tg{\ul{2.44}} & \tb{78.14} \\
    OnlineLabelSmooth~\cite{zhang2021onlinelabelsmooth}    & 84.75 & \tb{1.72} & 81.76 & \tg{81.13} & \tg{2.44} & \tg{76.57} & 87.26 & 1.25 & 84.28 & \tg{93.40} & 0.46 & \tg{93.08} & 84.36 & \tg{1.88} & 80.50 & \ul{74.14} & 7.24 & \ul{68.24} & \ul{82.70} & \tb{1.63} & 77.36 \\
    ConfidencePenalty~\cite{pereyra2017confidencepenalty}    & \tg{83.57} & \tg{2.05} & \tg{79.25} & \tg{78.22} & \tg{2.65} & \tg{71.93} & 87.03 & 1.33 & 84.12 & 93.87 & \tg{0.91} & 94.10 & \tb{87.89} & \ul{1.58} & \tb{85.14} & \tg{71.54} & 5.96 & \tg{64.70} & \tg{80.50} & \tg{3.28} & 76.57 \\
    DirichletLabelSmooth~\cite{cheng2021dirichletlabelsmoothLoss} & \ul{86.56} & \tg{1.97} & \ul{83.96} & \tb{85.53} & \ul{1.80} & \tb{81.13} & \tb{90.80} & \tb{0.93} & \tb{87.74} & \tb{97.25} & \tb{0.24} & \ul{97.25} & \tb{87.89} & 1.71 & \ul{84.67} & 73.19 & 5.98 & \tg{65.96} & \tg{82.31} & \tg{2.51} & \ul{78.07} \\
    Bootstrapping~\cite{reed2014bootstrapp}        & \tg{82.31} & \tg{2.13} & \tg{77.04} & \tg{72.64} & \tg{4.03} & \tg{64.86} & \tg{78.46} & \tg{2.35} & \tg{62.74} & 93.95 & 0.54 & 93.95 & $-$ & $-$ & $-$ & \tb{74.69} & \tb{5.74} & \tb{69.50} & 82.39 & \tg{2.66} & 76.97 \\
    \bottomrule
    \end{tabular}
}
\label{tab:sdumla_hmt}
\end{table*}

\textbf{Accuracy:} 
Across the five evaluated datasets, multi-image augmentation methods (MixUp, PuzzleMix, StarMixup) consistently dominate Top-1 Accuracy, particularly on palm vein datasets. For instance, as shown in Table~\ref{tab:vera220}, MixUp and PuzzleMix on VERA220 using ResNet18 achieve 95.27\% and 95.55\% respectively, yielding substantial gains over the Vanilla baseline (71.45\%). However, augmentation efficacy exhibits strong modality dependence. On the finger vein dataset SDUMLA-HMT (Table~\ref{tab:sdumla_hmt}), MixUp degrades ResNet18 accuracy to 79.87\% (\textit{vs.} 84.51\% baseline), whereas automated policies like RandAugment excel 89.70\%. Conversely, basic single-image geometric transformations (Flip, Rotate, Translate) frequently degrade performance below the baseline across most datasets, confirming that aggressive spatial operations easily disrupt the delicate topological structure of vascular patterns.

\textbf{EER and TAR@FAR=0.0001:} 
Strict biometric verification metrics reveal even more pronounced performance gaps. Mixup augmentations drastically reduce the EER and elevate the True Acceptance Rate (TAR). On Table~\ref{tab:scut1100}, MixUp reduces the ResNet18 EER from 0.30\% to 0.07\% and boosts TAR@FAR=0.0001 from 97.30\% to 99.63\%, highlighting its superiority in refining inter-class boundaries even on already saturated large-scale datasets. Similarly, on Table~\ref{tab:vera220}, MixUp boosts TAR from 51.00\% to 92.27\%. These results demonstrate that rigorous threshold-dependent metrics like TAR@FAR underscore the necessity of advanced mixing strategies for high-security real-world deployments. Figure~\ref{fig:roc} shows the ROC curves of 5 vein datasets using the ResNet18 model, and the results are shown in Appendix~\ref{sec:appendix}.

\begin{figure*}[t]
    \centering
    \includegraphics[width=1.0\linewidth]{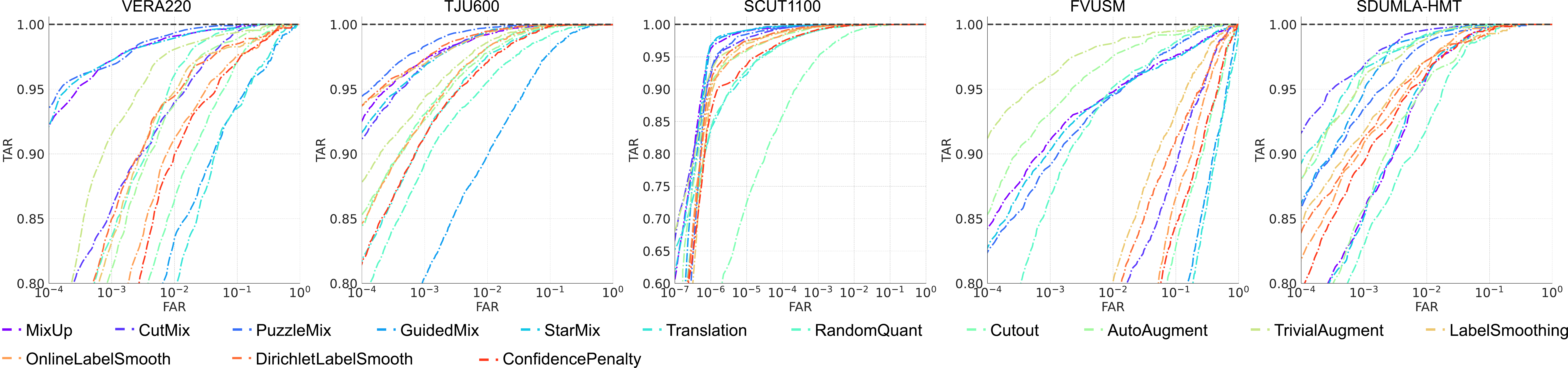}
    \caption{Receiver Operating Characteristic (ROC) curves of various data augmentation methods across five vein datasets using the ResNet18 backbone.}
    \label{fig:roc}
\end{figure*}

\begin{figure*}[t]
    \centering
    \includegraphics[width=1.0\linewidth]{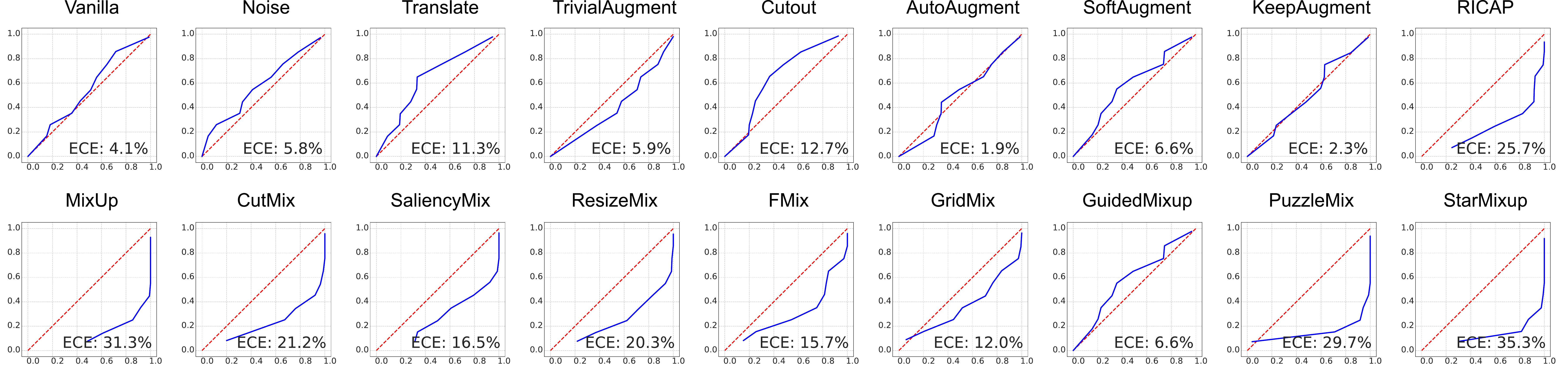}
    \vspace{-2.em}
    \caption{The confidence plots of different augmentations on the VERA220 dataset using ResNet18. The red line indicates the expected prediction tendency.}
    \vspace{-1.em}
    \label{fig:ece}
\end{figure*}

\subsection{Results of Robustness}
\label{sec:4.3}
Beyond recognition performance, we evaluate robustness across four complementary dimensions: calibration (Table~\ref{tab:calibration}), corruption (Table~\ref{tab:vera220_corruption}, Table~\ref{tab:tju600_corruption}), adversarial attacking (Table~\ref{tab:attack_tju600}, Table~\ref{tab:attack_scut1100}), and occlusion classification.

\textbf{Calibration:}  
Following the experiments of a lot of mixup methods~\cite{Verma2019manifold, Liu2022automix, qin2024adautomix, islam2024diffusemix}, we evaluate some well-performing vein classifiers with the calibration experiments, which aim to identify the consistency between the predictive confidence and the actual empirical accuracy of models. 
Figure.~\ref{fig:ece} shows that compared with multi-image augmentations, single-image augmentations could achieve a lower ECE score than a vanilla classifier. AutoAugment and KeepAugment achieve 1.9\% and 2.3\%, respectively. However, mixup-based methods are all worse than vanilla. Lots of them show an overconfident state. 
The full results were reported in Table~\ref{tab:calibration}.

While mixup has been reported to improve calibration in general object recognition by smoothing label distributions, its impact on the fine-grained topological features of vein images remains to be explored. Our experiments aim to reveal whether these augmentations help the vein classifier produce more ``honest'' confidence scores, thereby enhancing the reliability of the system in high-security, real-world deployment scenarios.

\textbf{Corruption:} To further evaluate the robustness, we conduct corruption experiments following the setting of AugMix~\cite{hendrycks2020augmix}. Table~\ref{tab:vear220_c1} and Table~\ref{tab:tju600_c1} show the C1 level results that mixup augmentations better than the single augmentations and label enhancement methods on both 2 datasets.
As shown in Tables~\ref{tab:vera220_corruption} and \ref{tab:tju600_corruption}, escalating corruption severity (C1--C3) degrades accuracy, with vanilla models collapsing below 30\% on TJU600 at C3. Despite this, MixUp-based methods (e.g., MixUp, PuzzleMix) demonstrate the most consistent robustness, substantially outperforming the vanilla baseline across all levels. Among single-image strategies, policy-based methods (notably TrivialAugment) provide the strongest resilience, whereas geometric augmentations frequently underperform the baseline even at mild levels. Label enhancement methods exhibit only moderate stability.
\begin{table}[t]
\caption{Top-1 Accuracy (\%) $\uparrow$ of corruption of various augmentations across different models on VEAR220-C1 dataset.}
\centering
\setlength{\tabcolsep}{1.4mm}
\resizebox{0.94\linewidth}{!}{
    \begin{tabular}{l ccccc}
    \toprule
    \textbf{Corruption}  & R18 & Mobv2 & FVN & APN& SLK-S \\
    \midrule
    Vanilla              & 69.59 & 69.19 & 39.09 & 73.18 & 65.10 \\
    \midrule
    Flip                 & \tg{63.49} & \tg{63.35} & \tg{30.21} & \tg{65.29} & \tg{60.29} \\
    Rotate               & \tg{62.75} & \tg{59.66} & \tg{32.18} & \tg{62.58} & \tg{58.33} \\
    Translation          & \tg{60.67} & \tg{64.14} & \tg{37.06} & \tg{63.54} & \tg{60.93} \\
    Noise                & \tg{67.32} & \tg{68.49} & \tg{37.18} & \tg{69.02} & 66.00 \\
    Cutout~\cite{devries2017cutout}               & \tg{68.06} & \tg{64.88} & \tg{33.95} & \tg{68.28} & \tg{62.82} \\
    GridMask~\cite{chen2020gridmask}             & \tg{64.93} & \tg{66.53} & \tg{24.81} & \tg{54.12} & \tg{54.71} \\
    RandomErasing~\cite{zhong2020randome}        & \tg{61.22} & \tg{66.89} & \tg{36.36} & \tg{72.54} & 65.86 \\
    RandomQuant~\cite{wu2023rq}          & \ul{79.59} & \tb{81.27} & \tg{22.27} & \ul{80.96} & 77.27 \\
    AutoAugment~\cite{cubuk2019autoaugment}          & 75.43 & 71.10 & \tg{35.12} & 76.56 & 72.82 \\
    RandAugment~\cite{cubuk2020randaugment}          & \tg{69.43} & 78.50 & \tg{21.72} & 75.60 & \tb{82.47} \\
    KeepAugment~\cite{gong2021keepaugment}          & 70.52 & 69.21 & \ul{41.97} & \tg{71.49} & \tg{63.63} \\
    TrivialAugment~\cite{muller2021trivialaugment}       & \tb{83.98} & \ul{78.78} & 41.33 & \tb{85.29} & \ul{81.99} \\
    TeachAugment~\cite{suzuki2022teachaugment}         & \tg{68.06} & \tg{57.19} & \tg{38.16} & \tg{56.84} & 65.31 \\
    SoftAugment~\cite{liu2023softaug}          & \tg{64.94} & \tg{68.21} & \tb{44.97} & 73.91 & 66.81 \\
    YOCO~\cite{han2022yoco}                 & \tg{62.06} & \tg{64.95} & \tg{29.35} & \tg{71.74} & \tg{58.83} \\
    \midrule
    RICAP~\cite{takahashi2019ricap}                & 73.15 & \tg{67.37} & \tg{28.27} & \tg{67.72} & \tg{61.02} \\
    MixUp~\cite{Zhang2018mixup}                & 85.51 & \cb{\tb{89.54}} & \cb{\tb{66.89}} & \ul{88.75} & \cb{\tb{85.81}} \\
    CutMix~\cite{Yun2019cutmix}               & 76.31 & 88.08 & \tg{35.65} & 77.01 & 70.47 \\
    FMix~\cite{Harris2020fmix}                 & 75.76 & 73.87 & 41.74 & 75.32 & 66.60 \\
    GridMix~\cite{Baek2021gridmix}              & 72.63 & \tg{69.17} & \tg{28.52} & \tg{67.33} & \tg{62.59} \\
    ResizeMix~\cite{Qin2020resizemix}            & 77.06 & 71.55 & \tg{34.02} & 77.72 & \tg{64.21} \\
    SaliencyMix~\cite{Uddin2020saliencymix}          & 77.42 & 75.05 & 42.47 & 78.55 & 68.96 \\
    PuzzleMix~\cite{Kim2020puzzle}            & \cb{\tb{87.05}} & \ul{88.47} & 60.10 & 84.77 & \ul{84.97} \\
    GuidedMixup~\cite{kang2023guidedmixup}          & \tg{61.85} & \tg{63.15} & \tg{22.61} & \tg{54.85} & \tg{61.98} \\
    StarMixup~\cite{jin2025starmixup}            & \ul{87.03} & 84.81 & \ul{66.49} & \cb{\tb{89.47}} & 84.66 \\
    \midrule
    LabelSmoothing~\cite{szegedy2016labelsmooth}       & \ul{76.51} & \ul{70.48} & \tb{51.84} & \ul{83.18} & \ul{66.38} \\
    OnlineLabelSmooth~\cite{zhang2021onlinelabelsmooth}    & 70.83 & \tg{67.14} & 48.56 & 76.01 & \tb{68.23} \\
    ConfidencePenalty~\cite{pereyra2017confidencepenalty}    & \tg{66.52} & \tg{65.17} & 45.54 & \tg{72.54} & \tg{60.91} \\
    DirichletLabelSmooth~\cite{cheng2021dirichletlabelsmoothLoss} & \tb{77.53} & \tb{71.53} & \ul{51.48} & \tb{84.94} & \tg{64.24} \\
    Bootstrapping~\cite{reed2014bootstrapp}        & \tg{61.85} & \tg{50.15} & 48.06 & \tg{71.49} & \tg{38.16} \\
    \bottomrule
    \end{tabular}
}
\label{tab:vear220_c1}
\end{table}
\begin{table}[t]
\caption{Top-1 Accuracy (\%) $\uparrow$ of corruption of various augmentations across different models on TJU600-C1 dataset.}
\centering
\setlength{\tabcolsep}{0.5mm}
\resizebox{1.0\linewidth}{!}{
    \begin{tabular}{l ccccccc}
    \toprule
    \textbf{Corruption}  & R18 & Mobv2 & FVN & APN& SLK-S & ViT-S & Swin-T \\
    \midrule
    Vanilla              & 76.99 & 77.64 & 45.82 & 70.11 & 60.90 & 57.05 & 63.47 \\
    \midrule
    Flip                 & \tg{67.48} & \tg{71.81} & \tg{39.58} & \tg{65.62} & \tg{52.43} & \tg{52.26} & \tg{59.35} \\
    Rotate               & \tg{72.36} & \tg{73.76} & \tg{39.74} & \tg{64.62} & \tg{55.62} & 58.75 & 64.13 \\
    Translation          & \tg{76.57} & \tg{76.25} & \tg{44.71} & \tg{68.70} & 61.87 & 59.65 & \tg{62.51} \\
    Noise                & 79.48 & 78.58 & 47.59 & \tg{63.63} & \ul{76.83} & 57.52 & 64.32 \\
    Cutout~\cite{devries2017cutout}               & \tg{75.82} & \tg{74.17} & 46.94 & \tg{64.47} & 68.08 & 58.92 & 64.38 \\
    GridMask~\cite{chen2020gridmask}             & \tg{71.11} & \tg{73.86} & \tg{39.40} & \tg{55.95} & 70.14 & 57.47 & \tg{63.33} \\
    RandomErasing~\cite{zhong2020randome}        & 79.53 & \tb{80.35} & 47.08 & \tb{75.83} & 67.57 & 58.20 & \tb{66.89} \\
    RandomQuant~\cite{wu2023rq}          & \tg{69.27} & \tg{76.97} & \ul{47.66} & \tg{67.39} & 72.37 & \tb{65.68} & \tg{47.00} \\
    AutoAugment~\cite{cubuk2019autoaugment}          & 77.11 & \tg{76.68} & \tg{42.08} & 70.22 & \tb{77.72} & \tg{52.97} & 64.53 \\
    RandAugment~\cite{cubuk2020randaugment}          & \ul{80.89} & 77.96 & \tg{45.45} & \ul{73.39} & 69.45 & \ul{65.25} & 65.64 \\
    KeepAugment~\cite{gong2021keepaugment}          & \tg{76.08} & 79.27 & \tg{44.26} & 70.83 & 74.81 & \tg{56.97} & 64.33 \\
    TrivialAugment~\cite{muller2021trivialaugment}       & \tb{82.67} & \ul{79.61} & \tb{48.62} & 73.22 & 74.72 & 64.40 & \ul{66.57} \\
    TeachAugment~\cite{suzuki2022teachaugment}         & \tg{71.21} & \tg{62.24} & \tg{33.52} & \tg{49.28} & \tg{58.22} & 64.16 & \tg{62.38} \\
    SoftAugment~\cite{liu2023softaug}          & 77.96 & 78.23 & 46.75 & 71.33 & 71.01 & \tg{56.73} & 65.58 \\
    YOCO~\cite{han2022yoco}                 & \tg{63.75} & \tg{67.72} & \tg{39.80} & \tg{68.13} & 65.21 & \tg{54.05} & \tg{51.64} \\
    \midrule
    RICAP~\cite{takahashi2019ricap}                & \tg{74.39} & \tg{71.76} & 48.40 & \tg{53.07} & 77.04 & \tg{46.15} & \tg{48.43} \\
    MixUp~\cite{Zhang2018mixup}                & \ul{88.72} & 88.63 & 61.52 & \ul{75.82} & 89.97 & \cb{\tb{75.80}} & \ul{80.19} \\
    CutMix~\cite{Yun2019cutmix}               & 83.12 & 83.50 & 54.30 & \tg{68.67} & 83.01 & 61.20 & 63.47 \\
    FMix~\cite{Harris2020fmix}                 & 82.50 & 79.51 & 53.00 & \tg{67.92} & 78.58 & 59.09 & 69.64 \\
    GridMix~\cite{Baek2021gridmix}              & 78.48 & 79.46 & \tg{30.89} & \tg{63.98} & 79.89 & 60.21 & 66.17 \\
    ResizeMix~\cite{Qin2020resizemix}            & 78.16 & \tg{76.65} & 48.23 & \tg{66.78} & 79.91 & 57.56 & \tg{61.24} \\
    SaliencyMix~\cite{Uddin2020saliencymix}          & 82.61 & 82.52 & 54.02 & 71.92 & 83.33 & 62.70 & 63.88 \\
    PuzzleMix~\cite{Kim2020puzzle}            & \tb{89.49} & \ul{88.92} & \cb{\tb{63.74}} & 75.04 & \ul{89.51} & \ul{74.24} & 73.81 \\
    GuidedMixup~\cite{kang2023guidedmixup}          & 63.44 & 60.42 & 23.50 & 54.54 & 63.09 & 56.00 & 40.75 \\
    StarMixup~\cite{jin2025starmixup}            & 88.64 & \cb{\tb{89.99}} & \ul{62.22} & \tb{78.95} & \cb{\tb{91.39}} & 73.73 & \cb{\tb{80.34}} \\
    \midrule
    LabelSmoothing~\cite{szegedy2016labelsmooth}       & \ul{89.53} & \ul{88.47} & \tb{59.15} & \cb{\tb{82.44}} & \ul{86.39} & \tb{60.95} & \ul{76.80} \\
    OnlineLabelSmooth~\cite{zhang2021onlinelabelsmooth}    & 79.33 & 80.50 & 47.70 & 75.32 & 81.60 & 59.62 & 69.21 \\
    ConfidencePenalty~\cite{pereyra2017confidencepenalty}    & 77.00 & 77.98 & \tg{45.66} & 73.55 & 74.36 & 58.12 & 64.06 \\
    DirichletLabelSmooth~\cite{cheng2021dirichletlabelsmoothLoss} & \cb{\tb{89.73}} & \tb{88.95} & \ul{58.38} & \ul{81.26} & \tb{87.47} & \ul{60.81} & \tb{77.56} \\
    Bootstrapping~\cite{reed2014bootstrapp}        & \tg{61.58} & \tg{50.43} & \tg{19.64} & 72.39 & \tg{55.50} & 57.64 & 64.58 \\
    \bottomrule
    \end{tabular}
}
\label{tab:tju600_c1}
\end{table}

\textbf{Adversarial Attack:}
To further explore adversarial attack resilience, we evaluate the Fast Gradient Sign Method (FGSM) and Projected Gradient Descent (PGD). FGSM uses an $\ell_{\infty}$ constraint with $\epsilon=0.2/255$. PGD is configured with $\epsilon=0.2/255$, $\alpha=0.05/255$, 10 iterations and the $\ell_{\infty}$ norm.

As shown in Table~\ref{tab:attack_tju600} and Table~\ref{tab:attack_scut1100}, our results indicate a striking dissociation between clean accuracy and adversarial robustness. Label enhancement methods, LabelSmoothing and DirichletLabelSmooth, exhibit the strongest adversarial robustness, with LabelSmoothing achieving 75.18\% and 70.37\% by FGSM and PGD on TJU600 under ResNet18, nearly doubling the Vanilla baseline of 40.93\% and 30.45\%.
Conversely, mixup-based methods suffer counterintuitive adversarial fragility: MixUp drops to 18.80\% and 4.87\% on TJU600 under ResNet18, well below Vanilla, because soft-label training blurs inter-class boundaries and expands the gradient-accessible attack surface.

\begin{table*}[t]
\caption{Top-1 Accuracy (\%) $\uparrow$ of various augmentations across different models under FGSM and PGD attacks on TJU600 dataset.}
\centering
\setlength{\tabcolsep}{2.0mm}
\resizebox{1.\linewidth}{!}{
    \begin{tabular}{lccccccccccccccc}
    \toprule
    \multirow{2}{*}{\textbf{TJU600}}
    & \multicolumn{2}{c}{\textbf{R18}}
    & \multicolumn{2}{c}{\textbf{Mobv2}}
    & \multicolumn{2}{c}{\textbf{FVN}}
    & \multicolumn{2}{c}{\textbf{APN}}
    & \multicolumn{2}{c}{\textbf{SLK-S}}
    & \multicolumn{2}{c}{\textbf{ViT-S}}
    & \multicolumn{2}{c}{\textbf{Swin-T}} \\
    \cmidrule(lr){2-3}
    \cmidrule(lr){4-5}
    \cmidrule(lr){6-7}
    \cmidrule(lr){8-9}
    \cmidrule(lr){10-11}
    \cmidrule(lr){12-13}
    \cmidrule(lr){14-15}
    & FGSM & PGD
    & FGSM & PGD
    & FGSM & PGD
    & FGSM & PGD
    & FGSM & PGD
    & FGSM & PGD
    & FGSM & PGD \\
    \midrule
    Vanilla              & 40.93 & 30.45 & 47.27 & 39.68 & 0.23 & 0.00 & 22.73 & 15.93 & 13.82 & 6.73 & 2.75 & 0.48 & 0.08 & 0.00 \\
    \midrule
    Flip                 & \tg{19.42} & \tg{9.85} & \tg{32.30} & \tg{22.65} & \tg{0.07} & \tg{0.00} & \tg{17.85} & \tg{11.30} & \tg{6.80} & \tg{1.08} & \tg{2.30} & 0.58 & \tg{0.08} & \tg{0.00} \\
    Rotate               & \tg{30.22} & \tg{18.77} & \tg{28.27} & \tg{15.47} & \tg{0.12} & \tg{0.00} & \tg{2.35} & \tg{0.30} & \tg{2.47} & \tg{0.17} & \tg{2.40} & \tg{0.38} & \tg{0.03} & \tg{0.00} \\
    Translation          & \tg{40.25} & \tg{30.23} & \tg{28.63} & \tg{12.98} & \tg{0.17} & \tg{0.00} & \tg{4.38} & \tg{0.32} & \tg{9.95} & \tg{3.57} & \tg{1.52} & \tg{0.20} & \tg{0.02} & \tg{0.00} \\
    Noise                & \tb{53.98} & \tb{49.55} & \tb{53.90} & \tb{48.07} & \tb{0.40} & \tb{0.07} & \tg{16.73} & \tg{10.47} & \tb{54.02} & \tb{49.52} & \tb{14.63} & \cb{\tb{9.77}} & \cb{\tb{8.52}} & \cb{\tb{4.23}} \\
    Cutout~\cite{devries2017cutout}               & \tg{32.80} & \tg{16.05} & \tg{28.67} & \tg{14.42} & \tg{0.17} & 0.02 & \tg{17.65} & \tg{9.03} & 25.82 & 10.58 & 4.58 & 1.25 & \tg{0.05} & \tg{0.00} \\
    GridMask~\cite{chen2020gridmask}             & 45.83 & 37.10 & \ul{49.73} & \ul{43.23} & \tg{0.12} & \tg{0.00} & \tg{15.15} & \tg{9.58} & 39.35 & 18.90 & 8.07 & 3.47 & \ul{2.75} & \ul{0.30} \\
    RandomErasing~\cite{zhong2020randome}        & 50.07 & 40.82 & \tg{46.17} & \tg{36.22} & 0.25 & \ul{0.03} & \tb{31.20} & \tb{24.05} & 25.52 & 13.77 & 3.17 & 0.55 & 0.13 & \tg{0.00} \\
    RandomQuant~\cite{wu2023rq}          & \tg{30.53} & \tg{16.60} & \tg{46.50} & \tg{34.92} & \tg{0.03} & \tg{0.00} & 27.45 & \ul{20.73} & 38.95 & 15.92 & \ul{13.68} & \ul{6.80} & 0.58 & 0.07 \\
    AutoAugment~\cite{cubuk2019autoaugment}          & \tg{37.23} & \tg{22.45} & \tg{25.48} & \tg{6.43} & \tg{0.00} & \tg{0.00} & \tg{22.53} & \tg{12.70} & 42.93 & 23.28 & \tg{0.67} & \tg{0.05} & \tg{0.02} & \tg{0.00} \\
    RandAugment~\cite{cubuk2020randaugment}          & 49.43 & 38.32 & \tg{22.10} & \tg{6.98} & \tg{0.00} & \tg{0.00} & 23.67 & \tg{14.08} & \tg{11.27} & \tg{1.37} & 3.20 & 0.58 & \tg{0.07} & \tg{0.00} \\
    KeepAugment~\cite{gong2021keepaugment}          & \tg{37.48} & \tg{26.60} & \tg{45.75} & \tg{36.47} & \tg{0.17} & \tg{0.00} & \tg{20.97} & \tg{14.00} & \ul{43.08} & \ul{31.45} & \tg{2.75} & 0.75 & \tg{0.05} & \tg{0.00} \\
    TrivialAugment~\cite{muller2021trivialaugment}       & \ul{53.20} & \ul{42.12} & \tg{25.48} & \tg{4.55} & \tg{0.08} & \tg{0.00} & \tg{15.55} & \tg{5.32} & 18.93 & \tg{5.57} & \tg{1.30} & \tg{0.08} & \tg{0.02} & \tg{0.00} \\
    TeachAugment~\cite{suzuki2022teachaugment}         & \tg{1.50} & \tg{0.00} & \tg{0.15} & \tg{0.02} & \tg{0.00} & \tg{0.00} & \tg{6.93} & \tg{0.00} & \tg{3.80} & \tg{0.87} & \tg{1.92} & \tg{0.32} & \tg{0.00} & \tg{0.00} \\
    SoftAugment~\cite{liu2023softaug}          & 46.68 & 37.65 & \tg{40.90} & \tg{30.43} & \ul{0.32} & \tg{0.00} & \ul{28.12} & 20.43 & 37.38 & 24.57 & \tg{2.27} & \tg{0.42} & \tg{0.05} & \tg{0.00} \\
    YOCO~\cite{han2022yoco}                 & \tg{13.47} & \tg{4.85} & \tg{4.40} & \tg{0.05} & \tg{0.05} & 0.02 & \tg{20.00} & \tg{8.50} & \tg{12.50} & \tg{1.25} & 2.85 & 0.50 & \tg{0.02} & \tg{0.00} \\
    \midrule
    RICAP~\cite{takahashi2019ricap}                & \tg{39.70} & \tg{16.43} & \tg{31.08} & \tg{6.70} & 0.57 & 0.02 & \tg{8.00} & \tg{4.05} & 30.80 & 9.95 & \tg{0.95} & \tg{0.20} & \tg{0.05} & \tg{0.00} \\
    MixUp~\cite{Zhang2018mixup}                & \tg{18.80} & \tg{4.87} & \tg{34.13} & \tg{12.33} & 0.48 & 0.12 & \tg{12.40} & \tg{3.45} & 24.82 & 7.22 & \ul{15.47} & \ul{5.23} & \ul{1.07} & \ul{0.02} \\
    CutMix~\cite{Yun2019cutmix}               & \tb{51.53} & \tg{\ul{28.13}} & \tg{\ul{46.02}} & \tg{17.73} & \ul{0.60} & \ul{0.13} & \tb{22.90} & \tg{13.08} & 41.27 & 14.37 & 6.23 & 0.95 & 0.27 & \tg{0.00} \\
    FMix~\cite{Harris2020fmix}                 & 42.58 & \tg{16.30} & \tg{32.75} & \tg{4.42} & 0.32 & 0.02 & \tg{20.87} & \tg{12.78} & 25.62 & \tg{4.68} & 7.80 & 2.63 & 0.18 & \tg{0.00} \\
    GridMix~\cite{Baek2021gridmix}              & 42.80 & \tg{26.32} & \tg{42.88} & \tg{\ul{26.85}} & \tg{0.10} & \tg{0.00} & \tg{8.30} & \tg{0.22} & 31.60 & 14.93 & 10.58 & 3.03 & 0.52 & \tg{0.00} \\
    ResizeMix~\cite{Qin2020resizemix}            & 48.25 & \tg{22.13} & \tg{38.68} & \tg{4.28} & \tg{0.05} & \tg{0.00} & \tg{18.02} & \tg{10.13} & 24.42 & \tg{2.58} & 7.03 & 1.60 & 0.17 & \tg{0.00} \\
    SaliencyMix~\cite{Uddin2020saliencymix}          & \ul{51.47} & \tg{\tb{29.93}} & \tg{42.47} & \tg{15.02} & 0.57 & 0.03 & \ul{22.88} & \tg{\tb{14.67}} & \ul{45.65} & \ul{19.90} & 10.17 & 2.83 & 0.43 & \tg{0.00} \\
    PuzzleMix~\cite{Kim2020puzzle}            & \tg{30.75} & \tg{6.70} & \tg{37.95} & \tg{8.32} & 0.35 & 0.02 & \tg{16.92} & \tg{9.35} & 30.35 & \tg{1.12} & 5.32 & 0.95 & 0.55 & \ul{0.02} \\
    GuidedMixup~\cite{kang2023guidedmixup}          & \tg{2.95} & \tg{0.00} & \tg{0.47} & \tg{0.00} & \tg{0.00} & \tg{0.00} & \tg{10.38} & \tg{4.38} & \tg{0.92} & \tg{0.00} & \tg{1.08} & \tg{0.10} & \tg{0.00} & \tg{0.00} \\
    StarMixup~\cite{jin2025starmixup}            & \tg{31.67} & \tg{14.08} & \tb{56.12} & \tg{\tb{39.07}} & \tb{0.72} & \tb{0.15} & \tg{21.48} & \tg{\ul{13.18}} & \tb{50.68} & \tb{27.17} & \cb{\tb{15.95}} & \tb{5.37} & \tb{1.88} & \tb{0.17} \\
    \midrule
    LabelSmoothing~\cite{szegedy2016labelsmooth}       & \cb{\tb{75.18}} & \cb{\tb{70.37}} & \cb{\tb{73.35}} & \cb{\tb{69.43}} & \cb{\tb{2.00}} & \cb{\tb{0.30}} & \cb{\tb{55.60}} & \cb{\tb{45.73}} & \ul{67.15} & \ul{58.73} & \ul{13.17} & \ul{6.67} & \tb{3.87} & \tb{0.73} \\
    OnlineLabelSmooth~\cite{zhang2021onlinelabelsmooth}    & 46.18 & 35.68 & 52.78 & 45.37 & 0.25 & 0.02 & 26.65 & 18.08 & 55.30 & 47.13 & 4.52 & 1.48 & 0.23 & \tg{0.00} \\
    ConfidencePenalty~\cite{pereyra2017confidencepenalty}    & 42.43 & 33.35 & 48.77 & 41.87 & 0.30 & \tg{0.00} & 26.58 & 19.40 & 41.33 & 30.12 & 2.92 & 0.52 & 0.10 & \tg{0.00} \\
    DirichletLabelSmooth~\cite{cheng2021dirichletlabelsmoothLoss} & \ul{74.30} & \ul{67.98} & \ul{70.55} & \ul{65.67} & \ul{1.73} & \ul{0.25} & \ul{52.48} & \ul{42.63} & \cb{\tb{70.80}} & \cb{\tb{63.48}} & \tb{13.28} & \tb{7.00} & \ul{3.52} & \ul{0.63} \\
    Bootstrapping~\cite{reed2014bootstrapp}        & \tg{23.20} & \tg{12.58} & \tg{6.35} & \tg{0.12} & 0.30 & 0.02 & 25.15 & 17.87 & \tg{3.42} & \tg{0.02} & 3.00 & 0.72 & \tg{0.07} & \tg{0.00} \\
    \bottomrule
    \end{tabular}
}
\vspace{-1.em}
\label{tab:attack_tju600}
\end{table*}

\begin{figure}[t]
    \centering
    \includegraphics[width=1.0\linewidth]{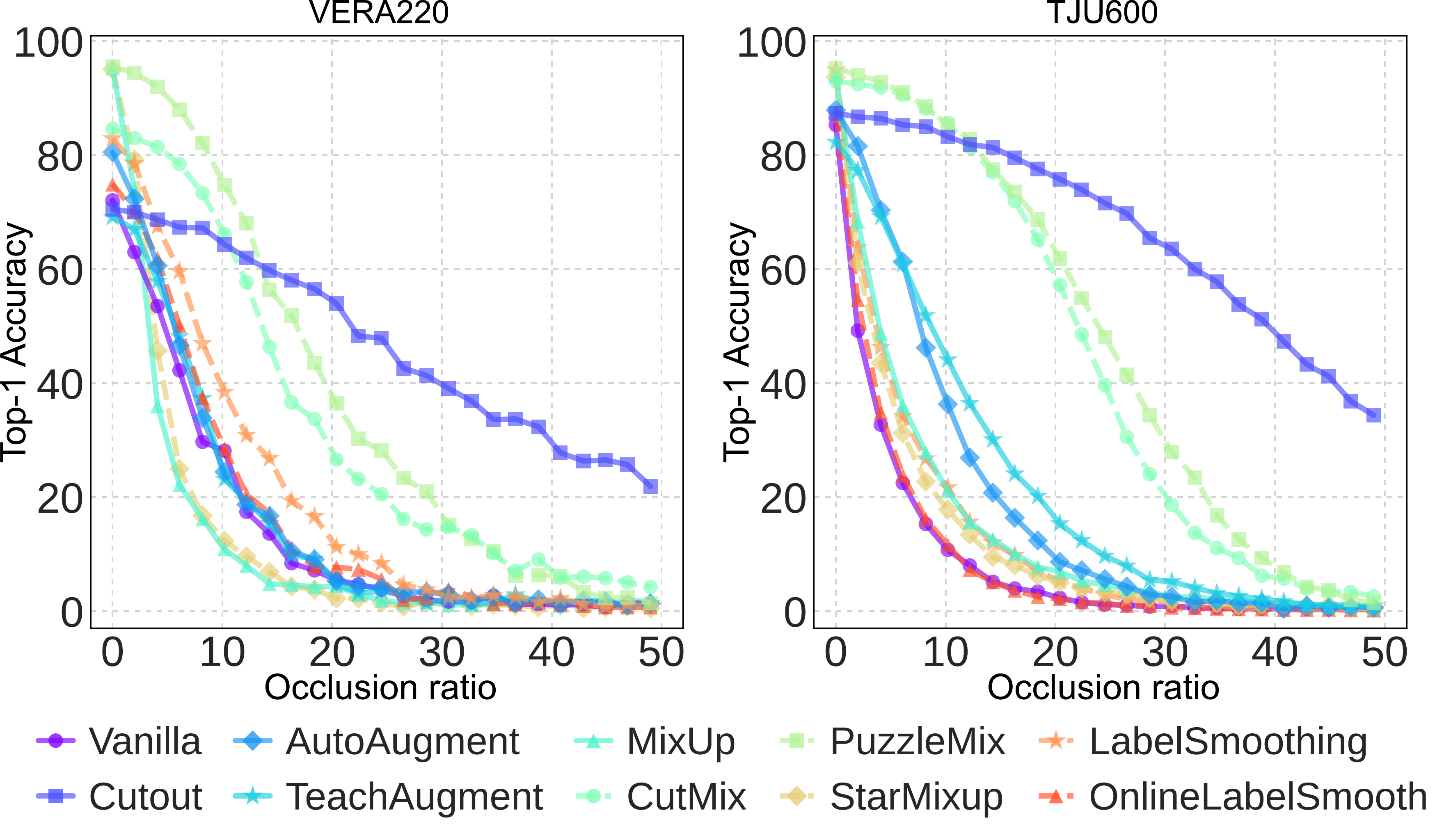}
    \vspace{-1.em}
    \caption{Top-1 Accuracy under varying occlusion ratios (0\% $\rightarrow$ 50\%) on the VERA220 and TJU600 datasets using ResNet18.}
    \vspace{-1.5em}
    \label{fig:occlusion_r18}
\end{figure}

\textbf{Occlusion Classification:}
To evaluate the robustness of augmentation methods against spatially continuous information loss, we conduct an occlusion classification experiment on the VERA220 and TJU600 datasets using ResNet18, where square regions are randomly masked with increasing ratios ($0\%$ to $50\%$). As shown in Figure \ref{fig:occlusion_r18}, cutting-based methods, including Cutout, CutMix, and PuzzleMix, maintain stable performance under increasing occlusion, while other methods degrade rapidly, indicating limited robustness to occlusion-induced spatial corruption and highlighting the effectiveness of masking-based augmentations in challenging scenarios.

\section{Analysis and Discussions}
\label{sec:6}

\subsection{Orthogonality of Augmentations}
\label{sec:6.1}
We study whether augmentations from different categories interact constructively by composing single-image, multi-image, and label-enhancement methods on ResNet18 with VERA220 and TJU600, as shown in Table~\ref{tab:multi_aug}.

Cross-category compositions consistently improve performance, indicating that different augmentation families provide complementary benefits. On VERA220, AutoAugment + LabelSmoothing improves accuracy from 80.82\% to 89.73\% and TAR@FAR from 65.09\% to 78.64\%. Label regularization also remains useful at high accuracy: MixUp + LabelSmoothing increases accuracy from 95.27\% to 97.18\%, while PuzzleMix + LabelSmoothing reduces EER from 0.83\% to 0.65\%. The full three-level composition (AutoAugment + PuzzleMix + LabelSmoothing) achieves the best overall results, reaching 98.00\%, 0.56\%, and 95.27\% on VERA220 and 96.50\%, 0.45\%, and 96.12\% on TJU600.

In contrast, intra-category combinations deliver marginal improvements. On VERA220, AutoAugment + MixUp only slightly outperforms vanilla MixUp (97.00\% \textit{vs.} 95.27\%), demonstrating that cross-category fusion works better.
\begin{table}[t]
\caption{Top-1 Accuracy (\%)$\uparrow$, EER (\%)$\downarrow$, and TAR@FAR=0.0001 (\%)$\uparrow$ of various composed augmentations on VERA220 and TJU600 datasets in ResNet18.}
\centering
\setlength{\tabcolsep}{1.mm}
\resizebox{1.\linewidth}{!}{
    \begin{tabular}{lccc ccc}
    \toprule
    \multirow{2}{*}{\textbf{Methods}}  &  \multicolumn{3}{c}{VERA220} &  \multicolumn{3}{c}{TJU600}  \\
    \cmidrule(lr){2-4} \cmidrule(lr){5-7}
                             & Acc. & EER & TAR@FAR & Acc. & EER & TAR@FAR \\
    
    \midrule
    Vanilla              & 71.45 & 5.20 & 51.00 & 85.55 & 1.72 & 81.23   \\
    \midrule
    % \multicolumn{7}{c}{\textit{\textbf{Single Image Augmentation with Label Enhancement}}} \\
    % \midrule
    AutoAugmentation     & 80.82 & 2.55 & 65.09 & 88.28 & 1.59 & 85.23    \\
    + LabelSmoothing     & 89.73 & 2.44 & 78.64 & 94.97 & 0.77 & 93.83   \\
    \midrule
    % \multicolumn{7}{c}{\textit{\textbf{Multi Image Augmentation with Label Enhancement}}} \\
    % \midrule
    MixUp                & 95.27 & 0.91 & 92.27 & 93.90 & 0.84 & 92.51   \\
    + LabelSmoothing     & 97.18 & 0.63 & 96.36 & 96.37 & 0.51 & 95.33   \\
    \midrule
    PuzzleMix            & 95.55 & 0.83 & 93.36 & 95.25 & 0.46 & 94.45    \\
    + LabelSmoothing     & 97.27 & 0.65 & 96.09 & 96.58 & 0.38 & 96.05   \\
    \midrule
    % \multicolumn{7}{c}{\textit{\textbf{Composed Augmentations}}} \\
    % \midrule
    AutoAugmentation     & 80.82 & 2.55 & 65.09 & 88.28 & 1.59 & 85.23   \\
    + MixUp              & 97.00 & 0.70 & 92.73 & 95.55 & 0.65 & 93.73   \\
    + LabelSmoothing     & 97.55 & 0.44 & 94.00 & 95.83 & 0.65 & 94.48   \\
    \midrule
    AutoAugmentation     & 80.82 & 2.55 & 65.09 & 88.28 & 1.59 & 85.23   \\
    + PuzzleMix          & 97.18 & 0.80 & 94.55 & 96.25 & 0.67 & 95.17   \\
    + LabelSmoothing     & 98.00 & 0.56 & 95.27 & 96.50 & 0.45 & 96.12   \\
    \bottomrule
    \end{tabular}
}
\label{tab:multi_aug}
\end{table}

\subsection{Efficiency Analysis}
\label{sec:6.2}
Our analysis is conducted on a lightweight backbone using the VERA220 dataset, with a standardized input resolution of $224 \times 224$ to ensure a fair comparison of overhead. Table~\ref{tab:vera220_mobilenetv2_apex} summarizes the efficiency--accuracy trade-off on MobileNetv2. Most methods keep the same inference cost as Vanilla (0.65 GFLOPs and 1.00M augmentation parameters), so their overhead mainly appears in training time and peak memory. MixUp offers the best balance, improving accuracy from 71.64\% to 95.55\% with nearly unchanged training time (5.04 \textit{vs.} 5.02 s/epoch). StarMixup is similarly efficient, reaching 92.55\% accuracy with near-Vanilla cost. PuzzleMix achieves the highest accuracy of 95.91\% but requires more computation and training time. In contrast, TeachAugment and SoftAugment incur high overhead or lower-than-Vanilla accuracy, indicating that lightweight mixing or quantization-based strategies are more practical for deployment-oriented vein recognition.

\subsection{Discussion}
\label{sec:6.3}
\textbf{\textit{When Do Augmentations Help?}} Augmentation efficacy is strongly conditioned on dataset scale and evaluation granularity. In data-scarce regimes (e.g., the VERA220 dataset with 2,200 images and the FV-USM dataset with 1,476 images), augmentation yields the largest absolute gains in accuracy. On larger datasets (e.g., SCUT1100 dataset with 11,000 images), vanilla baselines are already strong. Yet, augmentation continues to produce meaningful improvements at the harder metrics: MixUp reduces EER from 0.30\% to 0.07\% and boosts TAR@FAR=0.0001 from 97.30\% to 99.63\% on SCUT1100, revealing improvements that are invisible to accuracy-centric analysis. This observation confirms that stringent criteria like TAR@FAR=0.0001 and EER are the most sensitive and operationally relevant metrics for evaluating biometric augmentation strategies.
    
\textbf{\textit{Limitations of Current Augmentation Methods.}} Three systematic limitations characterize current augmentation strategies in the vein domain. First, geometric augmentations (Flip, Rotate, Translation, YOCO) consistently degrade performance across all datasets and architectures, as arbitrary spatial transformations violate the topological integrity of vein patterns that is essential for identity discrimination. Second, the highest accuracy methods (mixup-based) are simultaneously the worst calibrated (ECE >30\%) and the most adversarially fragile, e.g., MixUp drops to 4.87\% under PGD on TJU600, which reveals a multi-dimensional performance and robustness trade-off that no currently available single method resolves. Third, augmentation benefits are dataset-dependent: on SDUMLA-HMT, MixUp marginally underperforms the Vanilla baseline (79.87\% \textit{vs.} 84.51\% on ResNet18) while RICAP achieves the highest accuracy of 98.66\%, indicating that dataset-specific structural characteristics modulate which augmentation family is optimal and that blanket application of a single strategy across all vein datasets remains inadvisable.

\begin{table}[t]
  \caption{Efficient metrics and APEX Rank of augmentations by MobileNetv2 on the VEAR220 dataset.}
  \centering
  \setlength{\tabcolsep}{0.2mm}
  \resizebox{1.0\linewidth}{!}{
      \begin{tabular}{l cccccc}
      \toprule
      \textbf{Methods} & GFLOPs & Acc. $\uparrow$ & $T_\text{train}\downarrow$ &
  \textbf{$P_\text{aug}$} & $M_\text{peak}\downarrow$ & Rank \\
     \midrule
     Vanilla & 0.65 & 71.64 & 5.02 & 1.00 & 2507.10 & $-$ \\
     Flip & 0.65 & \tg{67.32} & \ul{\tg{5.03}} & 1.00 & \tg{2524.83} & \tg{3} \\
     Rotate & 0.65 & \tg{63.05} & \tg{5.28} & 1.00 & 2507.10 & \tg{4} \\
     Translation & 0.65 & \tg{65.45} & \tg{5.23} & 1.00 & 2507.10 & \tg{3} \\
     Noise & 0.65 & \tg{69.73} & \tg{5.04} & 1.00 & \tg{2566.45} & \tg{3} \\
     Cutout~\cite{devries2017cutout} & 0.65 & \tg{67.55} & \tg{6.57} & 1.00 & \tg{2524.83} & \tg{4} \\
     GridMask~\cite{chen2020gridmask} & 0.65 & \tg{68.50} & \tg{6.82} & 1.00 & \tg{2524.83} & \tg{4} \\
     RandomErasing~\cite{zhong2020randome} & 0.65 & \tg{69.86} & \tg{5.06} & 1.00 & 2488.38 & \ul{\tg{2}} \\
     RandomQuant~\cite{wu2023rq} & 0.65 & 82.23 & \tg{5.13} & 1.00 & \cb{\tb{2483.50}} & \cb{\tb{1}} \\
     AutoAugment~\cite{cubuk2019autoaugment} & 0.65 & 76.91 & \tg{7.81} & 1.00 & \tg{2507.72} & \tg{4} \\
     RandAugment~\cite{cubuk2020randaugment} & 0.65 & 82.09 & \tg{5.49} & 1.00 & \ul{2485.71} & \ul{\tg{2}} \\
     KeepAugment~\cite{gong2021keepaugment} & \tg{1.31} & \tg{70.91} & \tg{5.45} & 1.00 & \tg{2549.48} & \tg{4} \\
     TrivialAugment~\cite{muller2021trivialaugment} & 0.65 & 82.45 & \tg{5.19} & 1.00 & 2507.10 & \ul{\tg{2}} \\
     TeachAugment~\cite{suzuki2022teachaugment} & \tg{3.82} & \tg{59.86} & \tg{9.92} & \tg{2.37} & \tg{12156.80} & \tg{5} \\
     SoftAugment~\cite{liu2023softaug} & 0.65 & \tg{69.82} & \tg{11.79} & 1.00 & \tg{2652.35} & \tg{5} \\
     YOCO~\cite{han2022yoco} & 0.65 & \tg{66.14} & \tg{6.68} & 1.00 & 2506.12 & \tg{3} \\
     RICAP~\cite{takahashi2019ricap} & 0.65 & \tg{70.82} & \tg{5.09} & 1.00 & \tg{2508.17} & \tg{3} \\
     MixUp~\cite{Zhang2018mixup} & 0.65 & \ul{95.55} & \tg{5.04} & 1.00 & \tg{2545.46} & \cb{\tb{1}} \\
     CutMix~\cite{Yun2019cutmix} & 0.65 & 77.41 & \tg{5.05} & 1.00 & \tg{2507.72} & \cb{\tb{1}} \\
     FMix~\cite{Harris2020fmix} & 0.65 & 75.55 & \tg{5.05} & 1.00 & \tg{2524.83} & \ul{\tg{2}} \\
     GridMix~\cite{Baek2021gridmix} & 0.65 & 75.14 & \tg{5.07} & 1.00 & \tg{2524.83} & \tg{3} \\
     ResizeMix~\cite{Qin2020resizemix} & 0.65 & 75.86 & \tg{5.04} & 1.00 & 2488.38 & \cb{\tb{1}} \\
     SaliencyMix~\cite{Uddin2020saliencymix} & 0.65 & 79.14 & \tg{5.24} & 1.00 & 2507.10 & \tg{3} \\
     PuzzleMix~\cite{Kim2020puzzle} & \tg{1.31} & \cb{\tb{95.91}} & \tg{8.66} & 1.00 & \tg{2578.32} & \cb{\tb{1}} \\
     GuidedMixup~\cite{kang2023guidedmixup} & \tg{1.31} & \tg{69.68} & \tg{7.33} & 1.00 & \tg{2538.22} & \tg{4} \\
     StarMixup~\cite{jin2025starmixup} & 0.65 & 92.55 & \tg{5.09} & 1.00 & 2507.10 & \cb{\tb{1}} \\
     LabelSmoothing~\cite{szegedy2016labelsmooth} & 0.65 & 75.95 & \ul{\tg{5.03}} & 1.00 & \tg{2547.46} & \ul{\tg{2}} \\
     OnlineLabelSmooth~\cite{zhang2021onlinelabelsmooth} & 0.65 & \tg{70.18} & \tg{5.08} & 1.00 & \tg{2566.63} & \tg{4} \\
     ConfidencePenalty~\cite{pereyra2017confidencepenalty} & 0.65 & \tg{70.86} & \cb{\tb{5.02}} & 1.00 & \tg{2507.72} & \ul{\tg{2}} \\
     DirichletLabelSmooth~\cite{cheng2021dirichletlabelsmoothLoss} & 0.65 & 77.59 & \ul{\tg{5.03}} & 1.00 & \tg{2524.83} & \cb{\tb{1}} \\
     Bootstrapping~\cite{reed2014bootstrapp} & 0.65 & \tg{52.59} & \ul{\tg{5.03}} & 1.00 & \tg{2507.72} & \tg{3} \\
    \bottomrule
    \end{tabular}
  }
  \label{tab:vera220_mobilenetv2_apex}
\end{table}
\textbf{\textit{Trade-off between Accuracy \& Robustness Metrics.}}
Strong recognition performance and robustness are rarely co-realized by the same augmentation method. MixUp dominates on Accuracy, EER, and TAR@FAR yet ranks among the worst in calibration. For example, ECE score >30\% on TJU600 and adversarial robustness 4.87\% under PGD;
LabelSmoothing provides the strongest adversarial defense but with the most miscalibrated. Table~\ref{tab:calibration} shows that ECE score up to 47.88\% on TJU600 under ResNet18. In this paper, no single method currently offers an adequate all-in-one solution, and designing augmentations that jointly optimize across recognition accuracy and robustness dimensions remains an open research direction.

\textbf{\textit{Performance Consistency Across Vein Modalities.}} 
TrivialAugment and RandomQuant consistently rank among the top-performing single-image methods regardless of palm-vein or finger-vein images.
We hypothesize that their mechanisms, such as tuning-free operation sampling and discrete quantization, potentially reduce the sensitivity of the model to sensor-specific noise and micro-illumination variations, making them highly adaptable across diverse datasets. In contrast, mixup-based methods are highly effective on palm vein datasets but exhibit degraded or reversed benefits on certain finger vein configurations (e.g., SDUMLA-HMT). This confirms that no single augmentation strategy is universally optimal, motivating modality-aware selection and multi-method composition as standard practices in future vein recognition systems.

\textbf{\textit{Extensibility to Broader Biometric Modalities.}} Beyond vein recognition, the highly modular architecture of the AGVBench framework provides a clear pathway for extension to other biometric modalities, such as fingerprint, iris, or periocular recognition. Because our open-source codebase is deeply decoupled based on the MMCV ecosystem, researchers can seamlessly integrate new biometric tasks by simply customizing the data preprocessing pipelines (\textcolor{codemaroon}{\texttt{.agvbench.datasets}}), while fully reusing the comprehensive suite of standardized augmentation modules and evaluation protocols. This potential extensibility establishes AGVBench not merely as a vein-specific benchmark but as a unified, foundational testbed for exploring representation learning and data augmentation across the broader biometric security community.
\section{Conclusion}
We have presented AGVBench, a large-scale benchmark evaluating 30 augmentation strategies for vein recognition across five datasets, seven deep learning architectures, and six evaluation dimensions. Our study indicates a trend where multi-image augmentations generally yield superior recognition performance, whereas severe geometric transformations often pose risks to vascular topology. More importantly, we reveal a critical inconsistency across reliability metrics: methods achieving state-of-the-art clean accuracy frequently exhibit poor confidence calibration and increased adversarial fragility, despite their resilience to certain natural corruptions. We hope AGVBench motivates future augmentation designs that jointly optimize clean accuracy, adversarial security, and model calibration for reliable vein recognition.

\bibliographystyle{unsrt}
\bibliography{references}

\section*{Appendix}
\label{sec:appendix}

\subsection{Full Evaluation Protocol}
\label{sec:appendix_protocol}
\textbf{Performance Metrics:}
To quantitatively evaluate the recognition performance and the robustness of the learned feature representations, we employ three primary metrics: Top-1 Accuracy (Acc.), Equal Error Rate (EER), and True Acceptance Rate at a False Acceptance Rate of $10^{-4}$ (TAR@FAR=0.0001). Accuracy measures the proportion of correctly identified samples among the total test queries, reflecting the basic classification capability of models. Considering that biometric verification is essentially a threshold-dependent matching task, we adopt EER as a more rigorous criterion, defined as the point where the False Acceptance Rate (FAR) equals the False Rejection Rate (FRR). A lower EER indicates a better balance between general security and convenience. Furthermore, to evaluate the reliability under highly stringent security scenarios of models, we report TAR@FAR=0.0001. This metric specifically highlights the capacity of the system to recognize genuine users while strictly suppressing impostors, signifying superior discriminative power in the embedding space.

\textbf{Robustness Evaluation:}
To assess the reliability and robustness of vein recognition models in real-world complex scenarios, AGVBench integrates a multi-dimensional robustness evaluation suite. We introduce these evaluations as follows.

\textit{1) Calibration:} 
In biometric security systems like vein recognition, a well-calibrated model is crucial because the output probability should reliably reflect the likelihood of a correct match. However, modern deep neural networks, while highly accurate, often suffer from overconfidence, where the predicted probabilities are systematically higher than the actual precision, particularly when trained with intense data augmentations. To quantitatively assess this, we adopt the Expected Calibration Error (ECE) as the primary metric. The ECE partitions the predictions into $M$ equally spaced bins (e.g., $M=10$) based on their confidence scores and calculates the weighted average of the gap between accuracy and average confidence of each bin:
\begin{equation}
    \begin{aligned}
        \text{ECE} = \sum_{m=1}^{M} \frac{\vert B_m \vert}{n} \vert \text{Acc.}(B_m) - \text{Conf.}(B_m) \vert,
    \end{aligned}
\end{equation}
where $n$ is the total number of samples and $\vert B_m \vert$ is the number of samples in the $m$-th bin. In our benchmark, we investigate how different augmentation strategies affect model calibration.

\textit{2) Corruption:}
In practical deployments, vein recognition systems frequently encounter unpredictable environmental and sensor-induced degradations. Therefore, the corruption evaluation aims to explore the robustness and generalization of models against these out-of-distribution image distortions. Following previous works~\cite{hendrycks2020augmix, Liu2022automix, qin2024adautomix} and the robustness evaluation protocol of ImageNet-C~\cite{hendrycks2019robustness}, our benchmark implements a suite of 19 distinct corruption methods, including various degradations such as Gaussian noise and motion blur. 
Our experiments (Full results in Tables~\ref {tab:tju600_corruption} and~\ref {tab:vera220_corruption}) reveal a critical vulnerability: existing vein recognition models suffer catastrophic performance collapse even at the lowest severity level (C3). We hypothesize that this sensitivity arises because even minor corruptions are sufficient to obliterate the fine-grained topological textures of veins that are essential for accurate identity inference. Consequently, to enable a more granular and meaningful assessment, we expanded the evaluation spectrum by introducing two novel, lower-intensity severity levels (C1/C2) specifically tailored to the fragility of vein patterns.

\textit{3) Adversarial Attacks:}
To assess the robustness of security and features in vein recognition systems under malicious perturbations, we introduce a series of white-box adversarial attack experiments. Specifically, we implement two representative attack algorithms: FGSM~\cite{goodfellow2014explainingfgsm} and PGD~\cite{madry2018towardspgd}.

Considering the low contrast of vein textures, we set the maximum perturbation to $\epsilon = 0.2/255$ under the $\ell_{\infty}$ norm, which represents a significant challenge for vein recognition. For the PGD attack, we employ a step size of $\alpha = 0.05/255$ and 10 iterations to iteratively search for imperceptible perturbations within the $\ell_{\infty}$ bound that maximize the classification loss and provoke model misclassification.

\textit{4) Occlusion:}
In practical scenarios, vein patterns are frequently obscured by sensor smudges or finger misalignment. A robust biometric model must therefore leverage global topological structures rather than relying on easily compromised local patches. To systematically evaluate this fault tolerance, we introduce a spatial occlusion assessment by randomly masking continuous square regions in the input images with zero values. Specifically, we vary the occlusion ratio (the masked area relative to the total image dimensions) from $0\%$ to $50\%$ with a $2\%$ increment, constructing $25$ distinct test subsets. We evaluate the performance with 10 augmentations on these subsets.

\textbf{Efficiency Evaluation:}
We further evaluate the computational efficiency of the classifier integrated with various augmentation strategies. We postulate that an ideal augmentation approach should achieve a superior trade-off between performance gain and resource consumption. To quantify this, we employ three key metrics: training time occupancy ($T_\text{train}$) per epoch, peak memory footprint ($M_\text{peak}$), and computational complexity measured in GFLOPs. We also consider the ratio of extra learnable parameters, denoted as $P_\text{aug}$, defined as $P_\text{aug} = \frac{P_\text{extra} + P_\text{base}}{P_\text{base}}$, where $P_\text{base}$ and $P_\text{extra}$ denote the number of parameters of the vanilla backbone and the number of parameters of the additional modules introduced by the augmentation strategies, respectively.

To provide a holistic evaluation of these diverse augmentation strategies, we introduce the \underline{A}ugmentation \underline{P}erformance and \underline{E}fficiency E\underline{x}cellence (\textbf{APEX}) rank based on the principle of Pareto Efficiency. Instead of an arbitrary weighted sum, we identify the Non-dominated solutions across the multi-dimensional objective space (Performance, $T_\text{train}$, $M_\text{peak}$, GFLOPs, and $P_\text{aug}$) as a multi-objective optimization problem.

Non-dominated augmentations are assigned Rank-1 APEX, such that no other augmentation can improve one metric without degrading at least one other. By iteratively removing Rank-1 individuals, the subsequent non-dominated sets are defined as Rank-2 and higher. This ranking paradigm classifies methods beyond pure accuracy based on their strategic location on the efficiency-performance frontier and effectively eliminates redundant methods with high computational costs but only marginal performance gains.
\begin{table*}[t]
\caption{Summary of the datasets used in AGVBench.}
\centering
\setlength{\tabcolsep}{1.2mm}
\renewcommand{\tabularxcolumn}[1]{m{#1}}
\newcolumntype{L}{>{\raggedright\arraybackslash}X}
\begin{tabularx}{\linewidth}{lccccX}
    \toprule
    Dataset & Modality & Subjects & Samples & Total Images & \multicolumn{1}{c}{Description} \\ 
    \midrule
    % --- Palm Vein Datasets ---
    SCUT1100~\cite{luo2024scutpv} & Palm & 550 & 20 & 11,000 & Captured under unconstrained, ``on-the-fly'' conditions using dynamic sweeping and dropping motions. It features significant out-of-plane rotations and grayscale variations. \\ \addlinespace
    TJU600~\cite{Zhang2018palmprint} & Palm & 300 & 40 & 12,000 & A two-session dataset collected in a half-sheltered, wedge-shaped space. It incorporates variations in hand posture, positioning, and illumination. \\ \addlinespace
    VERA220~\cite{Tome2015vera} & Palm & 110 & 10 & 2,200 & Two-session dataset from an open environment with ultrasound-regulated positioning; exhibits minor pose variations and is susceptible to ambient light. \\ 
    \midrule
    % --- Finger Vein Datasets ---
    FVUSM~\cite{asaari2014fvusm} & Finger & 123 & 12 & 1,476 & A two-session collection of index and middle finger images. It evaluates model robustness against temporal intra-class variations. \\ \addlinespace
    SDUMLA-HMT~\cite{yin2011sdumla} & Finger & 106 & 36 & 3,816 & A multi-modal dataset providing finger vein samples of six fingers per subject. It features variability in finger placement and orientation. \\ 
    \bottomrule
\end{tabularx}
\label{tab:datasets}
\end{table*}

\subsection{Additional Full Results}
This appendix reports the more complete experimental tables that are extend with the main paper, which including calibration, adversarial attack, corruption, more ROC curve and hyperparameters of augmentations. Unless otherwise specified, the notation follows the main text.

\begin{figure*}[!t]
    \centering
    \includegraphics[width=0.9\linewidth]{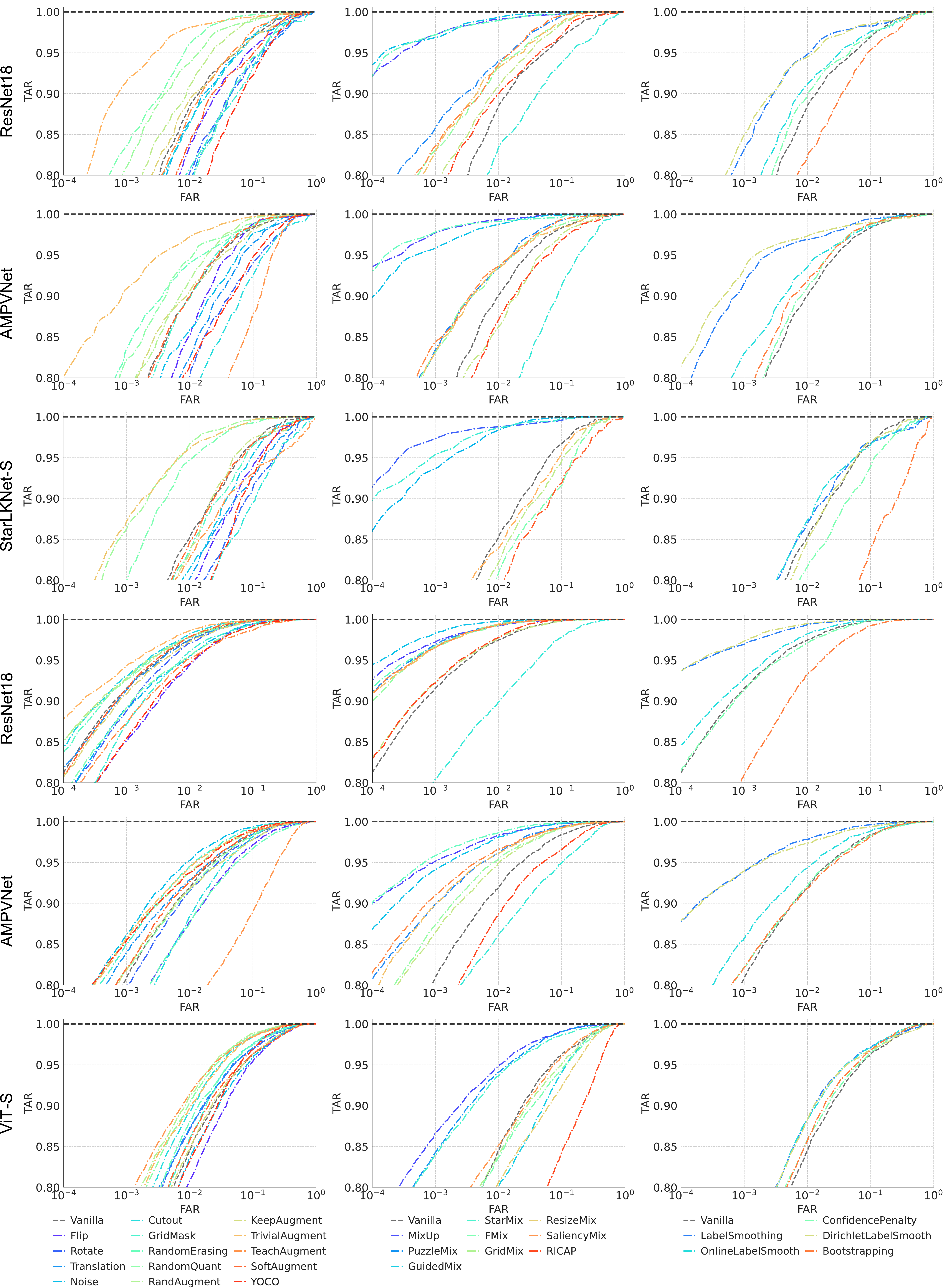}
    \caption{Receiver Operating Characteristic (ROC) curves of various data augmentation methods across two palm-vein datasets using different backbones. The results for VERA220 are shown in the top three rows, and the results for TJU600 are shown in the bottom three rows.}
    \label{fig:roc_vera220_tju600}
\end{figure*}

\begin{figure*}[!t]
    \centering
    \includegraphics[width=0.9\linewidth]{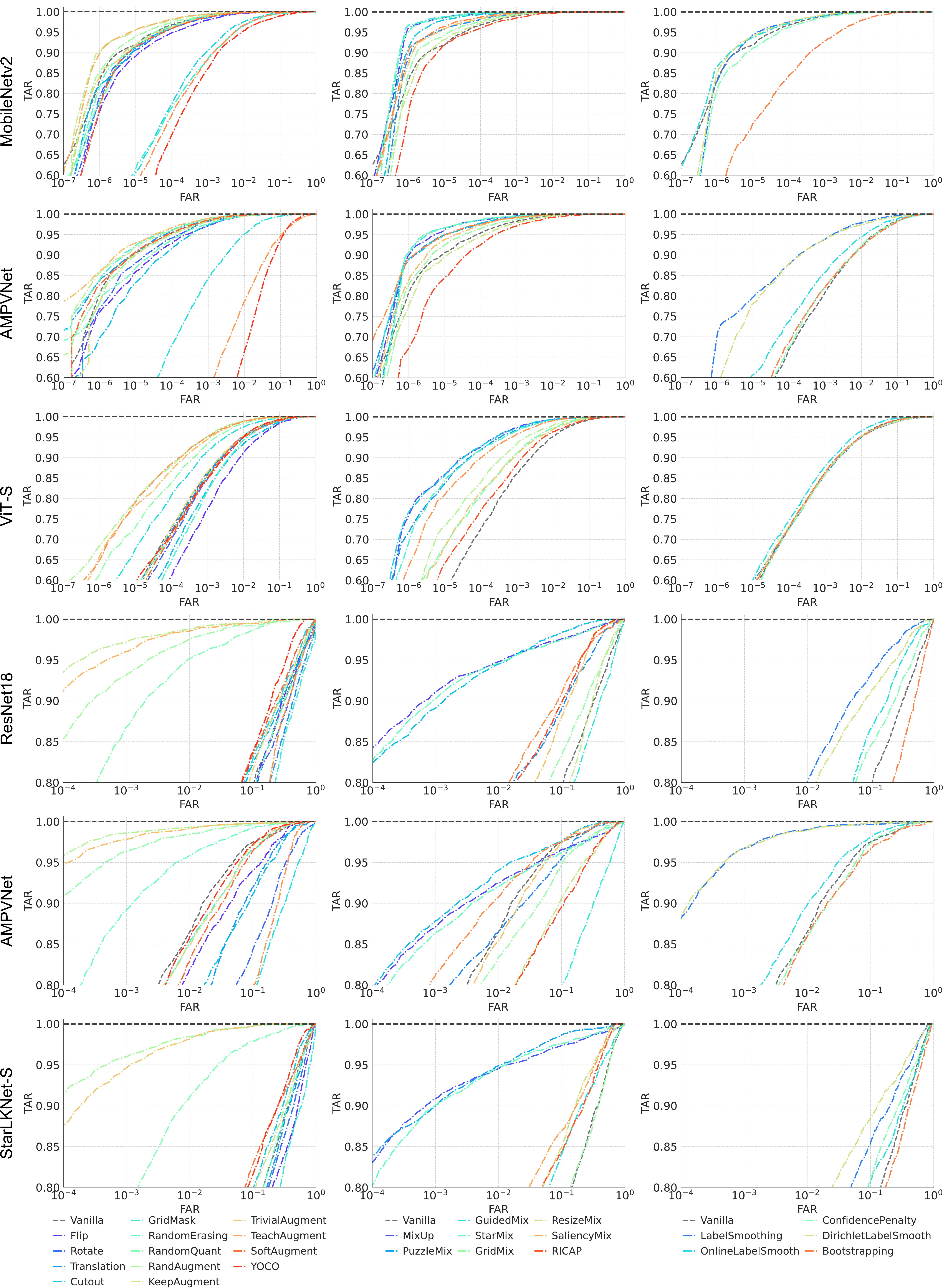}
    \caption{Receiver Operating Characteristic (ROC) curves of various data augmentation methods across two palm-vein datasets using different backbones. The results for SCTU1100 are shown in the top three rows, and the results for FV-USM are shown in the bottom three rows.}
    \label{fig:roc_scut1100_fvsum}
\end{figure*}

\begin{figure*}[!t]
    \centering
    \includegraphics[width=0.9\linewidth]{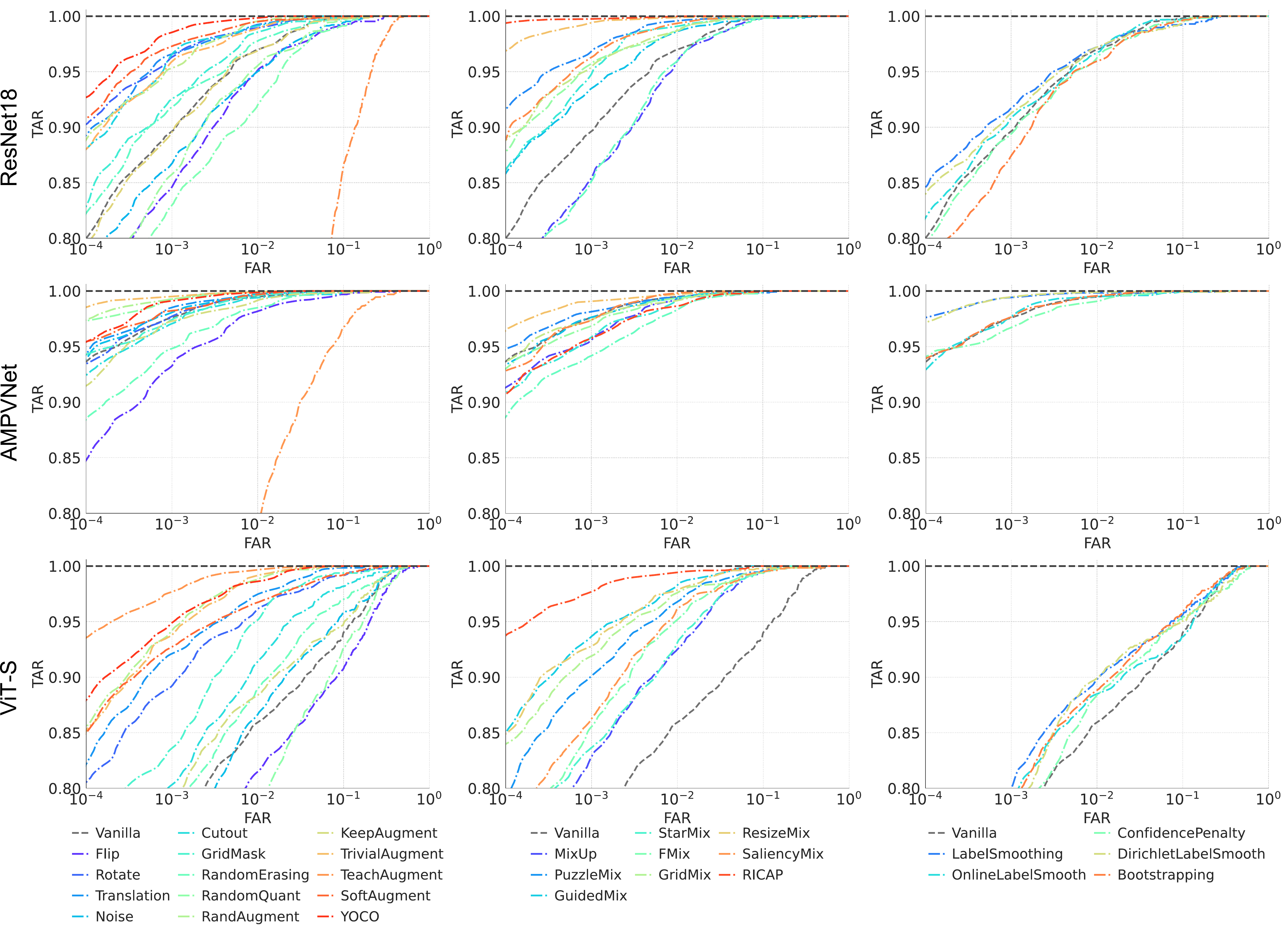}
    \caption{Receiver Operating Characteristic (ROC) curves of various data augmentation methods across SDUMLA-HMT datasets using different backbones. }
    \label{fig:roc_sdumla}
\end{figure*}

\begin{itemize}
    \item \textbf{Full Calibration Results.} Table~\ref{tab:calibration} reports the full calibration results measured by ECE across VERA220, TJU600, and SCUT1100. Lower ECE indicates better confidence calibration. For VERA220, results are reported for ResNet18, MobileNetv2, FVRASNet, AMPVNet, and StarLKNet-S. ViT-S and Swin-T are evaluated only on TJU600 and SCUT1100, where the larger-scale settings provide sufficient data for transformer-based backbones. These results expand the calibration discussion in the main paper and show that methods with high recognition accuracy are not necessarily well calibrated.
    
    \item \textbf{Full Adversarial Robustness Results.} Table~\ref{tab:attack_scut1100} reports the full SCUT1100 adversarial robustness results under FGSM and PGD attacks. Together with Table~\ref{tab:attack_tju600} in the main text, this table provides a complete view of adversarial behavior on the two vein datasets. The results further support the observation that recognition accuracy, calibration, corruption robustness, and adversarial robustness can favor different augmentation families.
    
    \item \textbf{Full Corruption Robustness Results.} Tables~\ref{tab:vera220_corruption} and~\ref{tab:tju600_corruption} provide the complete corruption robustness results under three corruption severity levels. The main paper reports the C1 results to keep the presentation compact, whereas the appendix includes C1, C2, and C3 for each backbone. These full results show how augmentation methods behave as the corruption level increases, which is important for distinguishing methods that only improve mild robustness from those that remain stable under severe degradation.

    \item \textbf{More ROC Curve Results.} Figures~\ref{fig:roc_vera220_tju600}, ~\ref{fig:roc_scut1100_fvsum} and ~\ref{fig:roc_sdumla} show more compressive results with different models.
\end{itemize}
\begin{table}[t]
\vspace{-7.em}
\centering
\caption{Representative hyperparameter settings used in AGVBench. Unless otherwise specified, default configurations from the original implementations are adopted.}
\label{tab:hyper_config}
\setlength{\tabcolsep}{1.mm}
\renewcommand{\arraystretch}{1.2}
\resizebox{\linewidth}{!}{
\begin{tabular}{llc}
\toprule
\textbf{Methods} & \textbf{Class} & \textbf{Configuration} \\
\midrule

Flip & Geometric & $p=0.5$ \\

Rotate & Geometric &
$M_{\rm level}=7$, $M_{\rm std}=0.5$ \\

Translation & Geometric &
$M_{\rm level}=7$, $M_{\rm std}=0.5$ \\

Noise & Pixel-level &
$p=0.1$ \\

RandomQuant & Pixel-level &
\makecell[c]{$N_{\rm region}=4$\\collapse: inside\_random} \\

Cutout & Cutting-based &
$\alpha=1.0$ \\

GridMask & Cutting-based &
$N_{\rm holes}\in[2,6]$ \\

RandomErasing & Cutting-based &
$p=0.25$, $s=(0.02,0.33)$ \\

AutoAugment & Policy-based &
25 sub-policies \\

RandAugment & Policy-based &
\makecell[c]{$N=2$, $M=7$\\std=0.5, max=10} \\

TrivialAugment & Policy-based &
max level=10 \\

YOCO & Policy-based &
$N_{\rm ops}=2$ \\

TeachAugment & Policy-based &
\makecell[c]{dropout $p=0.8$\\hidden $h=128$\\EMA $m=0.999999$} \\

KeepAugment & Saliency-based &
\makecell[c]{$\tau=0.5$, mode=paste\\$N_{\rm policy}=2$, $M=9$} \\

SoftAugment & Weighting-based &
$\tau_{\rm crop}=1.0$, $\sigma_{\rm crop}=12$ \\

RICAP & Cutting-based &
$N_{\rm images}=4$ \\

MixUp & Input-level &
$\alpha=1.0$ \\

CutMix & Cutting-based &
$\alpha=0.2$ \\

GridMix & Cutting-based &
$\alpha=0.2$ \\

FMix & Cutting-based &
$\alpha=0.2$ \\

SaliencyMix & Cutting-based &
$\alpha=0.2$ \\

StarMixup & Input-level &
$\alpha=1.0$ \\

ResizeMix & Cutting-based &
$\alpha=1.0$ \\

GuidedMixup & Input-level &
$\alpha=1.0$ \\

PuzzleMix & Feature-level &
\makecell[c]{$\alpha=2.0$, block=4\\transport=True, $\beta=1.2$} \\

LabelSmoothing & Loss Function &
$v=0.1$ \\

OnlineLabelSmoothing & Loss Function &
$\alpha=0.1$, mode=mean \\

ConfidencePenalty & Loss Function &
$\lambda_{\rm cp}=0.1$ \\

DirichletLabelSmoothing & Loss Function &
\makecell[c]{$v=0.1$, mode=soft\\Dirichlet $\alpha=0.1$} \\

Bootstrapping & Loss Function &
$\beta=0.1$ \\

\bottomrule
\end{tabular}
}
\end{table}

\begin{table*}[t]
    \caption{Results of calibration of various augmentations across different models on VEAR220, TJU600, and SCUT1100 datasets. ViT-S and Swin-T only with TJU600 and SCUT1100.}
    \centering
    \setlength{\tabcolsep}{0.8mm}
    \resizebox{1.0\linewidth}{!}{
        \begin{tabular}{lccccc ccccccc ccccccc}
        \toprule
        \multirow{2}{*}{\textbf{Calibration}}
        & \multicolumn{5}{c}{\textbf{VEAR220 ECE Score (\%) $\downarrow$}}
        & \multicolumn{7}{c}{\textbf{TJU600 ECE Score (\%) $\downarrow$}}
        & \multicolumn{7}{c}{\textbf{SCUT1100 ECE Score (\%) $\downarrow$}} \\
        \cmidrule(lr){2-6}
        \cmidrule(lr){7-13}
        \cmidrule(lr){14-20}
        & R18 & Mobv2 & FVN & APN & SLK-S
        & R18 & Mobv2 & FVN & APN & SLK-S & ViT-S & Swin-T
        & R18 & Mobv2 & FVN & APN & SLK-S & ViT-S & Swin-T \\
        \midrule
        Vanilla
        & 4.10 & 3.17 & 10.38 & 10.07 & 8.17 & 7.96 & 6.73 & 2.17 & 8.42 & 5.58 & 16.78 & 8.77 & 3.66 & 2.90 & 1.95 & 1.39 & 0.78 & 8.83 & 1.18 \\
        \midrule
        Flip
        & \tg{5.23} & \tg{6.02} & \tg{10.62} & \tg{13.48} & \tg{10.38} & 7.07 & 6.24 & \tg{3.93} & \tg{10.35} & \tg{6.00} & 16.27 & \tg{10.05} & 3.48 & 2.78 & 1.69 & \tg{2.17} & \tg{0.95} & \tg{12.87} & 0.66 \\
        Rotate
        & \tg{8.70} & \tg{9.91} & \tg{16.59} & \tg{15.77} & \tg{12.40} & 7.03 & \tg{6.95} & \tg{5.84} & \tg{9.84} & \tg{9.46} & 14.19 & \tg{9.37} & \tg{3.74} & \tg{3.34} & 1.27 & \tg{2.05} & 0.65 & \tg{10.03} & 1.01 \\
        Translation
        & \tg{11.33} & \tg{9.40} & \tg{18.85} & \tg{17.57} & \tg{12.18} & 7.72 & 5.44 & \tg{6.15} & \tg{8.96} & 5.26 & 14.00 & \tg{9.06} & \tg{4.09} & 2.84 & \ul{1.15} & \tg{1.83} & 0.66 & \tg{9.10} & \ul{0.49} \\
        Noise
        & \tg{5.80} & \tg{4.44} & \tg{15.54} & \tg{12.47} & \tg{9.82} & 7.70 & \tg{7.29} & \tg{3.75} & \tg{9.42} & 1.55 & \tg{16.86} & \tg{12.51} & \tg{6.40} & \tg{2.94} & \tg{2.52} & \tg{2.08} & 0.63 & \tg{10.51} & \tg{1.77} \\
        Cutout~\cite{devries2017cutout}
        & \tg{12.71} & \tg{15.44} & \tg{18.72} & \tg{14.29} & \tg{12.82} & \cb{\tb{0.94}} & 3.27 & \tg{10.98} & \tg{8.76} & 4.45 & 12.93 & 7.07 & \cb{\tb{1.13}} & \tg{3.25} & \tg{3.41} & \tg{4.17} & \tg{3.87} & 5.69 & 0.70 \\
        GridMask~\cite{chen2020gridmask}
        & \tg{11.15} & \tg{9.60} & \tg{22.00} & \tg{18.46} & \tg{16.33} & \ul{3.97} & \cb{\tb{1.19}} & \tg{17.31} & \tg{14.29} & \tg{5.64} & 16.46 & \tg{9.96} & 3.45 & \cb{\tb{2.26}} & \tg{6.99} & \tg{12.30} & \tg{4.67} & \tg{11.96} & \tg{2.54} \\
        RandomErasing~\cite{zhong2020randome}
        & \tg{9.45} & \tg{7.11} & \tg{16.03} & \tg{12.47} & \tg{9.90} & \tg{8.47} & \tg{6.78} & \tg{5.21} & 8.27 & 5.57 & 16.11 & \tg{8.78} & 3.51 & 2.80 & \cb{\tb{0.98}} & \tg{2.00} & \tg{0.90} & \tg{10.20} & 1.07 \\
        RandomQuant~\cite{wu2023rq}
        & \cb{\tb{1.62}} & \cb{\tb{1.49}} & \tg{11.73} & 5.70 & 5.44 & \tg{9.75} & 3.73 & \tg{11.33} & \tg{13.99} & 1.68 & 11.68 & \tg{16.79} & \tg{4.50} & \tg{3.73} & \tg{4.12} & \tg{3.54} & \tg{2.03} & 4.27 & 1.13 \\
        AutoAugment~\cite{cubuk2019autoaugment}
        & \ul{1.91} & \tg{3.44} & \ul{4.05} & 3.61 & \ul{2.78} & \tg{10.00} & 6.13 & \tg{5.98} & 7.43 & 1.71 & 14.92 & 6.16 & 3.11 & 2.55 & \tg{2.17} & 1.32 & \ul{0.56} & 3.71 & 0.54 \\
        RandAugment~\cite{cubuk2020randaugment}
        & 3.14 & \ul{2.78} & 5.53 & 4.98 & 3.16 & \tg{10.64} & 6.06 & \tg{4.54} & \ul{6.13} & 1.11 & \tb{9.84} & \cb{\tb{5.00}} & \tg{3.69} & \tg{3.08} & 1.56 & \tb{1.27} & \tg{1.28} & \cb{\tb{2.60}} & 0.55 \\
        KeepAugment~\cite{gong2021keepaugment}
        & 2.26 & \tg{4.55} & 9.56 & \ul{3.51} & \tg{11.53} & \tg{8.11} & \tg{7.01} & \tg{\ul{2.99}} & 8.09 & 2.83 & 16.13 & \tg{9.07} & 3.23 & 2.54 & 1.95 & \ul{1.31} & 0.65 & 8.45 & 0.67 \\
        TrivialAugment~\cite{muller2021trivialaugment}
        & \tg{5.88} & \tg{3.48} & \tb{2.96} & \tb{1.74} & \cb{\tb{1.34}} & \tg{9.92} & \tg{7.16} & \tg{2.22} & \tb{5.58} & \cb{\tb{0.54}} & 11.42 & \ul{5.77} & \ul{2.86} & \tg{3.01} & \tg{2.02} & 1.36 & \tg{1.02} & 3.46 & \cb{\tb{0.41}} \\
        TeachAugment~\cite{suzuki2022teachaugment}
        & 2.19 & \tg{8.10} & 4.58 & \tg{11.47} & \tg{9.89} & 5.58 & 3.50 & \tg{17.99} & \tg{10.89} & 4.92 & 10.14 & 6.77 & - & \tg{11.56} & - & 9.90 & - & \ul{2.61} & 0.57 \\
        SoftAugment~\cite{liu2023softaug}
        & \tg{6.65} & \tg{5.67} & 10.23 & \tg{10.24} & 9.45 & 4.65 & 4.36 & \tg{3.04} & \tg{9.23} & 4.13 & 15.07 & 8.63 & 2.95 & \ul{2.46} & 1.77 & \tg{1.71} & \cb{\tb{0.47}} & \tg{9.32} & 1.00 \\
        YOCO~\cite{han2022yoco}
        & \tg{15.98} & \tg{18.60} & \tg{22.95} & \tg{14.51} & \tg{14.17} & 5.61 & \ul{1.88} & \tg{12.01} & \tg{9.31} & \ul{1.08} & \ul{11.01} & \tg{10.56} & \tg{4.59} & \tg{3.02} & \tg{23.29} & \tg{42.93} & \tg{71.75} & 6.78 & 0.99 \\
        \midrule
        RICAP~\cite{takahashi2019ricap}
        & \tg{25.71} & \tg{18.99} & 7.19 & \tg{18.84} & \tg{18.08} & \tg{33.16} & \tg{35.17} & \tg{35.41} & \tg{36.57} & \tg{37.79} & \tg{26.46} & \tg{33.86} & \tg{9.57} & \tg{11.23} & \tg{17.25} & \tg{15.09} & \tg{9.77} & \tg{15.80} & \tg{8.22} \\
        MixUp~\cite{Zhang2018mixup}
        & \tg{31.26} & \tg{31.45} & \tg{37.74} & \tg{30.88} & \tg{35.60} & \tg{33.77} & \tg{36.09} & \tg{35.83} & \tg{35.59} & \tg{38.65} & \tg{26.56} & \tg{33.86} & \tg{8.93} & \tg{11.13} & \tg{17.03} & \tg{14.74} & \tg{9.14} & \tg{15.81} & \tg{8.22} \\
        CutMix~\cite{Yun2019cutmix}
        & \tg{21.20} & \tg{15.50} & 5.43 & \tg{13.15} & \tg{11.89} & \tg{24.99} & \tg{\ul{23.40}} & \tg{18.85} & \tg{20.65} & \tg{\ul{13.76}} & 8.87 & \tg{9.78} & \tg{13.91} & \tg{15.54} & \tg{11.33} & \tg{7.45} & \tg{7.84} & \tg{15.12} & \tg{6.29} \\
        FMix~\cite{Harris2020fmix}
        & \tg{15.73} & \tg{12.83} & 5.04 & \tg{12.82} & \ul{7.27} & \tg{26.47} & \tg{29.02} & \tg{20.61} & \tg{21.67} & \tg{20.87} & \ul{7.50} & \tg{10.01} & \tg{10.47} & \tg{11.15} & \tg{\ul{9.46}} & \tg{9.60} & \tg{7.61} & \tg{\ul{12.60}} & \tg{\tb{5.15}} \\
        GridMix~\cite{Baek2021gridmix}
        & \tg{\ul{11.99}} & \tg{12.23} & \ul{3.02} & \ul{3.47} & \tb{6.84} & \tg{17.30} & \tg{22.10} & \cb{\tb{2.09}} & \tg{11.73} & \tg{18.58} & \tb{2.79} & \tg{8.86} & \tg{11.44} & \tg{\ul{11.02}} & \tg{12.45} & \tg{\ul{7.37}} & \tg{8.47} & \tg{10.22} & \tg{6.62} \\
        ResizeMix~\cite{Qin2020resizemix}
        & \tg{20.33} & \tg{14.66} & \cb{\tb{2.92}} & \tg{14.42} & \tg{8.98} & \tg{\ul{21.65}} & \tg{25.10} & \tg{\ul{15.69}} & \tg{21.58} & \tg{14.76} & 13.07 & \tg{11.18} & \tg{9.68} & \tg{12.40} & \tg{11.49} & \tg{9.50} & \tg{\ul{7.04}} & \tg{18.37} & \tg{6.95} \\
        SaliencyMix~\cite{Uddin2020saliencymix}
        & \tg{16.53} & \tg{\ul{12.11}} & 4.10 & \tg{12.30} & \tg{8.46} & \tg{22.83} & \tg{25.08} & \tg{16.62} & \tg{\ul{17.44}} & \tg{\tb{13.49}} & 9.65 & \tg{\ul{9.61}} & \tg{\ul{8.91}} & \tg{11.05} & \tg{9.34} & \tg{6.26} & \tg{4.64} & \tg{13.65} & \tg{\ul{5.89}} \\
        PuzzleMix~\cite{Kim2020puzzle}
        & \tg{29.65} & \tg{28.93} & \tg{25.36} & \tg{25.76} & \tg{28.80} & \tg{31.13} & \tg{31.49} & \tg{23.27} & \tg{26.50} & \tg{40.65} & \tg{24.97} & \tg{24.91} & \tg{26.45} & \tg{25.82} & \tg{22.95} & \tg{19.28} & \tg{22.45} & \tg{26.54} & \tg{24.87} \\
        GuidedMixup~\cite{kang2023guidedmixup}
        & \tg{9.32} & \tg{10.57} & 8.41 & - & \tg{13.60} & \tg{33.40} & \tg{36.16} & \tg{36.05} & \tg{35.69} & \tg{39.87} & \tg{27.21} & \tg{33.86} & \tg{9.20} & \tg{11.09} & \tg{16.93} & \tg{12.86} & \tg{10.67} & \tg{15.93} & \tg{8.22} \\
        StarMixup~\cite{jin2025starmixup}
        & \tg{35.30} & \tg{33.69} & \tg{39.81} & \tg{37.06} & \tg{34.21} & \tg{38.34} & \tg{39.74} & \tg{38.80} & \tg{41.22} & \tg{45.50} & \tg{27.24} & \tg{35.75} & \tg{9.59} & \tg{12.42} & \tg{19.85} & \tg{15.29} & \tg{8.99} & \tg{15.89} & \tg{8.58} \\
        \midrule
        LabelSmoothing~\cite{szegedy2016labelsmooth}
        & \tg{38.42} & \tg{30.23} & \tg{25.16} & \tg{43.49} & 24.21 & \tg{47.88} & \tg{43.12} & \tg{36.33} & \tg{46.64} & \tg{33.45} & \tg{33.08} & \tg{47.57} & \tg{26.18} & \tg{21.07} & \tg{25.65} & \tg{25.24} & \tg{18.14} & \tg{36.40} & \tg{30.25} \\
        OnlineLabelSmooth~\cite{zhang2021onlinelabelsmooth}
        & \tg{4.41} & \tb{2.68} & \tb{3.88} & \ul{4.22} & 7.56 & \tg{11.74} & \tg{\ul{11.26}} & \tg{8.21} & \cb{\tb{4.22}} & \tg{14.52} & \cb{\tb{1.41}} & \ul{7.34} & \tg{5.74} & \tg{4.82} & \tg{6.77} & \tg{3.29} & \tg{\ul{6.32}} & \ul{5.21} & \tg{5.44} \\
        ConfidencePenalty~\cite{pereyra2017confidencepenalty}
        & \ul{3.57} & \tg{\ul{3.28}} & \ul{6.20} & 8.49 & 5.75 & \tg{16.99} & \tg{15.84} & \tg{\ul{5.96}} & \ul{4.91} & \tb{1.67} & \ul{10.81} & \tb{5.62} & \tg{\ul{7.03}} & \tg{\ul{6.58}} & \tg{\ul{4.22}} & \ul{0.92} & \tg{1.55} & \tb{4.90} & \tb{0.57} \\
        DirichletLabelSmooth~\cite{cheng2021dirichletlabelsmoothLoss}
        & \tg{39.10} & \tg{31.32} & \tg{25.35} & \tg{43.35} & 22.96 & \tg{47.80} & \tg{44.01} & \tg{38.53} & \tg{47.18} & \tg{34.69} & \tg{33.52} & \tg{47.47} & \tg{26.01} & \tg{22.38} & \tg{25.79} & \tg{25.36} & \tg{19.88} & \tg{36.62} & \tg{29.76} \\
        Bootstrapping~\cite{reed2014bootstrapp}
        & \tb{3.31} & \tg{4.14} & 8.56 & \tb{3.32} & 17.05 & \tg{\ul{15.92}} & \tg{10.05} & \tg{4.16} & 7.90 & \ul{2.95} & 15.80 & 8.05 & \tg{9.38} & \tg{11.90} & \tb{1.77} & \cb{\tb{0.76}} & \tg{7.69} & 8.45 & \ul{0.73} \\
        \bottomrule
        \end{tabular}
        }
    \label{tab:calibration}
    \end{table*}

\begin{table*}[t]
\caption{Top-1 Accuracy. (\%) $\uparrow$ of various augmentations across different models under FGSM and PGD attacks on SCUT1100 dataset.}
\centering
\setlength{\tabcolsep}{2.0mm}
\resizebox{1.\linewidth}{!}{
    \begin{tabular}{lccccccccccccccc}
    \toprule
    \multirow{2}{*}{\textbf{SCUT1100}}
    & \multicolumn{2}{c}{\textbf{R18}}
    & \multicolumn{2}{c}{\textbf{Mobv2}}
    & \multicolumn{2}{c}{\textbf{FVN}}
    & \multicolumn{2}{c}{\textbf{APN}}
    & \multicolumn{2}{c}{\textbf{SLK-S}}
    & \multicolumn{2}{c}{\textbf{ViT-S}}
    & \multicolumn{2}{c}{\textbf{Swin-T}} \\
    \cmidrule(lr){2-3}
    \cmidrule(lr){4-5}
    \cmidrule(lr){6-7}
    \cmidrule(lr){8-9}
    \cmidrule(lr){10-11}
    \cmidrule(lr){12-13}
    \cmidrule(lr){14-15}
    & FGSM & PGD
    & FGSM & PGD
    & FGSM & PGD
    & FGSM & PGD
    & FGSM & PGD
    & FGSM & PGD
    & FGSM & PGD \\
    \midrule
    Vanilla              & 90.38 & 89.86 & 85.80 & 83.75 & 8.73 & 0.14 & 86.20 & 85.27 & 86.42 & 85.76 & 49.91 & 46.40 & 43.66 & 30.36 \\
    \midrule
    Flip                 & \tg{87.53} & \tg{86.44} & \tg{83.13} & \tg{81.31} & 8.76 & 0.55 & \tg{82.14} & \tg{81.09} & \tg{80.44} & \tg{78.94} & \tg{42.89} & \tg{40.13} & \tg{35.26} & \tg{21.62} \\
    Rotate               & \tg{85.75} & \tg{84.00} & \tg{62.18} & \tg{48.62} & 15.40 & 2.67 & \tg{26.33} & \tg{9.74} & \tg{59.16} & \tg{45.18} & \tg{46.93} & \tg{42.95} & \tg{31.14} & \tg{16.20} \\
    Translation          & \tg{87.49} & \tg{86.18} & \tg{78.44} & \tg{71.09} & 21.78 & 2.49 & \tg{52.09} & \tg{33.96} & \tg{69.78} & \tg{61.00} & \tg{42.33} & \tg{37.27} & \tg{39.87} & \tg{25.40} \\
    Noise                & \tg{77.89} & \tg{77.40} & \tb{88.89} & \tb{88.25} & \cb{\tb{61.71}} & \cb{\tb{57.20}} & \tg{85.80} & 85.38 & \tb{88.78} & \tb{88.33} & 60.07 & 59.66 & \tb{74.67} & \tb{73.24} \\
    Cutout~\cite{devries2017cutout}               & \tg{65.44} & \tg{62.66} & \tg{57.76} & \tg{53.96} & \tg{8.38} & 0.24 & \tg{33.07} & \tg{16.34} & \tg{65.86} & \tg{62.96} & 63.44 & 61.00 & 45.60 & 31.64 \\
    GridMask~\cite{chen2020gridmask}             & \tg{71.09} & \tg{70.00} & \tg{69.07} & \tg{68.31} & 14.22 & 5.00 & \tg{15.09} & \tg{6.24} & \tg{64.89} & \tg{63.98} & 58.64 & 57.82 & 71.55 & 69.80 \\
    RandomErasing~\cite{zhong2020randome}        & 90.84 & \ul{90.36} & 86.78 & 85.20 & \ul{41.45} & 20.82 & \tb{88.87} & \tb{87.78} & \ul{86.53} & \tg{\ul{85.55}} & 52.16 & 49.45 & 46.76 & 33.11 \\
    RandomQuant~\cite{wu2023rq}          & \tg{87.02} & \tg{83.44} & 87.02 & \tg{82.62} & 38.69 & \ul{27.69} & \ul{86.53} & \ul{85.94} & \tg{85.55} & \tg{81.06} & \tb{73.55} & \tb{72.67} & \ul{74.11} & \ul{71.24} \\
    AutoAugment~\cite{cubuk2019autoaugment}          & \tg{86.31} & \tg{82.33} & \tg{77.47} & \tg{64.55} & \tg{4.86} & \tg{0.04} & \tg{71.91} & \tg{64.49} & \tg{66.71} & \tg{42.53} & 61.04 & 57.34 & \tg{11.51} & \tg{0.71} \\
    RandAugment~\cite{cubuk2020randaugment}          & \tg{89.11} & \tg{86.36} & \tg{82.56} & \tg{74.89} & \tg{3.64} & \tg{0.00} & \tg{81.42} & \tg{77.93} & \tg{74.40} & \tg{54.67} & \ul{65.93} & \ul{61.76} & \tg{32.18} & \tg{15.09} \\
    KeepAugment~\cite{gong2021keepaugment}          & 90.82 & 90.20 & 87.24 & 85.71 & \tg{8.54} & \tg{0.04} & \tg{85.27} & \tg{84.14} & \tg{86.16} & \tg{85.09} & \tg{49.76} & 46.85 & 44.87 & 31.27 \\
    TrivialAugment~\cite{muller2021trivialaugment}       & \tb{91.64} & \tg{89.44} & \tg{83.96} & \tg{77.84} & \tg{4.76} & \tg{0.02} & \tg{73.47} & \tg{65.71} & \tg{80.86} & \tg{75.11} & 63.16 & 59.00 & 44.33 & \tg{27.69} \\
    TeachAugment~\cite{suzuki2022teachaugment}         & \tg{0.26} & \tg{0.11} & \tg{20.11} & \tg{10.07} & \tg{0.00} & \tg{0.00} & \tg{15.53} & \tg{0.56} & \tg{0.16} & \tg{0.16} & 64.13 & 59.73 & \tg{14.93} & \tg{2.40} \\
    SoftAugment~\cite{liu2023softaug}          & \ul{91.36} & \tb{90.93} & \ul{87.75} & \ul{87.07} & 20.07 & 2.85 & \tg{86.18} & \tg{85.16} & \tg{85.93} & \tg{85.22} & \tg{48.27} & \tg{45.15} & \tg{39.66} & \tg{25.71} \\
    YOCO~\cite{han2022yoco}                 & \tg{24.69} & \tg{21.71} & \tg{39.42} & \tg{33.73} & \tg{1.76} & \tg{0.06} & \tg{10.22} & \tg{8.91} & \tg{0.98} & \tg{0.49} & \tg{48.29} & \tg{44.42} & \tg{25.42} & \tg{13.47} \\
    \midrule
    RICAP~\cite{takahashi2019ricap}                & 91.71 & \tg{89.24} & 86.27 & \tg{83.27} & 16.31 & 2.44 & \tg{65.07} & \tg{54.13} & 88.36 & \tg{83.53} & 57.02 & 54.47 & \tg{39.80} & \tg{26.05} \\
    MixUp~\cite{Zhang2018mixup}                & 91.73 & \tg{85.58} & \ul{92.27} & 87.69 & \tg{6.87} & \tg{0.02} & \tg{85.87} & \tg{76.25} & \tg{85.93} & \tg{75.66} & \ul{75.40} & 72.56 & \tg{27.31} & \tg{2.80} \\
    CutMix~\cite{Yun2019cutmix}               & \tb{95.53} & \tb{94.55} & 91.86 & \ul{90.13} & \tb{41.13} & \ul{9.67} & 90.87 & 89.84 & \ul{91.67} & \ul{89.82} & 72.24 & 70.22 & 58.89 & 41.09 \\
    FMix~\cite{Harris2020fmix}                 & 93.82 & 92.42 & 89.73 & 87.69 & 27.69 & 3.07 & 90.25 & 89.71 & \tg{82.45} & \tg{73.33} & 64.34 & 62.62 & \ul{61.85} & \ul{42.00} \\
    GridMix~\cite{Baek2021gridmix}              & 93.71 & 93.00 & 89.18 & 87.96 & 34.34 & 8.96 & \tg{79.16} & \tg{71.76} & 89.29 & 87.38 & 69.76 & 68.25 & 48.93 & 31.22 \\
    ResizeMix~\cite{Qin2020resizemix}            & \ul{94.84} & \ul{94.11} & \tb{92.45} & \tb{91.36} & \ul{38.80} & \tb{12.31} & \tb{92.09} & \tb{91.64} & \tb{92.47} & \tb{91.09} & 65.84 & 64.36 & 55.51 & 37.64 \\
    SaliencyMix~\cite{Uddin2020saliencymix}          & 94.78 & 93.60 & 91.64 & 89.27 & 33.76 & 8.80 & \ul{91.96} & \ul{91.20} & 90.91 & 87.31 & 74.64 & \cb{\tb{72.75}} & \tb{63.71} & \tb{44.84} \\
    PuzzleMix~\cite{Kim2020puzzle}            & 93.49 & \tg{89.55} & 88.33 & \tg{76.58} & \tg{6.62} & \tg{0.02} & \tg{85.45} & \tg{80.96} & \tg{84.34} & \tg{61.71} & 70.25 & 67.18 & 61.60 & 39.05 \\
    GuidedMixup~\cite{kang2023guidedmixup}          & 91.91 & \tg{85.62} & 90.09 & \tg{83.20} & \tg{7.18} & \tg{0.06} & \tg{78.62} & \tg{60.38} & 86.75 & \tg{78.20} & \cb{\tb{75.42}} & \ul{72.64} &  & 0.18 \\
    StarMixup~\cite{jin2025starmixup}            & 92.64 & \tg{87.11} & 92.02 & 88.56 & 8.93 & \tg{0.06} & \tg{83.75} & \tg{64.07} & 90.06 & \tg{84.02} & 74.38 & 71.31 & \tg{34.53} & \tg{3.54} \\
    \midrule
    LabelSmoothing~\cite{szegedy2016labelsmooth}       & \ul{96.34} & \ul{96.18} & \ul{93.00} & \cb{\tb{92.51}} & \ul{47.07} & \ul{11.73} & \ul{95.58} & \ul{95.49} & \ul{92.76} & \ul{92.27} & \ul{61.96} & \ul{61.16} & \ul{77.60} & \ul{73.18} \\
    OnlineLabelSmooth~\cite{zhang2021onlinelabelsmooth}    & 93.06 & 92.51 & 89.34 & 88.16 & 16.36 & 0.66 & 89.71 & 89.02 & 90.40 & 89.73 & 56.27 & 53.85 & 56.89 & 44.42 \\
    ConfidencePenalty~\cite{pereyra2017confidencepenalty}    & 91.64 & 91.13 & 87.16 & 85.86 & 11.31 & 0.20 & 86.29 & 85.49 & 86.69 & \tg{85.71} & 50.47 & 46.93 & 45.85 & 30.98 \\
    DirichletLabelSmooth~\cite{cheng2021dirichletlabelsmoothLoss} & \cb{\tb{96.53}} & \cb{\tb{96.44}} & \cb{\tb{93.04}} & \ul{92.47} & \tb{51.87} & \tb{17.42} & \cb{\tb{95.73}} & \cb{\tb{95.64}} & \cb{\tb{93.36}} & \cb{\tb{92.96}} & \tb{62.73} & \tb{61.87} & \cb{\tb{78.51}} & \cb{\tb{74.09}} \\
    Bootstrapping~\cite{reed2014bootstrapp}        & \tg{83.31} & \tg{80.91} & \tg{57.00} & \tg{48.45} & \tg{7.86} & \tg{0.13} & \tg{86.00} & \tg{84.98} & \tg{54.40} & \tg{44.27} & 50.55 & 47.31 & 43.84 & \tg{29.67} \\
    \bottomrule
    \end{tabular}
}
\label{tab:attack_scut1100}
\end{table*}
\begin{table*}[t]
\caption{Top-1 Accuracy (\%) $\uparrow$ of corruption of various augmentations across different models on VEAR220 dataset.}
\centering
\setlength{\tabcolsep}{1.5mm}
\resizebox{1.0\linewidth}{!}{
    \begin{tabular}{lccccccccccccccc}
    \toprule
    \multirow{2}{*}{\textbf{Corruption}}
    & \multicolumn{5}{c}{\textbf{VERA220-C1}}
    & \multicolumn{5}{c}{\textbf{VERA220-C2}}
    & \multicolumn{5}{c}{\textbf{VERA220-C3}} \\
    \cmidrule(lr){2-6}
    \cmidrule(lr){7-11}
    \cmidrule(lr){12-16}
    & R18 & Mobv2 & FVN & APN & SLK-S
    & R18 & Mobv2 & FVN & APN & SLK-S
    & R18 & Mobv2 & FVN & APN & SLK-S \\
    \midrule
    Vanilla
    & 69.59 & 69.19 & 39.09 & 73.18 & 65.10
    & 61.99 & 62.70 & 25.57 & 63.35 & 59.12
    & 46.70 & 50.69 & 9.81 & 41.96 & 48.61 \\
    \midrule
    Flip
    & \tg{63.49} & \tg{63.35} & \tg{30.21} & \tg{65.29} & \tg{60.29}
    & \tg{56.01} & \tg{56.60} & \tg{18.47} & \tg{56.82} & \tg{54.47}
    & \tg{40.05} & \tg{44.57} & \tg{6.77} & \tg{37.13} & \tg{43.95} \\
    Rotate
    & \tg{62.75} & \tg{59.66} & \tg{32.18} & \tg{62.58} & \tg{58.33}
    & \tg{53.66} & \tg{53.28} & \tg{20.91} & \tg{51.75} & \tg{52.13}
    & \tg{38.49} & \tg{40.53} & \tg{8.80} & \tg{35.86} & \tg{37.18} \\
    Translation
    & \tg{60.67} & \tg{64.14} & \tg{37.06} & \tg{63.54} & \tg{60.93}
    & \tg{53.90} & \tg{56.51} & \tg{25.10} & \tg{52.90} & \tg{55.14}
    & \tg{38.13} & \tg{43.52} & 10.86 & \tg{36.51} & \tg{42.97} \\
    Noise
    & \tg{67.32} & \tg{68.49} & \tg{37.18} & \tg{69.02} & 66.00
    & \tg{61.32} & \tg{61.63} & 28.42 & \tg{61.27} & 60.22
    & \ul{48.66} & \tg{50.38} & \tb{16.60} & 41.99 & 48.80 \\
    Cutout~\cite{devries2017cutout}
    & \tg{68.06} & \tg{64.88} & \tg{33.95} & \tg{68.28} & \tg{62.82}
    & \tg{58.73} & \tg{57.77} & \tg{21.51} & \tg{56.70} & \tg{56.05}
    & \tg{40.50} & \tg{43.64} & \tg{8.56} & \tg{38.28} & \tg{42.32} \\
    GridMask~\cite{chen2020gridmask}
    & \tg{64.93} & \tg{66.53} & \tg{24.81} & \tg{54.12} & \tg{54.71}
    & \tg{57.99} & \tg{59.86} & \tg{15.02} & \tg{44.40} & \tg{48.68}
    & \tg{45.02} & \tg{47.75} & \tg{6.63} & \tg{33.16} & \tg{37.61} \\
    RandomErasing~\cite{zhong2020randome}
    & \tg{61.22} & \tg{66.89} & \tg{36.36} & \tg{72.54} & 65.86
    & \tg{55.31} & \tg{60.41} & \tg{25.12} & 63.92 & 59.90
    & \tg{44.35} & \tg{50.00} & 11.17 & 44.83 & 49.47 \\
    RandomQuant~\cite{wu2023rq}
    & \ul{79.59} & \tb{81.27} & \tg{22.27} & \ul{80.96} & 77.27
    & \ul{68.83} & \tb{74.86} & \tg{15.10} & \ul{72.92} & 70.14
    & 48.56 & \cb{\tb{67.66}} & \tg{6.34} & \tb{59.09} & 56.91 \\
    AutoAugment~\cite{cubuk2019autoaugment}
    & 75.43 & 71.10 & \tg{35.12} & 76.56 & 72.82
    & 62.01 & \tg{61.55} & \tg{22.06} & 63.37 & 62.77
    & \tg{45.05} & \tg{42.99} & 10.38 & 48.04 & \tg{42.20} \\
    RandAugment~\cite{cubuk2020randaugment}
    & \tg{69.43} & 78.50 & \tg{21.72} & 75.60 & \tb{82.47}
    & \tg{57.73} & 69.41 & \tg{13.88} & \tg{62.09} & \tb{74.96}
    & \tg{41.42} & 52.28 & \tg{5.49} & 43.89 & \ul{58.33} \\
    KeepAugment~\cite{gong2021keepaugment}
    & 70.52 & 69.21 & \ul{41.97} & \tg{71.49} & \tg{63.63}
    & \tg{61.43} & \tg{60.77} & \ul{28.60} & \tg{62.34} & \tg{57.54}
    & \tg{45.24} & \tg{50.32} & 10.97 & 44.60 & \tg{45.21} \\
    TrivialAugment~\cite{muller2021trivialaugment}
    & \tb{83.98} & \ul{78.78} & 41.33 & \tb{85.29} & \ul{81.99}
    & \tb{73.18} & \ul{70.17} & 28.21 & \tb{73.74} & \ul{74.20}
    & \tb{55.45} & \ul{55.51} & 13.25 & \ul{54.91} & \tb{60.27} \\
    TeachAugment~\cite{suzuki2022teachaugment}
    & \tg{68.06} & \tg{57.19} & \tg{38.16} & \tg{56.84} & 65.31
    & \tg{60.67} & \tg{49.88} & 27.17 & \tg{50.11} & \tg{57.07}
    & \tg{45.00} & \tg{37.66} & 13.28 & \tg{38.12} & \tg{44.02} \\
    SoftAugment~\cite{liu2023softaug}
    & \tg{64.94} & \tg{68.21} & \tb{44.97} & 73.91 & 66.81
    & \tg{58.75} & \tg{61.88} & \tb{33.40} & 65.81 & 59.86
    & \tg{43.76} & \tg{49.88} & \ul{14.66} & 45.34 & \tg{48.06} \\
    YOCO~\cite{han2022yoco}
    & \tg{62.06} & \tg{64.95} & \tg{29.35} & \tg{71.74} & \tg{58.83}
    & \tg{53.25} & \tg{56.32} & \tg{21.55} & \tg{61.32} & \tg{52.32}
    & \tg{39.68} & \tg{43.32} & \tg{8.97} & 45.69 & \tg{40.86} \\
    \midrule
    RICAP~\cite{takahashi2019ricap}
    & 73.15 & \tg{67.37} & \tg{28.27} & \tg{67.72} & \tg{61.02}
    & 64.29 & \tg{59.57} & \tg{18.23} & \tg{59.20} & \tg{53.86}
    & 47.88 & \tg{46.16} & \tg{8.23} & 45.26 & \tg{40.91} \\
    MixUp~\cite{Zhang2018mixup}
    & 85.51 & \cb{\tb{89.54}} & \cb{\tb{66.89}} & \ul{88.75} & \cb{\tb{85.81}}
    & 75.05 & \cb{\tb{82.49}} & \ul{46.72} & \ul{81.16} & \cb{\tb{77.40}}
    & \ul{57.19} & \tb{65.86} & \ul{21.84} & \cb{\tb{62.82}} & \cb{\tb{60.89}} \\
    CutMix~\cite{Yun2019cutmix}
    & 76.31 & 88.08 & \tg{35.65} & 77.01 & 70.47
    & 68.04 & 80.89 & \tg{23.36} & 69.07 & 63.02
    & 51.81 & 62.11 & 10.03 & 51.49 & \tg{47.24} \\
    FMix~\cite{Harris2020fmix}
    & 75.76 & 73.87 & 41.74 & 75.32 & 66.60
    & 66.60 & 63.90 & 29.35 & 66.38 & \tg{58.33}
    & 49.52 & \tg{48.68} & 13.06 & 49.33 & \tg{44.02} \\
    GridMix~\cite{Baek2021gridmix}
    & 72.63 & \tg{69.17} & \tg{28.52} & \tg{67.33} & \tg{62.59}
    & 63.31 & \tg{60.10} & \tg{19.34} & \tg{59.39} & \tg{54.77}
    & \tg{46.68} & \tg{46.54} & \tg{7.87} & 44.40 & \tg{41.77} \\
    ResizeMix~\cite{Qin2020resizemix}
    & 77.06 & 71.55 & \tg{34.02} & 77.72 & \tg{64.21}
    & 68.99 & 63.37 & \tg{23.11} & 70.16 & \tg{56.75}
    & 51.10 & \tg{48.89} & 10.23 & 52.21 & \tg{42.92} \\
    SaliencyMix~\cite{Uddin2020saliencymix}
    & 77.42 & 75.05 & 42.47 & 78.55 & 68.96
    & 68.61 & 65.31 & 29.26 & 71.05 & 62.16
    & 51.48 & \tg{50.69} & 12.86 & 52.12 & \tg{48.16} \\
    PuzzleMix~\cite{Kim2020puzzle}
    & \cb{\tb{87.05}} & \ul{88.47} & 60.10 & 84.77 & \ul{84.97}
    & \cb{\tb{77.10}} & \ul{81.45} & 41.97 & 76.01 & 75.40
    & \cb{\tb{58.75}} & \ul{65.71} & 19.23 & 57.19 & \ul{59.18} \\
    GuidedMixup~\cite{kang2023guidedmixup}
    & \tg{61.85} & \tg{63.15} & \tg{22.61} & \tg{54.85} & \tg{61.98}
    & \tg{54.34} & \tg{55.19} & \tg{15.68} & \tg{47.88} & \tg{54.37}
    & \tg{40.67} & \tg{42.66} & \tg{7.56} & \tg{35.80} & \tg{41.14} \\
    StarMixup~\cite{jin2025starmixup}
    & \ul{87.03} & 84.81 & \ul{66.49} & \cb{\tb{89.47}} & 84.66
    & \ul{76.54} & 75.06 & \cb{\tb{47.96}} & \cb{\tb{82.02}} & \ul{76.49}
    & 57.18 & 56.09 & \cb{\tb{24.11}} & \ul{61.91} & 59.10 \\
    \midrule
    LabelSmoothing~\cite{szegedy2016labelsmooth}
    & \ul{76.51} & \ul{70.48} & \tb{51.84} & \ul{83.18} & \ul{66.38}
    & \ul{68.60} & \ul{62.09} & \tb{36.56} & \ul{73.80} & \ul{58.62}
    & \ul{50.10} & \ul{48.06} & \tb{14.93} & \ul{52.17} & \ul{45.62} \\
    OnlineLabelSmooth~\cite{zhang2021onlinelabelsmooth}
    & 70.83 & \tg{67.14} & 48.56 & 76.01 & \tb{68.23}
    & 63.34 & \tg{59.54} & 34.42 & 66.52 & \tb{60.59}
    & 47.96 & \tg{47.47} & \ul{14.88} & 48.04 & \tb{47.66} \\
    ConfidencePenalty~\cite{pereyra2017confidencepenalty}
    & \tg{66.52} & \tg{65.17} & 45.54 & \tg{72.54} & \tg{60.91}
    & \tg{59.04} & \tg{57.82} & 31.42 & \tg{63.78} & \tg{54.39}
    & \tg{43.90} & \tg{45.28} & 13.44 & 46.51 & \tg{43.62} \\
    DirichletLabelSmooth~\cite{cheng2021dirichletlabelsmoothLoss}
    & \tb{77.53} & \tb{71.53} & \ul{51.48} & \tb{84.94} & \tg{64.24}
    & \tb{69.65} & \tb{63.18} & \ul{35.40} & \tb{75.60} & \tg{56.55}
    & \tb{51.04} & \tb{49.20} & \ul{14.88} & \tb{53.47} & \tg{43.52} \\
    Bootstrapping~\cite{reed2014bootstrapp}
    & \tg{61.85} & \tg{50.15} & 48.06 & \tg{71.49} & \tg{38.16}
    & \tg{55.24} & \tg{43.20} & 33.56 & \tg{62.38} & \tg{33.45}
    & \tg{41.51} & \tg{32.74} & 13.88 & 46.33 & \tg{26.15} \\
    \bottomrule
    \end{tabular}
}
\label{tab:vera220_corruption}
\end{table*}
\begin{table*}[t]
\caption{Top-1 Accuracy (\%) $\uparrow$ of corruption of various augmentations across different models on TJU600 dataset.}
\centering
\setlength{\tabcolsep}{0.8mm}
\resizebox{1.0\linewidth}{!}{
    \begin{tabular}{lccccccccccccccccccccc}
    \toprule
    \multirow{2}{*}{\textbf{Corruption}}
    & \multicolumn{7}{c}{\textbf{TJU600-C1}}
    & \multicolumn{7}{c}{\textbf{TJU600-C2}}
    & \multicolumn{7}{c}{\textbf{TJU600-C3}} \\
    \cmidrule(lr){2-8}
    \cmidrule(lr){9-15}
    \cmidrule(lr){16-22}
    & R18 & Mobv2 & FVN & APN & SLK-S & ViT-S & Swin-T
    & R18 & Mobv2 & FVN & APN & SLK-S & ViT-S & Swin-T
    & R18 & Mobv2 & FVN & APN & SLK-S & ViT-S & Swin-T \\
    \midrule
    Vanilla
    & 76.99 & 77.64 & 45.82 & 70.11 & 60.90 & 57.05 & 63.47
    & 59.31 & 67.13 & 31.96 & 57.81 & 48.51 & 45.22 & 48.40
    & 30.01 & 42.52 & 12.84 & 36.87 & 24.43 & 25.29 & 22.26 \\
    \midrule
    Flip
    & \tg{67.48} & \tg{71.81} & \tg{39.58} & \tg{65.62} & \tg{52.43} & \tg{52.26} & \tg{59.35}
    & \tg{50.34} & \tg{59.77} & \tg{27.34} & \tg{52.70} & \tg{39.39} & \tg{40.84} & \tg{44.10}
    & \tg{21.48} & \tg{35.85} & \tg{11.23} & \tg{33.56} & \tg{10.79} & \tg{22.75} & \tg{19.20} \\
    Rotate
    & \tg{72.36} & \tg{73.76} & \tg{39.74} & \tg{64.62} & \tg{55.62} & 58.75 & 64.13
    & \tg{55.84} & \tg{61.23} & \tg{26.53} & \tg{51.27} & \tg{39.88} & 46.85 & 50.29
    & \tg{28.03} & \tg{39.53} & \tg{11.32} & 38.06 & \tg{16.86} & 28.75 & 25.31 \\
    Translation
    & \tg{76.57} & \tg{76.25} & \tg{44.71} & \tg{68.70} & 61.87 & 59.65 & \tg{62.51}
    & \tg{58.85} & \tg{62.83} & \tg{31.70} & \tg{54.50} & \tg{46.28} & 46.65 & \tg{47.54}
    & \tg{29.78} & \tg{40.96} & 12.88 & 39.67 & 26.80 & \tg{24.71} & \tg{22.09} \\
    Noise
    & 79.48 & 78.58 & 47.59 & \tg{63.63} & \ul{76.83} & 57.52 & 64.32
    & 66.68 & \ul{69.15} & \tb{34.72} & \tg{50.90} & \tb{67.34} & 50.50 & \tb{54.99}
    & \cb{\tb{44.38}} & 47.80 & 13.54 & \tg{36.53} & \tb{45.05} & \ul{34.97} & \cb{\tb{35.85}} \\
    Cutout~\cite{devries2017cutout}
    & \tg{75.82} & \tg{74.17} & 46.94 & \tg{64.47} & 68.08 & 58.92 & 64.38
    & \tg{56.95} & \tg{59.36} & 32.53 & \tg{51.28} & 52.38 & 46.67 & 49.19
    & \tg{25.25} & \tg{34.86} & 13.07 & 38.05 & \tg{23.48} & 26.96 & \tg{22.15} \\
    GridMask~\cite{chen2020gridmask}
    & \tg{71.11} & \tg{73.86} & \tg{39.40} & \tg{55.95} & 70.14 & 57.47 & \tg{63.33}
    & \tg{56.69} & \tg{64.08} & \tg{26.61} & \tg{45.24} & 58.69 & 47.04 & 51.83
    & 30.80 & 43.44 & \tg{11.87} & \tg{32.42} & 34.78 & 28.34 & \ul{34.63} \\
    RandomErasing~\cite{zhong2020randome}
    & 79.53 & \tb{80.35} & 47.08 & \tb{75.83} & 67.57 & 58.20 & \tb{66.89}
    & 63.53 & \tb{69.88} & 33.28 & \tb{65.07} & 53.50 & 46.56 & \ul{53.84}
    & 37.66 & 45.76 & 15.01 & 40.08 & 30.30 & 27.29 & 26.66 \\
    RandomQuant~\cite{wu2023rq}
    & \tg{69.27} & \tg{76.97} & \ul{47.66} & \tg{67.39} & 72.37 & \tb{65.68} & \tg{47.00}
    & \tg{54.49} & 69.13 & 33.35 & \ul{59.96} & 59.03 & \tb{58.27} & \tg{40.46}
    & \tg{23.54} & \cb{\tb{51.87}} & \ul{16.54} & 41.72 & \tg{23.66} & \cb{\tb{41.51}} & \tg{17.46} \\
    AutoAugment~\cite{cubuk2019autoaugment}
    & 77.11 & \tg{76.68} & \tg{42.08} & 70.22 & \tb{77.72} & \tg{52.97} & 64.53
    & 60.48 & \tg{63.35} & \tg{29.39} & \tg{56.71} & \ul{64.39} & \tg{41.28} & 52.02
    & 34.26 & \tg{34.72} & 14.68 & 44.41 & \ul{40.25} & \tg{21.89} & 24.49 \\
    RandAugment~\cite{cubuk2020randaugment}
    & \ul{80.89} & 77.96 & \tg{45.45} & \ul{73.39} & 69.45 & \ul{65.25} & 65.64
    & \tb{67.05} & \tg{66.13} & \tg{31.86} & 59.71 & 51.65 & \ul{53.46} & 52.71
    & 42.70 & \ul{49.28} & 16.36 & \tb{47.46} & 33.03 & 34.50 & 25.92 \\
    KeepAugment~\cite{gong2021keepaugment}
    & \tg{76.08} & 79.27 & \tg{44.26} & 70.83 & 74.81 & \tg{56.97} & 64.33
    & \tg{58.23} & 68.84 & \tg{31.07} & \tg{57.29} & 60.87 & 45.39 & 49.33
    & \tg{28.86} & 42.87 & \tg{12.24} & 37.30 & 35.95 & \tg{25.00} & 23.32 \\
    TrivialAugment~\cite{muller2021trivialaugment}
    & \tb{82.67} & \ul{79.61} & \tb{48.62} & 73.22 & 74.72 & 64.40 & \ul{66.57}
    & \ul{66.80} & \tg{65.38} & \ul{34.09} & 59.88 & 51.79 & 52.46 & 53.45
    & \ul{44.35} & 46.39 & \tb{17.96} & \ul{46.29} & 35.98 & 33.14 & 27.01 \\
    TeachAugment~\cite{suzuki2022teachaugment}
    & \tg{71.21} & \tg{62.24} & \tg{33.52} & \tg{49.28} & \tg{58.22} & 64.16 & \tg{62.38}
    & \tg{54.51} & \tg{49.08} & \tg{24.71} & \tg{38.36} & \tg{44.73} & 51.03 & \tg{47.42}
    & \tg{29.11} & \tg{21.80} & \tg{12.82} & \tg{27.58} & 24.43 & 28.43 & \tg{21.74} \\
    SoftAugment~\cite{liu2023softaug}
    & 77.96 & 78.23 & 46.75 & 71.33 & 71.01 & \tg{56.73} & 65.58
    & 62.06 & \tg{66.87} & 33.00 & 59.59 & 58.58 & 45.82 & 52.88
    & 33.33 & \tg{41.67} & 13.38 & 37.98 & 32.55 & 25.86 & 25.79 \\
    YOCO~\cite{han2022yoco}
    & \tg{63.75} & \tg{67.72} & \tg{39.80} & \tg{68.13} & 65.21 & \tg{54.05} & \tg{51.64}
    & \tg{47.30} & \tg{49.60} & \tg{26.83} & \tg{55.17} & \tg{47.20} & \tg{41.97} & \tg{36.90}
    & \tg{18.74} & \tg{19.11} & \tg{11.16} & 37.84 & \tg{16.67} & \tg{22.67} & \tg{15.65} \\
    \midrule
    RICAP~\cite{takahashi2019ricap}
    & \tg{74.39} & \tg{71.76} & 48.40 & \tg{53.07} & 77.04 & \tg{46.15} & \tg{48.43}
    & \tg{57.25} & \tg{53.63} & 34.20 & \tg{43.05} & 59.08 & \tg{35.22} & \tg{35.38}
    & \tg{26.70} & \tg{22.96} & 14.14 & \tg{30.18} & 28.54 & \tg{18.66} & \tg{15.03} \\
    MixUp~\cite{Zhang2018mixup}
    & \ul{88.72} & 88.63 & 61.52 & \ul{75.82} & 89.97 & \cb{\tb{75.80}} & \ul{80.19}
    & \cb{\tb{75.66}} & \ul{75.06} & 47.48 & \ul{63.22} & 76.76 & \cb{\tb{62.23}} & \cb{\tb{65.17}}
    & \tb{40.20} & \ul{43.42} & \cb{\tb{21.92}} & \cb{\tb{49.26}} & 40.33 & \tb{38.83} & \ul{34.65} \\
    CutMix~\cite{Yun2019cutmix}
    & 83.12 & 83.50 & 54.30 & \tg{68.67} & 83.01 & 61.20 & 63.47
    & 64.57 & \tg{64.81} & 38.41 & \tg{56.53} & 66.71 & 47.98 & 48.82
    & 31.39 & \tg{36.81} & 16.01 & 42.41 & 37.11 & 27.91 & \tg{21.91} \\
    FMix~\cite{Harris2020fmix}
    & 82.50 & 79.51 & 53.00 & \tg{67.92} & 78.58 & 59.09 & 69.64
    & 64.30 & \tg{63.52} & 37.82 & \tg{54.83} & 61.73 & 47.27 & 55.00
    & 34.77 & \tg{40.68} & 17.85 & 42.79 & 36.61 & 28.52 & 26.73 \\
    GridMix~\cite{Baek2021gridmix}
    & 78.48 & 79.46 & \tg{30.89} & \tg{63.98} & 79.89 & 60.21 & 66.17
    & 61.33 & 67.89 & \tg{23.62} & \tg{52.13} & 67.09 & 48.46 & 53.11
    & 31.84 & 43.14 & \tg{10.83} & 39.97 & 40.22 & 28.73 & 23.61 \\
    ResizeMix~\cite{Qin2020resizemix}
    & 78.16 & \tg{76.65} & 48.23 & \tg{66.78} & 79.91 & 57.56 & \tg{61.24}
    & 62.69 & \tg{58.66} & 35.29 & \tg{54.80} & 61.53 & \tg{45.12} & \tg{45.59}
    & 33.65 & \tg{34.08} & 16.80 & 41.75 & 33.02 & 27.67 & 24.81 \\
    SaliencyMix~\cite{Uddin2020saliencymix}
    & 82.61 & 82.52 & 54.02 & 71.92 & 83.33 & 62.70 & 63.88
    & 64.48 & \tg{63.66} & 38.12 & 58.23 & 65.64 & 49.66 & \tg{48.36}
    & 31.69 & \tg{32.41} & 16.07 & 43.14 & 35.42 & 30.37 & \tg{22.20} \\
    PuzzleMix~\cite{Kim2020puzzle}
    & \tb{89.49} & \ul{88.92} & \cb{\tb{63.74}} & 75.04 & \ul{89.51} & \ul{74.24} & 73.81
    & 73.25 & 74.62 & \cb{\tb{47.87}} & 61.82 & \ul{76.89} & \ul{60.25} & 56.35
    & 37.99 & \tg{39.86} & \ul{21.11} & 47.68 & \ul{40.92} & 37.95 & 27.80 \\
    GuidedMixup~\cite{kang2023guidedmixup}
    & 63.44 & 60.42 & 23.50 & 54.54 & 63.09 & 56.00 & 40.75
    & 50.74 & 47.41 & 16.48 & 46.42 & 49.53 & 45.11 & 35.49
    & 24.36 & 23.64 & 6.63 & 33.54 & 26.15 & 24.86 & 16.68 \\
    StarMixup~\cite{jin2025starmixup}
    & 88.64 & \cb{\tb{89.99}} & \ul{62.22} & \tb{78.95} & \cb{\tb{91.39}} & 73.73 & \cb{\tb{80.34}}
    & \ul{74.39} & \tb{77.97} & \ul{47.54} & \tb{65.34} & \cb{\tb{79.00}} & 60.12 & \ul{64.83}
    & \ul{40.01} & \tb{49.60} & 20.98 & \ul{48.90} & \cb{\tb{47.73}} & \ul{37.58} & \tb{35.72} \\
    \midrule
    LabelSmoothing~\cite{szegedy2016labelsmooth}
    & \ul{89.53} & \ul{88.47} & \tb{59.15} & \cb{\tb{82.44}} & \ul{86.39} & \tb{60.95} & \ul{76.80}
    & \tb{72.22} & \cb{\tb{78.85}} & \tb{43.23} & \cb{\tb{67.84}} & \ul{73.25} & \tb{48.29} & \ul{59.58}
    & \ul{40.13} & \cb{\tb{52.00}} & \tb{18.22} & \tb{47.53} & \ul{41.10} & \tb{27.96} & \tb{32.05} \\
    OnlineLabelSmooth~\cite{zhang2021onlinelabelsmooth}
    & 79.33 & 80.50 & 47.70 & 75.32 & 81.60 & 59.62 & 69.21
    & 61.26 & 70.15 & 33.66 & 62.08 & 69.68 & 47.44 & 53.16
    & 31.74 & 43.45 & 13.57 & 39.87 & 41.06 & 26.80 & 25.35 \\
    ConfidencePenalty~\cite{pereyra2017confidencepenalty}
    & 77.00 & 77.98 & \tg{45.66} & 73.55 & 74.36 & 58.12 & 64.06
    & 59.33 & 67.84 & \tg{31.93} & 62.00 & 60.00 & 46.27 & 49.25
    & \tg{29.95} & 43.11 & \tg{12.69} & 38.48 & 35.45 & 25.90 & 22.95 \\
    DirichletLabelSmooth~\cite{cheng2021dirichletlabelsmoothLoss}
    & \cb{\tb{89.73}} & \tb{88.95} & \ul{58.38} & \ul{81.26} & \tb{87.47} & \ul{60.81} & \tb{77.56}
    & \ul{72.13} & \ul{78.42} & \ul{42.96} & \ul{66.83} & \tb{75.49} & \ul{48.05} & \tb{60.10}
    & \tb{40.89} & \ul{50.15} & \ul{18.18} & \ul{47.36} & \tb{43.90} & \ul{27.85} & \ul{31.94} \\
    Bootstrapping~\cite{reed2014bootstrapp}
    & \tg{61.58} & \tg{50.43} & \tg{19.64} & 72.39 & \tg{55.50} & 57.64 & 64.58
    & \tg{45.15} & \tg{35.79} & \tg{12.46} & 60.06 & \tg{40.70} & 46.01 & 49.26
    & \tg{18.70} & \tg{15.26} & \tg{4.79} & 37.65 & \tg{18.18} & 25.58 & 23.18 \\
    \bottomrule
    \end{tabular}
    }
\label{tab:tju600_corruption}
\end{table*}

\clearpage
\newpage

\newpage
\clearpage

\end{document}